\documentclass[11pt]{article}

\usepackage{geometry}
\usepackage{amssymb} 
\usepackage[acronym]{glossaries}
\usepackage{setspace}
\usepackage{caption} 
\usepackage{graphicx}
\usepackage{subcaption} 
\usepackage{todonotes}
\usepackage{url}
\usepackage[normalem]{ulem}
\usepackage{enumerate}
\usepackage{tabularray}
\usepackage{makecell}
\usepackage{multirow}

\usepackage{mathptmx}
\usepackage{multirow}
\usepackage[T1]{fontenc}
\usepackage{booktabs}
\usepackage{lscape}

\usepackage{siunitx}

\usepackage[natbibapa]{apacite}

\newenvironment{biseabstract}{%
\begin{quote} \bf}
{\end{quote}}

\newenvironment{bisekeywords}{%
\begin{quote} \it \textbf{Keywords -}}
{\end{quote}}

\newcommand{\msd}[2]{\makecell[c]{#1 \\ \small{(#2)}}}

\title{Challenging the Performance-Interpretability Trade-off: An Evaluation of Interpretable Machine Learning Models}

\author{
Sven Kruschel$^{1}$, Nico Hambauer$^{1}$, Sven Weinzierl$^{2}$, Sandra Zilker$^{2,3}$, Mathias Kraus$^{1}$, Patrick Zschech$^{4\ast}$\\
\\
\normalsize{$^{1}$University of Regensburg, Chair for Explainable AI in Business Value Creation, }\\
\normalsize{\phantom{$^{1}$}Bajuwarenstraße 4, 93053 Regensburg, Germany}\\
\normalsize{$^{2}$Friedrich-Alexander-Universität Erlangen-Nürnberg, Chair of Digital Industrial Service Systems,}\\
\normalsize{\phantom{$^{2}$}Fürther Straße 248, 90429 Nuremberg, Germany}\\
\normalsize{$^{3}$Technische Hochschule Nürnberg Georg Simon Ohm, Professorship for Business Analytics,}\\
\normalsize{\phantom{$^{3}$}Hohfederstraße 40, 90489 Nuremberg, Germany}\\
\normalsize{$^{4}$Leipzig University, Professorship for Intelligent Information Systems and Processes,}\\
\normalsize{\phantom{$^{4}$}Grimmaische Straße 12, 04109 Leipzig, Germany}\\
\normalsize{$^\ast$To whom correspondence should be addressed; E-mail:  patrick.zschech@uni-leipzig.de.}
}

\date{}

\newacronym{pbpm}{PBPM}{predictive business process monitoring}
\newacronym{bpm}{BPM}{business process management}
\newacronym{dsr}{DSR}{design science research}
\newacronym{dnn}{DNN}{deep neural network}
\newacronym{lstm}{LSTM}{long short-term memory}
\newacronym{bilstm}{BiLSTM}{bi-directional long short-term memory}
\newacronym{ml}{ML}{machine learning}
\newacronym{dl}{DL}{deep learning}
\newacronym{pm}{PM}{process mining}
\newacronym{it}{IT}{information technology}
\newacronym{is}{IS}{information systems}
\newacronym{pdp}{PDP}{partial dependence plots}
\newacronym{ebm}{EBM}{explainable boosting machine}
\newacronym{gbm}{GBM}{gradient boosting machine}

\newacronym{xai}{XAI}{explainable artificial intelligence}
\newacronym{shap}{SHAP}{Shapley additive explanations}
\newacronym{lime}{LIME}{local interpretable model-agnostic explanations}
\newacronym{gam}{GAM}{generalized additive model}
\newacronym{gaim}{GAIM}{generalized additive index model}
\newacronym{nam}{NAM}{neural additive model}
\newacronym{gannm}{GANNM}{generalized additive neural network model}
\newacronym{xnn}{xNN}{explainable neural network}
\newacronym{exnn}{ExNN}{enhanced explainable neural network}
\newacronym{ppr}{PPR}{projection pursuit regression}
\newacronym{gaminet}{GAMI-Net}{generalized additive models with structured interactions}
\newacronym{igann}{IGANN}{interpretable generalized additive neural networks}
\newacronym{vi}{VI}{variable importance}
\newacronym{ice}{ICE}{individual conditional expectation}
\newacronym{rf}{RF}{random forest}
\newacronym{dt}{DT}{decision tree}
\newacronym{svm}{SVM}{support vector machine}
\newacronym{lr}{LR}{linear/logistic regression}
\newacronym{xgb}{XGB}{extreme gradient boosting}
\newacronym{psplines}{P-Splines}{penalized B-splines}
\newacronym{tpsplines}{TP-Splines}{penalized thin plate regression splines with shrinkage}
\newacronym{mlp}{MLP}{multi-layer perceptron}
\newacronym{hpo}{HPO}{hyperparameter optimization}
\newacronym{auroc}{AUROC}{area under the receiver operating characteristic curve}

\newacronym{rmse}{RMSE}{root mean squared error}
\newacronym{mse}{MSE}{mean squared error}
\newacronym{mae}{MAE}{mean absolute error}

\begin{document} 

\baselineskip24pt


\maketitle

\begin{biseabstract}

Machine learning is permeating every conceivable domain to promote data-driven decision support. The focus is often on advanced black-box models due to their assumed performance advantages, whereas interpretable models are often associated with inferior predictive qualities. More recently, however, a new generation of generalized additive models (GAMs) has been proposed that offer promising properties for capturing complex, non-linear patterns while remaining fully interpretable. To uncover the merits and limitations of these models, this study examines the predictive performance of seven different GAMs in comparison to seven commonly used machine learning models based on a collection of twenty tabular benchmark datasets. To ensure a fair and robust model comparison, an extensive hyperparameter search combined with cross-validation was performed, resulting in 68,500 model runs. In addition, this study qualitatively examines the visual output of the models to assess their level of interpretability. Based on these results, the paper dispels the misconception that only black-box models can achieve high accuracy by demonstrating that there is no strict trade-off between predictive performance and model interpretability for tabular data. Furthermore, the paper discusses the importance of GAMs as powerful interpretable models for the field of information systems and derives implications for future work from a socio-technical perspective. 
\end{biseabstract}

\begin{bisekeywords}
Decision support, Predictive analytics, Interpretable machine learning, Generalized additive models, Explainable artificial intelligence
\end{bisekeywords}


\newpage

\section{Introduction}
\label{sec:intro}
\Gls{ml} has made significant advancements in recent years, allowing for the automation of many tasks related to predictive decision-making \citep{janiesch_machine_2021}. Promising examples can be found in various fields such as e-commerce \citep{ghavamipoor_reinforcement_2020}, transportation \citep{balster_eta_2020}, industrial maintenance \citep{landwehr_design_2022, zschech_prognostic_2019}, and business process monitoring \citep{oberdorf_predictive_2022, kratsch_machine_2021}. However, many advanced \gls{ml} models, such as boosted decision trees and deep neural networks, suffer from black-box characteristics. This means that the mathematical functions they learn between input features and target values are so complex that it is practically impossible for humans to understand how these models generate a prediction for a given data instance \citep{bauer_explain_2021}. This lack of transparency can lead to a lack of trust in using these models for high-stakes decision tasks, such as in healthcare, finance, and criminal justice \citep{rudin2019stop, thiebes_trustworthy_2021}.

To address the lack of transparency, interpretable \gls{ml} models have been developed \citep{barredo_arrieta_explainable_2020}. The structure of interpretable models is constrained in some way to provide a better understanding of how predictions are generated \citep{rudin2019stop}. Traditional representatives are linear models and decision trees, which are easy to analyze and comprehend, but often too limited to capture more complex relationships. As a remedy, more advanced models have been proposed to mitigate the trade-off between predictive performance and interpretability. Of particular interest are \textit{\glspl{gam}}, which are currently experiencing a renaissance in the debate about comprehensible decision support \citep{nori2019interpretml, barredo_arrieta_explainable_2020, zschech_game_2022}.
 
In general, \Glspl{gam} are a type of \gls{ml} model that allows for the estimation of non-linear relationships between predictor variables (i.e., features) and a response variable (i.e., target). They are an extension of linear models, in which the linearity assumption is relaxed, allowing for more flexible and powerful modeling capabilities \citep{hastie_generalized_1986}. More specifically, the relationship between each feature and the target is modeled separately in so-called \textit{shape functions}, and the results are combined in an additive manner. This allows the model to capture arbitrary relationships while remaining fully interpretable, which provides crucial benefits for model analysis and debugging purposes \citep{lou_intelligible_2012}.

In recent years, multiple versions of \glspl{gam} have been proposed. Although they all follow the same principle of using non-linear functions to map input features to the target space, they integrate different algorithmic approaches. For example, while the original \gls{gam} uses splines for fitting shape functions \citep{hastie_generalized_1986, wood_generalized_2017}, more recent variants are based on trees \citep{lou_intelligible_2012, caruana2015intelligible} or tailored neural networks \citep{agarwal2021neural, yang_gami-net_2021, kraus_interpretable_2023}. Taken together, these different extensions show promising directions for increasing model performance while remaining fully transparent. As such, they constitute pivotal instruments in the area of interpretable \gls{ml} and comprehensible decision support, which is of central importance for the \gls{is} community.

Despite their promising potential, there have been only a few evaluation studies to empirically assess the merits and limitations of different \gls{gam} variants. Each model has already been tested in individual settings and domains, especially by the respective authors and developers. However, it currently lacks an overarching cross-model comparison that considers the different particularities. Likewise, it lacks a thorough analysis that examines the performance gap between these interpretable models and commonly applied black-box models. We believe that such findings are essential to advancing and promoting the use of transparent \gls{ml} models in research and industry alike.

For this reason, we conduct a comprehensive evaluation study in which we analyze a series of modern \gls{gam} variants and compare them with traditional \gls{ml} models based on a large collection of tabular datasets. In particular, this work contributes to the existing body of knowledge in the following ways:

\begin{itemize}
\setlength\itemsep{-1mm}
\vspace{-2.5mm}
    \item We assess and compare the predictive performance of seven different \glspl{gam} as well as seven commonly used \gls{ml} models based on a collection of twenty benchmark datasets.
    \item We conduct an extensive hyperparameter search and perform a cross-validation to ensure a fair and robust model comparison, resulting in more than $68,500$ training runs across all datasets and models.
    \item We examine the visual outputs of all seven \glspl{gam}, verify their intrinsic interpretability, assess similarities and differences between them, and evaluate their level of interpretability based on six evaluation criteria.
    \item We integrate all models and experiments into a unified evaluation pipeline and provide a publicly available repository that is free to use for researchers and practitioners.
    \item We summarize our results and dispel the misconception that only black-box models can achieve high accuracy by demonstrating that there is no strict trade-off between predictive performance and model interpretability for tabular data.
    \item We discuss the importance of \glspl{gam} as a powerful interpretable model for the \gls{is} community and the broader field of decision support by outlining their overall suitability for high-stakes decision tasks and derive implications for further research from a socio-technical perspective.
\end{itemize}

The remaining paper is structured as follows: Section~\ref{sec:background} provides an overview of relevant foundations and related work. Subsequently, Section~\ref{sec:method} describes the experimental set-up of our evaluation study, followed by the presentation of the results in Section~\ref{sec:results}. Section~\ref{sec:discussion} offers a discussion of our findings, outlines the potentials of advanced \glspl{gam} for the \gls{is} community, and highlights promising avenues for future research. Section~\ref{sec:conclusion} concludes the paper with summarizing thoughts.

\section{Conceptual Background and Related Work}
\label{sec:background}

\subsection{Model Interpretability}

The need for model interpretability has grown substantially due to the widespread use of \gls{ml} models, particularly in critical domains like credit scoring, healthcare, and criminal justice \citep{rudin_why_2019, bauer_explain_2021}. In general, model interpretability refers to a model's ability to explain or present its decision logic in a human-understandable way \citep{du_techniques_2019, doshi2017towards}. This can take various forms depending on the specific use case and stakeholder requirements. What is enough for one use case may not be enough for another, and different stakeholders may have distinct expectations and requirements for interpretability. To this end, \citet{doshi2017towards} distinguish between application-grounded, human-grounded, and functionally grounded interpretability. While the first two involve user-centric evaluation criteria, such as perceived complexity or required cognitive effort, the latter refers to a formal definition of interpretability, allowing for a more objective assessment of interpretability. In this paper, we focus explicitly on the \textit{functionally grounded} perspective of interpretability and consider the extent to which a model is able to visually reveal how model inputs (i.e., features) are processed in order to produce a predicted outcome (i.e.,~target).

Apart from that, it is possible to distinguish between post-hoc interpretability and intrinsic interpretability \citep{rudin2019stop, du_techniques_2019}. \emph{Post-hoc interpretability} requires the construction of a second model to explain the behavior of an existing model that usually exhibits a high degree of complexity, such as boosted decision trees or deep neural networks. Common post-hoc techniques include \gls{shap} \citep{lundberg_local_2020} and \gls{lime} \citep{ribeiro_why_2016}. They are often also subsumed under the term \gls{xai} as they provide explanations that attempt to simplify a complex mathematical function so that it is digestible by a human user \citep{barredo_arrieta_explainable_2020}. Although post-hoc explanations can lead to valuable insights \citep[e.g.,][]{senoner_using_2022}, they must be considered with caution as they might not be reliable and can result in misleading conclusions \citep{rudin2019stop, babic_beware_2021}.

By contrast, \emph{intrinsic interpretability} is achieved by constructing models that are interpretable by design \citep{rudin2019stop}. Such models provide not just an approximate explanation of a model's functional behavior, but an \emph{exact} description of how a model computes a prediction. That is, the underlying mathematical function (which fully defines the model) is simple enough so that users can access and analyze it directly. To achieve this goal, the models are restricted in their structure by incorporating practical constraints, such as linearity, monotonicity, additivity, and sparsity \citep{rudin_why_2019, sudjianto_unwrapping_2020}. Common representatives of intrinsically interpretable models are linear models, point systems, and decision trees, which have often been favored in the past in high-stakes decisions due to their transparency and ease of understanding \citep{lou_intelligible_2012, rudin_why_2019, yang_gami-net_2021}. At the same time, however, such simple models often lack the flexibility to capture more complex relationships, which naturally occur in real-world applications.

\subsection{Generalized Additive Models}

In this study, we focus on the specific model family of \textit{\glspl{gam}} as a particularly powerful class of intrinsically interpretable \gls{ml} models that balance transparency with high accuracy. To achieve this goal, \glspl{gam} build relationships between input features and the target by summing up several distinct non-linear mappings, known as \textit{shape functions}\footnote{Alternatively, some works use the term \textit{smooth function} \citep{hastie_generalized_1986} to emphasize the property that the function is infinitely differentiable and results in a smooth graph.} \citep{lou_intelligible_2012}. As such, they combine the simplicity of linear models with the flexibility of non-linear mappings by replacing static model coefficients with flexible shape functions to capture more complex relationships between input features and the prediction target. 

More formally, this model structure can be expressed as follows. Let $D=\left(X, y\right)$ denote a training dataset, where $X=(x_{1},\dots ,x_{n}) \in \mathbb{R}^{N \times n}$ is the feature matrix comprising $N$ samples and $n$ features and ${y}$ denotes the target variable. For a regression task, ${y}$ is a real value ($y\in\mathbb{R}^N$), while for a binary classification task, ${y}$ is a binary value ($y \in \{1,0\}^N$). A generalized additive model is then defined in the following form by taking a feature matrix $X$ as input and calculating a prediction $\hat{y}$ as output
\newcommand{\R}{\mathbb{R}}
\begin{equation}
    g(\hat{y})=f_1(x_1)+ \cdots + f_n(x_n),
    \label{eq:1}
\end{equation}
where $g(\cdot)$ is called link function\footnote{If the link function is the identity function, Equation~\ref{eq:1} describes a regression model, and if the link function is the logistic function, Equation~\ref{eq:1} describes a classification model.} and $f_i(\cdot)$ is the shape function for a feature $x_i$ \citep{lou_intelligible_2012}. 

By employing separate functions for each feature, the overall structure of \glspl{gam} remains simply interpretable, as it allows model users and developers to verify the importance of each feature. In other words, it is directly observable how each feature, through its corresponding shape function, affects the predicted output.
For the examination of the individual feature effects, the corresponding shape functions can be visualized using two-dimensional shape plots, in which the $x$-axis represents the feature values and the $y$-axis represents the impact on the predicted output. On this basis, human experts can evaluate the established relationships and verify them with their domain knowledge.

\begin{figure}[h!]
\centering
\includegraphics[width=0.65\textwidth]{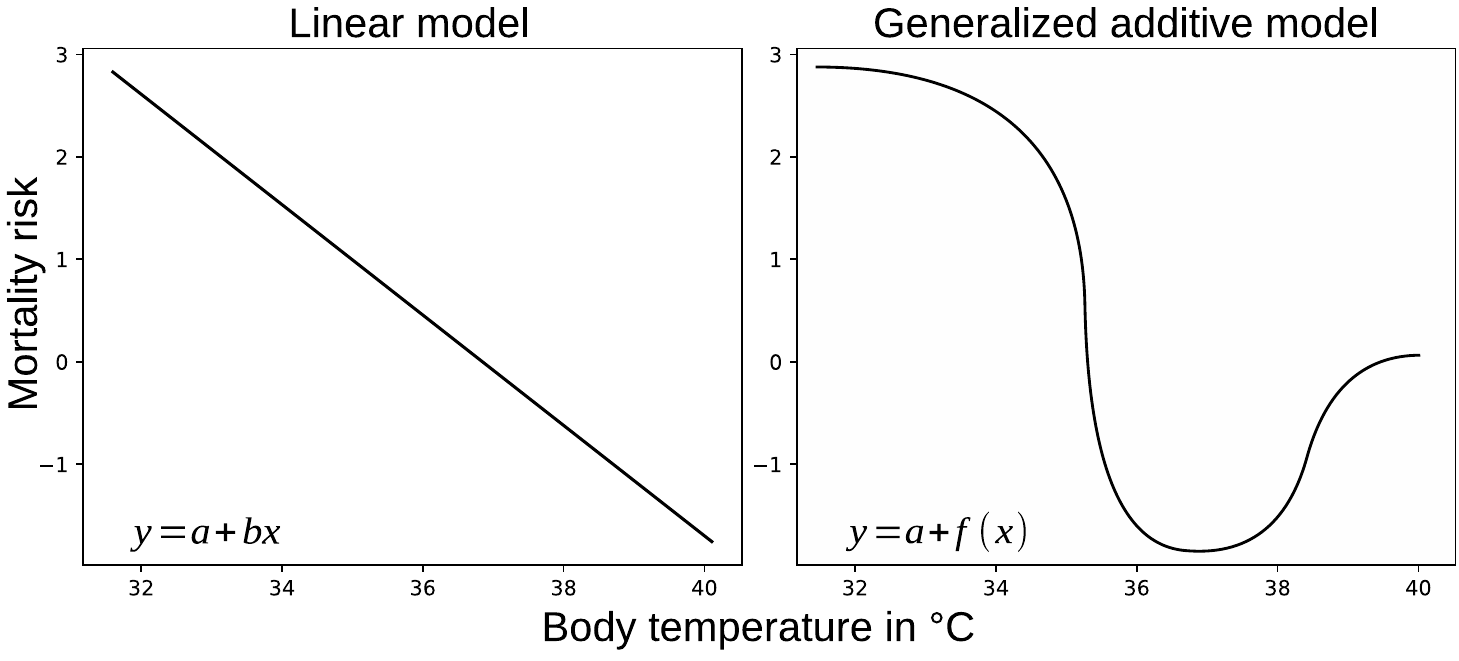}
\caption{Comparison of shape functions of a linear model and a \gls{gam}.}
\label{fig:gam_vs_lr}
\end{figure}

Figure~\ref{fig:gam_vs_lr} on the right shows an example of a shape plot where a \gls{gam} captures the non-linear relationship between body temperature and the probability (in log odds) of mortality. In contrast, on the left, a linear model is visualized which assumes a linear relationship between body temperature and the probability of mortality. This comparison highlights the flexibility of \glspl{gam} in modeling complex patterns, providing more accurate and interpretable predictions as opposed to linear alternatives.

\subsection{Variants of Generalized Additive Models}

The original \gls{gam} refers to a model in which the shape functions represent \textit{regression splines} \citep{hastie_generalized_1986}. Splines are a class of mathematical functions defined over intervals, typically composed of piecewise segments joined at specific points known as knots. They serve as flexible tools for approximating complex relationships in data, employing a variety of functions, such as polynomial functions or radial basis functions, to achieve smoothness and high flexibility in modeling. To address unnecessary model complexity and avoid overfitting, spline-based \glspl{gam} can also be enriched with advanced techniques such as shrinkage and automated smoothness estimation techniques to encourage smoother and more parsimonious models \citep{wood_generalized_2017}. Moreover, some authors integrate spline-based \glspl{gam} into ensemble learning strategies in order to further enhance their predictive performance \citep[e.g.,][]{de_bock_ensemble_2010}.

A second stream of research focuses on tree-based learning approaches for fitting shape functions in \glspl{gam} \citep{ caruana2015intelligible}. To this end, \citet{lou_intelligible_2012} argue that splines are often too smooth for real-world applications and that higher predictive performance is achievable by using tree-based step functions. More specifically, the authors suggested using bagged and boosted tree ensembles to increase model flexibility and to fit more detailed shape functions. The authors further extended their approach by incorporating pairwise interaction terms, which can still be visualized as two-dimensional heatmaps \citep{lou_accurate_2013}. Subsequently, the authors made their algorithm publicly available as an easy-to-use implementation known as \textit{\gls{ebm}} \citep{nori2019interpretml}.

Another stream of research has shifted the focus to the development of \gls{gam} extensions inspired by tailored artificial neural networks. For example, \cite{agarwal2021neural} proposed \textit{\gls{nam}} in which feature-wise shape functions are learned via deep neural networks consisting of multiple hidden layers. To fit the rapid changes in the target variables of real-world data, the authors introduce exp-centered hidden units. These units allow to capture sharp changes in the output.


\citet{yang_gami-net_2021} proposed \textit{\gls{gaminet}}. The basic architecture is similar to that of a \gls{nam}, using a simple \gls{gam}-based structure in the form of a disentangled feed-forward network with multiple additive sub-networks. To reduce unnecessary model complexity and avoid overfitting, \gls{gaminet} integrates sparsity constraints to select the most relevant features to receive a compact model. Furthermore, \gls{gaminet} has a similar strength to \gls{ebm} in that it is able to incorporate pairwise interactions between individual features, which can lead to better predictive performance. At the same time, this functionality requires further model constraints such as heredity and marginal clarity constraints to retain structural interpretability and avoid mutual absorption between main effects and pairwise interactions.

A different architectural design was sought with the proposal of \textit{\gls{xnn}} \citep{vaughan_explainable_2018} and its successor \textit{\gls{exnn}} \citep{yang_enhancing_2021}. Similar to \gls{gaminet}, both models are based on multiple sub-networks and regularization terms to retain sparsity and receive a compact model. However, instead of using a simple \gls{gam} structure, both models are based on the structure of additive index models by adopting the idea of a projection pursuit regression \citep{friedman_projection_1981}. This structure generally violates the principle of univariate feature mappings due to an additional projection layer that fully connects all input features to the following sub-networks. As a result, each feature can possibly have a partial contribution to all corresponding shape functions, to achieve a higher level of predictive performance.

More recently, \citet{kraus_interpretable_2023} proposed another novel \gls{gam} variant, called \textit{\gls{igann}}. In the first step, the model initializes shape functions linearly and then incrementally adapts to potential non-linearities. That is, \gls{igann} is based on the concept of linear modeling and only resorts to non-linearities if the underlying data requires it. For the incremental adaptation to non-linear shapes, \gls{igann} incorporates the principle of gradient boosting. Specifically, it uses a boosted ensemble of tailored sparsified neural networks, where each network represents an extreme learning machine and acts as a weak learner to gradually improve the performance.

\subsection{Comparative Evaluation Studies}

All of the \gls{gam}-based models presented above offer innovative algorithmic concepts and components to balance model accuracy and transparency. While the main principle remains almost identical for all models, some of the approaches differ greatly in their incorporated learning principles and model constraints. Therefore, it is worthwhile to evaluate and compare their merits and limitations for different prediction tasks and datasets. 

A few authors have conducted experiments to compare their proposed \gls{gam} extensions to traditional \gls{ml} baselines and competing models. For instance, \citet{lou_intelligible_2012} compared spline-based models with tree-based \glspl{gam} as well as \gls{lr} and \gls{rf} as lower and upper bound baselines. Similarly, \citet{yang_gami-net_2021} evaluated \gls{gaminet} against several benchmark models, including \gls{ebm}, splines, \gls{lr}, \gls{rf}, \gls{xgb}, and \gls{mlp}, using a large number of datasets. \citet{agarwal2021neural} benchmarked \gls{nam} against \gls{ebm} and traditional approaches, albeit with a limited number of datasets.

Beyond that, only a few studies have assessed the characteristics of different intrinsically interpretable models so far. For instance, \citet{chang_how_2021} examined a series of \glspl{gam}, such as spline-based models like penalized cubic regression splines and tree-based models like \gls{ebm}, using benchmark datasets from real-world applications and simulations. They evaluated the models both qualitatively and quantitatively. \citet{hohman_gamut_2019} incorporated \glspl{gam} into a visual analytics tool to examine how data scientists interact with shape functions. Meanwhile, \citet{kaur_interpreting_2020} explored how data science experts use and evaluate interpretable models, including both intrinsically and post-hoc interpretable models like \gls{shap} to compare the results of both paradigms.

In summary, when considering the focus of related work, it currently lacks a cross-model comparison to evaluate the merits and limitations of different \glspl{gam} from a broader and more neutral perspective. Likewise, it lacks a thorough analysis that examines the performance gap between these advanced interpretable models and commonly applied black-box models. Especially for application-oriented sciences such as the IS community, it is crucial to understand how different interpretable models behave in direct comparison as well as how they perform in comparison to commonly used \gls{ml} models. We narrow this gap by contributing new insights through a comprehensive comparative evaluation study based on a large collection of datasets and models.

\section{Research Approach}
\label{sec:method}

We perform a series of computational experiments, which we describe in the following. In the first part, we focus on the assessment of the predictive performance, for which we outline the selection of evaluated models~(Section~\ref{sec:method_models}), our collection of datasets~(Section~\ref{sec:method_data}), and our experimental set-up and evaluation pipeline~(Section~\ref{sec:method_setup}). 
Thereafter, we consider the models' visual outputs to assess their level of functional interpretability (Section~\ref{sec:method_interpret}). 

\subsection{Selection of Models}
\label{sec:method_models}

For our comparative evaluation study, we consider seven different variants of \glspl{gam}, for which publicly accessible implementations are available. This includes (i)~\gls{psplines} \citep{eilers_flexible_1996}, (ii)~\gls{tpsplines} \citep{wood_thin_2003}, (iii)~\gls{ebm} \citep{nori_interpretml_2021}, (iv)~\gls{nam} \citep{agarwal2021neural}, (v)~\gls{gaminet}~\citep{yang_gami-net_2021}, (vi)~\gls{exnn} \citep{yang_enhancing_2021}, and (vii)~\gls{igann} \citep{kraus_interpretable_2023}.
The implementations are provided by the respective authors of the proposed \gls{gam} variants. For \gls{psplines}, we use the Python package pyGAM \citep{serven_pygam_2021}. For \gls{tpsplines}, we adopt models from the mgcv package implemented in R \citep{wood_mgcv_2023} using a Python wrapper \citep{chang_how_2021}. Additionally, we also experiment with other spline-based models from the mgcv package, including penalized cubic regression splines with and without shrinkage and different configurations. However, since they only achieve subordinate predictive performance compared to \gls{tpsplines}, they are excluded from further model comparison (see Appendix~\ref{app:additional_experiments}). 

Furthermore, we include several baseline models for a broader comparison, which are commonly used in the \gls{ml} and \gls{is} community for prediction tasks using tabular data. Specifically, we include linear models (logistic/linear regression, LR) and \glspl{dt} as common representatives of intrinsically interpretable models. To consider widely known black-box models, we include \gls{rf}, \gls{xgb}, CatBoost, \gls{mlp}, and TabNet \citep{de_caigny_hybrid_2024, schoormann_artificial_2023}. For the implementation of the models, we use the corresponding Python packages from the scikit-learn library, except for \gls{xgb}~\citep{chen_xgboost_2016}, CatBoost~\citep{prokhorenkova_catboost_2018}, and TabNet \citep{arik_tabnet_2021}, where we use Python implementations from the respective developers/authors. Further details on the model implementations can be found in Appendix~\ref{app:models}.

\subsection{Selection of Datasets}
\label{sec:method_data}

For an extensive comparison of models, we utilize a variety of benchmark datasets. We align our dataset selection with previous evaluation studies \citep[e.g.,][]{roy_performance_2019, yang_gami-net_2021}, using datasets from recognized public repositories such as the UCI Machine Learning Repository\footnote{http://archive.ics.uci.edu/ml/} and Kaggle\footnote{https://www.kaggle.com/}.
These repositories offer diverse, high-quality datasets for benchmarking purposes, allowing for a thorough evaluation of the different models.
After reviewing the repositories, we select twenty diverse datasets that cover a broad range of real-world applications relevant to various business, organizational, and societal issues. To ensure a balanced distribution between different prediction tasks, we use ten datasets tailored for predicting categorical outcomes (binary classification task, CLS) and an equal number for predicting numerical values (regression task, REG). 
Moreover, we limit our experiments to medium-sized datasets with up to $150,000$ samples to maintain a manageable level of computational complexity. This is particularly important given the extensive hyperparameter tuning in our model evaluation procedure~(cf. Section~\ref{sec:method_setup}).

Table~\ref{tab:datasets_overview} provides an overview of the overall collection with summary statistics of the inherent dataset properties (after cleaning and data preprocessing) and a short description of the prediction targets. The number of samples ranges from $205$ to $142,193$, and the number of preprocessed and encoded predictors varies between $8$ and $100$ with a mixed combination of numerical and categorical features. Thus, we consider a variety of benchmark scenarios for assessing and comparing all models. Further details on the data repositories and the applied preprocessing steps can be found in Appendix~\ref{app:datasets}.

\begin{table}[ht]
\caption{Overview of selected datasets covering classification (CLS) and regression (REG) tasks.}
\label{tab:datasets_overview}
\sisetup{group-digits=true, group-separator={,}, group-minimum-digits=4}
\resizebox{\textwidth}{!}{
\begin{tabular}{llS[table-format=6.0]rrrl}
\toprule
\multicolumn{1}{l}{\textbf{Type}} & \multicolumn{1}{l}{\textbf{Dataset}}& \multicolumn{1}{l}{\textbf{Samples}} & \multicolumn{3}{l}{\textbf{Features}} & \multicolumn{1}{l}{\textbf{Prediction Target}} \\ \cmidrule(l){4-6} &  & & \textbf{num*} & \textbf{cat*} & \textbf{cat**} & \\\midrule 
\multirow{10}{*}{CLS} & College  \citep{college_source} & 1000 & 4 & 6 & 10 & Will a high school student go to college?\\
 & Water potability \citep{water_source} & 3276 & 9 & 0 & 0 & Will the water be safe for consumption?\\
 & Stroke \citep{stroke_source} & 5110 & 3 & 7 & 16 & Will a patient suffer from a stroke?\\
 & Customer churn \citep{IBM} & 7043 & 3 & 16 & 37 & Will a customer leave the company?\\
 & Recidivism \citep{angin_machine_2016} & 7214 & 7 & 5 & 11 & Will a defendant recidivate?\\
 & Credit scoring \citep{fico_source} & 10459 & 21 & 2 & 16 & Will a client repay within 2 years?\\
 & Income adults \citep{Kohavi.1996} & 32561 & 6 & 8 & 59 & Will the income exceed \$50,000 per year?\\
 & Bank marketing \citep{moro2014-driven_2014}& 45211 & 6 & 9 & 41 & Will a client subscribe to a deposit?\\
 & Airline satisfaction \citep{tj_klein_airline} & 103904 & 18 & 4 & 6 & Will a passenger be satisfied?\\
 & Weather forecast \citep{weather_australia} & 142193 & 16 & 5 & 54 & Will it rain the next day in Australia?\\ \\
\multirow{10}{*}{REG} & Car price \citep{kibler_instance-based_1989} & 205 & 13 & 11 & 63 & What is the price of a car?\\
 & Student grade \citep{cortez_using_2008} & 649 & 13 & 17 & 30 & What is a student's final grade?\\
 & Productivity \citep{Imran_2019}& 1197 & 9 & 4 & 26 & What is the productivity of a team?\\
 & Medical insurance \citep{lantz_2015}& 1338 & 3 & 3 & 6 & What are the costs for a patient?\\
 & Violent crimes \citep{redmond_data-driven_2002} & 1994 & 100 & 0 & 0 &How many violent crimes will happen?\\
 & Crab farming \citep{gursewak_singh_sidhu_2021} & 3893 & 7 & 1 & 3 & What is the age of a crab?\\
 & Wine quality \citep{cortez_2009}& 4898 & 11 & 0 & 0 & What is the quality of a produced wine? \\
 & Bike rental \citep{fanaee-t_event_2014} & 17379 & 7 & 5 & 5 & How many bikes will be rented per hour?\\
 & House price \citep{pace_sparse_1997}& 20640 & 8 & 0 & 0 & What is the value of a house?\\
 & Diamond price \citep{garside_2021} & 53943 & 6 & 3 & 20 & What is the price of a diamond?\\ \bottomrule
 \multicolumn{7}{l}{\textit{Note:} * after data preprocessing; ** after feature encoding}
 \end{tabular}
 }
\end{table}

\subsection{Experimental Set-up and Evaluation Pipeline}
\label{sec:method_setup}

To provide a fair comparison, we integrate all datasets and models into a shared environment to run the experiments under the same conditions.\footnote{In our experiments, we use a workstation with the following setting: Single GPU NVIDIA RTX A 6000 with 48GB VRAM, Intel i7-12700 10 CPU cores with 20 threads, and 128 GB RAM, Python 3.9.13, Pytorch 1.12.1, and cudatoolkit 11.6.0} 
The predictive performance is measured in terms of commonly used evaluation metrics. For classification tasks, we primarily focus on the \gls{auroc} to assess the models' ability to correctly rank positive and negative instances. For regression tasks, we consider the \gls{rmse} to quantify the average magnitude of errors between the models' predicted values and the actual values. To determine the best-performing models, we calculate the average ranks across all datasets, with higher ranks (i.e., first, second, third, etc.) indicating better predictive performance. The procedure of ranking algorithms aligns with good practice for evaluating multiple classifiers on multiple datasets \citep{demsar_statistical_2006, brazdil2000comparison}.

For the sake of completeness, we also calculate additional metrics that can be found in the online repository, which is referred to at the end of this section.\footnote{For classification tasks, we additionally measure accuracy, precision, recall, F1-score (macro average, weighted average), and support. For regression tasks, we additionally measure \gls{mse}, \gls{mae}, explained variance, $r^2$, and maximum residual error (max error).} Moreover, we measure the training time (in seconds) of all models to assess their computational intensity. The results of this examination are reported and discussed in Appendix~\ref{app:training_times}.

Furthermore, all models are evaluated in two settings. In the first setting, we keep all models in their default hyperparameter configuration. In the second setting, we repeat our evaluation using a grid search tuning our models' hyperparameters out-of-sample. For this purpose, we select the main hyperparameters and establish a grid of parameter values that tend to have a high impact on model performance.

\begin{figure}[htp]
    \centering
    \includegraphics[width=\textwidth]{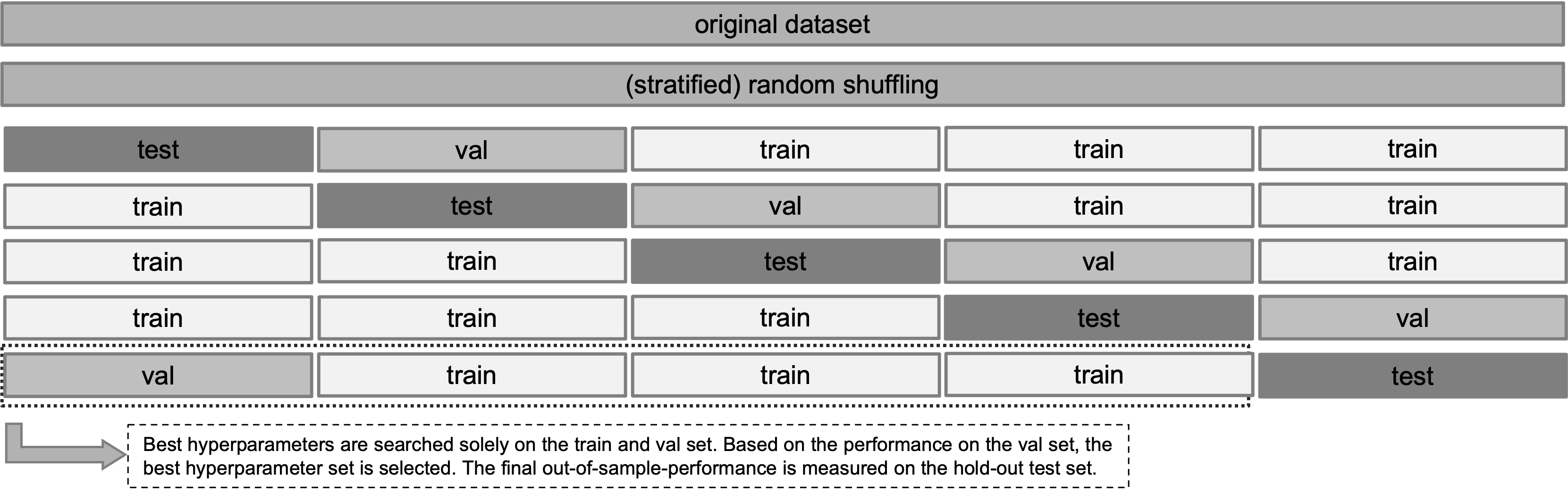}
    \caption{Illustration of our evaluation strategy using 5-fold (stratified) cross-validation for error estimation and inner train-validation splitting for hyperparameter tuning with final retraining in each iteration.}
    \label{fig:hpo_pipeline}
\end{figure}

To mitigate the effects of random outliers and to assess the stability and robustness of all models, we apply a well-defined evaluation strategy.
We illustrate the strategy in a simplified way in Figure~\ref{fig:hpo_pipeline}.
Specifically, we use a 5-fold (stratified) cross-validation with random shuffling, which allows us to perform error estimation by measuring the out-of-sample performance on each test fold and then calculating the mean and standard deviation across all folds. 
In addition, for the hyperparameter-tuned setting, we apply an inner-split-validation to tune all hyperparameters. 
During the random shuffling strategy used in both the cross-validation and the inner-split-validation, seed values are set to ensure reproducible results.
Our designed approach is a mild variation of well-established pipelines \citep{de_caigny_new_2018, kuhl_how_2021}, with modifications justified by the need for comparability and out-of-sample hyperparameter tuning.
Overall, this results in $68,500$ model runs across all datasets and models, with a total training time exceeding $279$ computing hours. Detailed information on the choice of hyperparameters and the best hyperparameter configurations for each dataset and model can be found in Appendix~\ref{app:hyperparameter}.

To ensure the reproducibility of our experiments, but also to promote the reusability of our results, we incorporate all models, datasets, dataset-agnostic preprocessing steps, metrics, and evaluation procedures into a unified evaluation pipeline, which is publicly available in an online repository\footnote{
\url{https://github.com/NicoHambauer/Model-Performance-vs-Interpretability}}. The evaluation pipeline is freely available to researchers and practitioners, and can be easily extended for individual needs (e.g., new models, custom datasets, additional metrics, etc.).

\subsection{Assessment of Interpretability}
\label{sec:method_interpret}

To comprehensively analyze the behavior of all models and assess their level of functional interpretability, our investigations are divided into three parts, offering different perspectives.

In the \textbf{first part}, we consider each model individually and examine the \glspl{gam}' general ability to externalize their inner workings through human-understandable feature plots to verify their degree of intrinsic interpretability. More specifically, we follow the notion that an intrinsically interpretable model must be able to provide transparency at the level of individual predictions (i.e., \textit{local interpretability}), but also at the level of the entire model (i.e., \textit{global interpretability}), in order to enable a full understanding of how a model works \citep{du_techniques_2019}. To this end, we look into the \glspl{gam}' entire output as well as specific shape functions and assess the extent to which each model is able to visually reveal how input features are processed in order to produce a predicted outcome.

In the \textbf{second part}, we contrast and compare the behavior of the different models by looking at them simultaneously. More specifically, we qualitatively examine the \glspl{gam}' different feature shape plots across all 20 datasets and models. To ensure a fair and consistent evaluation of the feature plots, we develop custom plot functions for each model and plot type (e.g., for numerical, categorical, and interaction feature effects). On this basis, we identify similarities and differences that allow us to make informed statements about their merits and limitations for different interpretability contexts.

Finally, in the \textbf{third part}, we build on our previous findings and evaluate all models using an objective, aggregated interpretability score. For this purpose, we adopt the six interpretability criteria (see Table~\ref{tab:interpret_metrics}) suggested by \citet{Sudjianto2021DesigningII}, which allow us to reflect on the strengths and weaknesses of each model. \citet{Sudjianto2021DesigningII} propose these six criteria as part of a qualitative assessment framework to determine the level of model-based interpretability in the context of tabular data. 
For our study, we make some adjustments to the framework to avoid ambiguous evaluations. On the one hand, we adjust the scoring system by simplifying the scales from four levels to three levels. Thus, a model receives two points if it meets a given criterion by default, one point if it meets the criterion after adjusting related hyperparameters, or zero points if a model cannot meet the criterion.
On the other hand, we also refine the evaluation criteria in order to increase their conceptual sharpness and to achieve a better common understanding among all authors of this study. For example, the \textit{visualizability} criterion is refined to more clearly emphasize a model's ability to visualize how changes in the model input affect the model output without re-evaluating the model. In Table~\ref{tab:interpret_metrics}, we summarize the refined evaluation criteria and provide representative examples for each criterion.

The actual evaluation was performed by three of the authors of this paper, who gained extensive experience with each of the models over the course of the study. To ensure an objective assessment, each author independently rated each criterion for each model. The results were then discussed and inter-rater reliability was measured using Krippendorff's alpha \citep{krippendorff2018content}, with an initial agreement of 0.7476. The disagreements were mainly due to different interpretations of the original definitions of the evaluation criteria, which led to the aforementioned refinements. After the refinements, a second round of independent evaluations was performed, resulting in an agreement of 0.9254. The remaining disagreements were due to uncertainties about the effects of certain model hyperparameters to assess whether or not certain interpretability criteria could be met by the corresponding models. To resolve these issues, additional model experiments were performed to reach a final consensus for a correct assessment. Further details on the scoring process, discussion points that arose during the evaluation, and justifications for final scoring decisions can be found in Appendix~\ref{app:interpret_criteria}.

\begin{table}[htbp!]
\caption{Evaluation criteria for interpretability assessment.}
\label{tab:interpret_metrics}
\scriptsize
\begin{tabular}{p{1.8cm}p{7.0cm}p{5.8cm}}
     \toprule
     \textbf{Criteria} & \textbf{Descriptions} & \textbf{Examples} \\
     \midrule
     \emph{Additivity} & Whether the model allows for additive decomposition and modular aggregation of individual features so that each feature effect can be analyzed separately. & Linear models and \glspl{gam} capture features separately and then sum up the feature effects to obtain the overall prediction value. \\[8mm] 
     \emph{Sparsity} &  Whether the model uses an internal selection or regularization mechanism to identify a relevant subset of features, with the goal of maximizing their relevance while reducing model complexity. &  \gls{gaminet} promotes model compactness by incorporating sparsity constraints to include only non-trivial main effects.  \\[8mm]
     \emph{Linearity} &  Whether the model is able to capture linear or piecewise linear relationships between the features and the target, despite the presence of noise in the dataset. & \gls{igann} promotes linearity by initializing shape functions linearly and only deviating from this when required by the underlying features. \\[8mm]
     \emph{Smoothness} &  Whether the model avoids erratic curves and abrupt jumps by ensuring that small changes in the input lead to small changes in the output. & \gls{tpsplines}, \gls{gaminet}, and \gls{igann} promote smoothness by preserving continuous and gradually changing shape functions. \\[8mm]
     \emph{Monotonicity} &  Whether monotonic constraints can be injected into the model to maintain a steadily increasing or decreasing trend in the feature-target relationships. & \gls{ebm} and \gls{gaminet} provide functionalities or hyperparameters to impose monotonically increasing or decreasing feature effects.  \\[8mm]
     \emph{Visualizability} &  Whether changes in the model's output can be visually understood unambiguously without re-evaluating the overall model when modifying a single feature. & \gls{psplines}, \gls{tpsplines}, \gls{ebm}, \gls{igann}, and \gls{gaminet} produce comprehensible shape plots to visually inspect the impact of individual features.  \\
     \bottomrule
\end{tabular}
\end{table}

\section{Results}
\label{sec:results}

\subsection{Assessment of Predictive Performance}
\label{sec:results_pred}

This section presents the evaluation results for assessing the predictive performance. First, we consider the setting with default model configurations, and then we continue with the results of the models with tuned hyperparameters. Finally, we also consider the performance differences between both settings. The best overall performance for each dataset is highlighted in bold, whereas the best result among the interpretable models is underlined.\footnote{If multiple models achieve the same average predictive performance values, the model with the lower standard deviation is considered the better one.} Moreover, the combination of certain datasets with \gls{tpsplines} results in some anomalies where the model does not converge on all five folds, so a subset of folds is used to compute the aggregated performance results. These cases are highlighted in italics.

\begin{table}[htp]
\caption{Predictive performance using default hyperparameters. Classification tasks are assessed using \gls{auroc}, whereas regression tasks are measured using \gls{rmse}.}
\label{tab:pred_perf_default_hp}
\resizebox{1\textwidth}{!}{%
\begin{tabular}{@{}l@{\hspace{0.1cm}}lllllllllllllll@{}}
\toprule
& & \multicolumn{9}{c}{\textbf{Interpretable Models}} & \multicolumn{5}{c}{\textbf{Black-box Models}} \\ \cmidrule(l){3-11} \cmidrule(l){12-16} 
& & \multicolumn{7}{c}{\textbf{GAMs}} & \multicolumn{2}{c}{\textbf{Traditional}} & & & \\ \cmidrule(l){3-9} \cmidrule(l){10-11}
\textbf{Task} & \textbf{Dataset} & \rotatebox{45}{\textbf{P-Splines}} & \rotatebox{45}{\textbf{TP-Splines}} & \rotatebox{45}{\textbf{EBM}} & \rotatebox{45}{\textbf{NAM}} & \rotatebox{45}{\textbf{GAMI-Net}} & \rotatebox{45}{\textbf{ExNN}} & \rotatebox{45}{\textbf{IGANN}} & \rotatebox{45}{\textbf{LR}} & \rotatebox{45}{\textbf{DT}} & \rotatebox{45}{\textbf{RF}} & \rotatebox{45}{\textbf{XGB}}  & \rotatebox{45}{\textbf{CatBoost}} & \rotatebox{45}{\textbf{MLP}} & \rotatebox{45}{\textbf{TabNet}} \\ \midrule
\multicolumn{1}{l}{\multirow{25}{*}{CLS}} & college          & \msd{0.951}{0.004} & \msd{0.951}{0.006} & \msd{0.965}{0.008} & \msd{0.889}{0.025} & \msd{\underline{0.969}}{0.008} & \msd{0.752}{0.033} & \msd{0.952}{0.008} & \msd{0.935}{0.012} & \msd{0.839}{0.025} & \msd{0.964}{0.011} & \msd{0.962}{0.015}  & \msd{\textbf{0.972}}{0.007} & \msd{0.968}{0.013} & \msd{0.949}{0.013} \\[15pt]
& water            & \msd{0.561}{0.016} & \msd{0.585}{0.015} & \msd{0.678}{0.026} & \msd{0.564}{0.021} & \msd{0.679}{0.034} & \msd{\underline{0.680}}{0.025} & \msd{0.573}{0.033} & \msd{0.500}{0.016} & \msd{0.562}{0.016} & {\msd{0.686}{0.017}} & \msd{0.656}{0.014}  & \msd{\textbf{0.689}}{0.017} & \msd{0.681}{0.031} & \msd{0.636}{0.020} \\[15pt]
& stroke           & \msd{0.834}{0.022} & \msd{0.835}{0.024} & \msd{0.837}{0.017} & \msd{0.802}{0.020} & \msd{0.833}{0.023} & \msd{0.702}{0.034} & \msd{\underline{\textbf{0.838}}}{0.027} & \msd{0.837}{0.026} & \msd{0.560}{0.012} & \msd{0.800}{0.018} & \msd{0.803}{0.011} & \msd{0.806}{0.015} & \msd{0.610}{0.197}  & \msd{0.805}{0.051} \\[15pt]
& churn            & \msd{\underline{\textbf{0.850}}}{0.011} & \msd{0.849}{0.012} & \msd{0.849}{0.011} & \msd{0.827}{0.008} & \msd{0.847}{0.014} & \msd{0.774}{0.020} & \msd{0.847}{0.014} & \msd{0.845}{0.015} & \msd{0.662}{0.015} & \msd{0.821}{0.013} & \msd{0.824}{0.010} & \msd{0.837}{0.012}  & \msd{0.842}{0.016}  & \msd{0.835}{0.015}\\[15pt]
& recidivism           & \msd{0.744}{0.017} & \msd{0.744}{0.018} & \msd{\underline{\textbf{0.748}}}{0.016} & \msd{0.705}{0.026} & \msd{0.741}{0.018} & \msd{0.634}{0.014} & \msd{0.739}{0.019} & \msd{0.727}{0.017} & \msd{0.602}{0.012} & \msd{0.674}{0.015} & \msd{0.724}{0.017} & \msd{0.740}{0.015} & \msd{0.735}{0.012} & \msd{0.727}{0.016}  \\[15pt]
& credit             & \msd{0.806}{0.008} & \textit{\msd{\underline{\textbf{0.810}}}{0.010}} & \msd{0.808}{0.010} & \msd{0.791}{0.012} & \msd{0.804}{0.010} & \msd{0.783}{0.017} & \msd{0.802}{0.010} & \msd{0.797}{0.012} & \msd{0.636}{0.015} & \msd{0.797}{0.011} & \msd{0.781}{0.008}  & \msd{0.794}{0.007} & \msd{0.799}{0.010} & \msd{0.785}{0.013} \\[15pt]
& income            & \msd{0.916}{0.002} & \textit{\msd{0.916}{0.002}} & \msd{\underline{0.927}}{0.002} & \msd{0.897}{0.002} & \msd{0.914}{0.002} & \msd{0.609}{0.008} & \msd{0.915}{0.002} & \msd{0.907}{0.002} & \msd{0.748}{0.002} & \msd{0.903}{0.004} & \msd{0.927}{0.001} & \msd{\textbf{0.930}}{0.002} & \msd{0.913}{0.003}  & \msd{0.908}{0.004}  \\[15pt]
& bank             & \msd{0.780}{0.011} & \msd{0.779}{0.011} & \msd{\underline{0.797}}{0.011} & \msd{0.758}{0.013} & \msd{0.780}{0.011} & \msd{0.615}{0.009} & \msd{0.774}{0.012} & \msd{0.765}{0.009} & \msd{0.613}{0.006} & \msd{0.781}{0.011} & \msd{0.790}{0.009}  & \msd{\textbf{0.805}}{0.009} & \msd{0.782}{0.011} & \msd{0.775}{0.007} \\[15pt]
& airline          & \msd{0.980}{0.001} & \msd{0.980}{0.001} & \msd{0.989}{0.001} & \msd{0.963}{0.002} & \msd{0.985}{0.001} & \msd{\underline{0.990}}{0.002} & \msd{0.980}{0.001} & \msd{0.927}{0.002} & \msd{0.944}{0.001} & \msd{0.993}{0.000} & \msd{\textbf{0.995}}{0.000}  & \msd{\textbf{0.995}}{0.000} & \msd{0.994}{0.001}  & \msd{0.993}{0.002} \\[15pt]
& weather          & \msd{0.874}{0.003} & \msd{0.874}{0.003} & \msd{\underline{0.875}}{0.003} & \msd{0.852}{0.003} & \msd{\underline{0.875}}{0.003} & \msd{0.579}{0.003} & \multicolumn{1}{l}{\msd{0.873}{0.003}} & \msd{0.868}{0.002} & \msd{0.696}{0.005} & \msd{0.883}{0.002} & \msd{0.886}{0.003} & \msd{\textbf{0.887}}{0.002} & \msd{0.885}{0.003} & \msd{0.878}{0.003}   \\[15pt] \midrule
\multicolumn{1}{l}{\multirow{25}{*}{REG}} & car              & \msd{0.367}{0.105} & \textit{\msd{0.405}{0.113}}  & \msd{\underline{\textbf{0.256}}}{0.071} & \msd{0.715}{0.399} & \msd{0.381}{0.139} & \msd{1.712}{0.899} & \msd{0.350}{0.179} & \msd{0.470}{0.263} & \msd{0.357}{0.102} & \msd{0.304}{0.095} & \msd{0.301}{0.071}  & \msd{0.288}{0.091} & \msd{0.317}{0.076} & \msd{0.437}{0.236} \\[15pt]
& student          & \msd{0.882}{0.121} & \msd{0.863}{0.132} & \msd{0.856}{0.127} & \msd{0.954}{0.154} & \msd{0.900}{0.170} & \msd{1.682}{0.386} & \msd{\underline{0.850}}{0.116} & \msd{0.878}{0.151} & \msd{1.238}{0.135} & \msd{0.842}{0.143} & \msd{0.883}{0.120}  & \msd{\textbf{0.832}}{0.146} & \msd{0.876}{0.146} & \msd{0.921}{0.093}       \\[15pt]
& productivity     & \msd{0.796}{0.093} & \msd{0.789}{0.081} & \msd{\underline{0.723}}{0.048} & \msd{0.909}{0.054} & \msd{0.742}{0.052} & \msd{1.036}{0.060} & \msd{0.813}{0.076} & \msd{0.905}{0.043} & \msd{0.960}{0.085} & \msd{0.707}{0.065} & \msd{0.744}{0.042} & \msd{\textbf{0.698}}{0.036} & \msd{0.784}{0.055} & \msd{0.902}{0.024}      \\[15pt]
& insurance          & \msd{0.500}{0.020} & \msd{0.497}{0.019} & \msd{\underline{\textbf{0.379}}}{0.027} & \msd{0.683}{0.028} & \msd{0.383}{0.026} & \msd{0.625}{0.020} & \msd{0.495}{0.020} & \msd{0.863}{0.071} & \msd{0.523}{0.036} & \msd{0.400}{0.021} & \msd{0.435}{0.019} & \msd{0.388}{0.027} & \msd{0.397}{0.016} & \msd{0.405}{0.031}         \\[15pt]
& crimes           & \msd{0.737}{0.047} & \msd{0.586}{0.059} & \msd{0.590}{0.065} & \msd{0.774}{0.048} & \msd{0.610}{0.059} & \msd{0.684}{0.050} & \msd{\underline{0.584}}{0.054} & \msd{0.600}{0.066} & \msd{0.844}{0.072} & \msd{0.595}{0.061} & \msd{0.623}{0.048}  & \msd{\textbf{0.571}}{0.061} & \msd{0.612}{0.062} & \msd{0.642}{0.067}          \\[15pt]
& farming             & \msd{0.673}{0.042} & \msd{0.665}{0.043} & \msd{0.687}{0.043} & \msd{0.793}{0.036} & \msd{\underline{0.655}}{0.043} & \msd{0.668}{0.026} & \msd{0.668}{0.046} & \msd{0.798}{0.048} & \msd{0.931}{0.041} & \msd{0.682}{0.033} & \msd{0.725}{0.040} & \msd{0.673}{0.037} & \msd{\textbf{0.655}}{0.042} & \msd{0.677}{0.052}  \\[15pt]
& wine             & \msd{0.828}{0.046} & \msd{0.815}{0.026} & \msd{\underline{0.772}}{0.021} & \msd{0.880}{0.037} & \msd{0.798}{0.019} & \msd{0.811}{0.019} & \msd{0.813}{0.024} & \msd{0.893}{0.028} & \msd{0.966}{0.037} & \msd{\textbf{0.682}}{0.032} & \msd{0.707}{0.027} & \msd{0.729}{0.022}  & \msd{0.777}{0.019} & \msd{0.803}{0.020}        \\[15pt]
& bike             & \msd{0.553}{0.010} & \msd{0.581}{0.013} & \msd{0.315}{0.006} & \msd{0.748}{0.076} & \msd{\underline{0.306}}{0.009} & \msd{0.913}{0.018} & \msd{0.636}{0.016} & \msd{0.830}{0.020} & \msd{0.325}{0.008} & \msd{0.235}{0.008} & \msd{\textbf{0.227}}{0.006}  & \msd{0.230}{0.007} & \msd{0.255}{0.006} & \msd{0.311}{0.022}        \\[15pt]
& house          & \msd{0.524}{0.017} & \msd{0.536}{0.018} & \msd{\underline{0.445}}{0.014} & \msd{0.704}{0.029} & \msd{0.477}{0.015} & \msd{0.501}{0.015} & \msd{0.547}{0.018} & \msd{0.771}{0.009} & \msd{0.599}{0.018} & \msd{0.424}{0.014} & \msd{0.415}{0.011} &  \msd{\textbf{0.411}}{0.014} & \msd{0.465}{0.013} & \msd{0.471}{0.016}          \\[15pt]
& diamond          & \msd{0.383}{0.154} & \msd{0.634}{0.724} & \msd{0.230}{0.010} & \msd{0.413}{0.026} & \msd{0.214}{0.009} & \msd{0.980}{0.013} & \msd{0.268}{0.008} & \msd{0.471}{0.011} & \msd{\underline{0.189}}{0.008} & \msd{0.138}{0.004} & \msd{\textbf{0.138}}{0.003}  & \msd{0.139}{0.004} & \msd{0.147}{0.018} & \msd{0.148}{0.005}         \\[15pt] \midrule
\multicolumn{2}{l}{Average rank CLS} & \multicolumn{1}{c}{6.15}  & \multicolumn{1}{c}{6.00}   & \multicolumn{1}{c}{3.55}   & \multicolumn{1}{c}{11.20}   & \multicolumn{1}{c}{5.35}   & \multicolumn{1}{c}{11.50}   & \multicolumn{1}{c}{6.95}   & \multicolumn{1}{c}{9.70}  & \multicolumn{1}{c}{13.40} & \multicolumn{1}{c}{7.30}  & \multicolumn{1}{c}{6.45}  & \multicolumn{1}{c}{3.55}  & \multicolumn{1}{c}{5.70}  & \multicolumn{1}{c}{8.20}   \\ 
\multicolumn{2}{l}{Average rank REG}  & \multicolumn{1}{c}{9.20}  & \multicolumn{1}{c}{8.00}   & \multicolumn{1}{c}{4.60}   & \multicolumn{1}{c}{12.30}   & \multicolumn{1}{c}{5.90}   & \multicolumn{1}{c}{11.30}   & \multicolumn{1}{c}{7.30}   & \multicolumn{1}{c}{11.50} & \multicolumn{1}{c}{11.20} & \multicolumn{1}{c}{3.60}  & \multicolumn{1}{c}{5.00}  & \multicolumn{1}{c}{2.30}  & \multicolumn{1}{c}{4.80}  & \multicolumn{1}{c}{8.00}   \\ \midrule
\multicolumn{2}{l}{Average rank total}  &  \multicolumn{1}{c}{7.68}  & \multicolumn{1}{c}{7.00}   & \multicolumn{1}{c}{4.08}   & \multicolumn{1}{c}{11.75}   & \multicolumn{1}{c}{5.63}   & \multicolumn{1}{c}{11.40}   & \multicolumn{1}{c}{7.13}   & \multicolumn{1}{c}{10.60} & \multicolumn{1}{c}{12.30} & \multicolumn{1}{c}{5.45}  & \multicolumn{1}{c}{5.73}  & \multicolumn{1}{c}{2.93}  & \multicolumn{1}{c}{5.25}  & \multicolumn{1}{c}{8.10}  
  \\ \bottomrule
\end{tabular}%
}
\end{table}

\noindent \textbf{Default setting.} Table~\ref{tab:pred_perf_default_hp} summarizes the prediction results of the default setting for the classification and regression tasks. The results show that the best prediction scores for each dataset are spread across a variety of models. That is, there is no single model that achieves the best results across all datasets. However, there are some remarkable tendencies. In particular, it can be seen that the intrinsically interpretable \glspl{gam} collectively deliver the highest performance in 6 out of 20 datasets. Specifically for classification, this is the case for 4 out of 10 datasets, and for regression, this is the case for 2 out of 10 datasets.

In contrast, the black-box models perform best in 14 out of 20 datasets. Nevertheless, the performance difference between the best black-box models and the best interpretable models is only marginal. For example, for the classification tasks, the smallest difference in performance between the best interpretable model and the best black-box model is 0.003 (college: \gls{gaminet} 0.969 vs. CatBoost 0.972; income: \gls{ebm} 0.927 vs. CatBoost 0.930) and the largest difference is 0.012 (weather: \gls{ebm} 0.875 vs. CatBoost 0.887). Similarly, for the regression tasks, the smallest difference is almost 0 (farming: \gls{gaminet} 0.655 $\pm$ 0.043 vs. MLP 0.655 $\pm$ 0.042), whereas the largest difference is 0.090 (wine: \gls{ebm} 0.772 vs. \gls{rf} 0.682). These results clearly outline that there is no remarkable loss of model accuracy in favor of interpretability.

A more detailed analysis reveals the strengths of the individual models. CatBoost turns out to be the best-performing model overall, showing the best results in 10 out of 20 datasets, with an average rank of 2.93. No other black-box model achieves comparable results. Instead, the strongest competitor is \gls{ebm}, where it is striking to observe that the interpretable model achieves a remarkable average rank of 4.08. As such, it even dominates the four black-box models \gls{mlp}, \gls{rf}, \gls{xgb}, and TabNet with average ranks of 5.25, 5.45, 5.73, and 8.10, respectively. If we look only at the results of the classification tasks, \gls{ebm} even takes the top position together with CatBoost, where both models achieve an average rank of 3.55.

The second best interpretable model is \gls{gaminet} with an average rank of 5.58, followed by \gls{tpsplines}, \gls{igann}, and \gls{psplines} with average ranks of 7.00, 7.13, and 7.68, respectively. Thus, all four models also show strong prediction qualities that are not too far behind their black-box counterparts. Apart from that, the remaining interpretable models \gls{nam}, \gls{exnn}, \gls{lr}, and \gls{dt} generally underperform in this setting, as indicated by their inferior average ranks.

In summary, the results challenge the prevalent tendency in the \gls{ml} community to favor black-box models over interpretable alternatives, which in fact clearly show competitive performance, as our evaluation results reveal.

\begin{table}[htp]
\caption{Predictive performance using tuned hyperparameters. Classification tasks are assessed using \gls{auroc}, whereas regression tasks are measured using \gls{rmse}.}
\label{tab:hpo_predictive_performance}
\resizebox{1\textwidth}{!}{%
\begin{tabular}{@{}l@{\hspace{0.1cm}}lllllllllllllll@{}}
\toprule
& & \multicolumn{9}{c}{\textbf{Interpretable Models}} & \multicolumn{5}{c}{\textbf{Black-box Models}} \\ \cmidrule(l){3-11} \cmidrule(l){12-16} 
& & \multicolumn{7}{c}{\textbf{GAMs}} & \multicolumn{2}{c}{\textbf{Traditional}} & & & \\ \cmidrule(l){3-9} \cmidrule(l){10-11}
\textbf{Task} & \textbf{Dataset} & \rotatebox{45}{\textbf{P-Splines}} & \rotatebox{45}{\textbf{TP-Splines}} & \rotatebox{45}{\textbf{EBM}} & \rotatebox{45}{\textbf{NAM}} & \rotatebox{45}{\textbf{GAMI-Net}} & \rotatebox{45}{\textbf{ExNN}} & \rotatebox{45}{\textbf{IGANN}} & \rotatebox{45}{\textbf{LR}} & \rotatebox{45}{\textbf{DT}} & \rotatebox{45}{\textbf{RF}} & \rotatebox{45}{\textbf{XGB}}  & \rotatebox{45}{\textbf{CatBoost}} & \rotatebox{45}{\textbf{MLP}} & \rotatebox{45}{\textbf{TabNet}} \\ \midrule
\multicolumn{1}{l}{\multirow{25}{*}{CLS}} & college    & \msd{0.951}{0.006} & \msd{0.952}{0.007} & \msd{\underline{0.967}}{0.008} & \msd{0.935}{0.011} & \msd{\underline{0.967}}{0.008} & \msd{0.768}{0.025} & \msd{0.953}{0.023} & \msd{0.929}{0.015} & \msd{0.849}{0.027} & \msd{0.964}{0.012} & \msd{0.960}{0.017} & \msd{0.970}{0.008} & \msd{\textbf{0.973}}{0.009} & \msd{0.944}{0.012} \\[15pt]
 & water      & \msd{0.565}{0.019} & \msd{0.589}{0.018} & \msd{0.679}{0.027} & \msd{0.578}{0.023} & \msd{0.675}{0.030} & \msd{\underline{0.680}}{0.017} & \msd{0.664}{0.049} & \msd{0.495}{0.013} & \msd{0.589}{0.027} & \msd{0.684}{0.015} & \msd{0.658}{0.011} & \msd{\textbf{0.688}}{0.021} & \msd{0.677}{0.044} & \msd{0.629}{0.040} \\[15pt]
 & stroke     & \msd{\underline{0.835}}{0.023} &\textit{ \msd{0.828}{0.022}} & \msd{0.834}{0.017} & \msd{0.804}{0.044} & \msd{0.834}{0.026} & \msd{0.701}{0.018} & \msd{\underline{\textbf{0.838}}}{0.027} & \msd{0.834}{0.032} & \msd{0.815}{0.035} & \msd{0.836}{0.018} & \msd{0.823}{0.027} & \msd{0.830}{0.019} & \msd{0.662}{0.176} & \msd{0.795}{0.037} \\[15pt]
 & churn      & \msd{0.849}{0.014} & \msd{0.849}{0.013} & \msd{\underline{\textbf{0.849}}}{0.011} & \msd{0.846}{0.014} & \msd{0.847}{0.015} & \msd{0.776}{0.017} & \msd{0.846}{0.013} & \msd{0.838}{0.014} & \msd{0.823}{0.016} & \msd{0.841}{0.013} & \msd{0.844}{0.014} & \msd{0.848}{0.012} & \msd{0.842}{0.018} & \msd{0.832}{0.016} \\[15pt]
 & recidivism & \msd{0.741}{0.019} & \msd{0.743}{0.017} & \msd{\underline{\textbf{0.747}}}{0.017} & \msd{0.742}{0.017} & \msd{0.740}{0.017} & \msd{0.632}{0.014} & \msd{0.744}{0.019} & \msd{0.726}{0.017} & \msd{0.709}{0.015} & \msd{0.736}{0.016} & \msd{0.744}{0.018} & \msd{0.742}{0.026} & \msd{0.740}{0.017} & \msd{0.722}{0.016} \\[15pt]
 & credit     & \msd{0.807}{0.009} & \msd{\underline{\textbf{0.810}}}{0.008}& \msd{0.807}{0.010} & \msd{0.802}{0.009} & \msd{0.804}{0.010} & \msd{0.787}{0.017} & \msd{0.802}{0.010} & \msd{0.795}{0.013} & \msd{0.768}{0.011} & \msd{0.801}{0.011} & \msd{0.802}{0.012} & \msd{0.797}{0.006} & \msd{0.796}{0.009} & \msd{0.779}{0.021} \\[15pt]
 & income     & \msd{0.916}{0.002} & \textit{\msd{0.920}{0.003}} & \msd{\underline{0.928}}{0.002} & \msd{0.916}{0.001} & \msd{0.915}{0.002} & \msd{0.609}{0.007} & \msd{0.916}{0.002} & \msd{0.906}{0.002} & \msd{0.902}{0.003} & \msd{0.917}{0.003} & \msd{\textbf{0.929}}{0.002} & \msd{\textbf{0.929}}{0.002} & \msd{0.915}{0.002} & \msd{0.906}{0.002} \\[15pt]
 & bank       & \msd{0.777}{0.011} & \msd{0.778}{0.009} & \msd{\underline{0.798}}{0.011} & \msd{0.772}{0.014} & \msd{0.772}{0.014} & \msd{0.609}{0.008} & \msd{0.772}{0.007} & \msd{0.764}{0.008} & \msd{0.719}{0.026} & \msd{0.795}{0.006} & \msd{0.797}{0.008} & \msd{\textbf{0.801}}{0.006} & \msd{0.787}{0.009} & \msd{0.747}{0.016} \\[15pt]
 & airline    & \msd{0.980}{0.001} & \msd{0.980}{0.001} & \msd{\underline{0.991}}{0.001} & \msd{0.976}{0.001} & \msd{0.989}{0.001} & \msd{\underline{0.991}}{0.001} & \msd{0.989}{0.001} & \msd{0.927}{0.002} & \msd{0.958}{0.003} & \msd{0.994}{0.000} & \msd{\textbf{0.995}}{0.000} & \msd{\textbf{0.995}}{0.000} & \msd{0.994}{0.000} & \msd{0.994}{0.000} \\[15pt]
 & weather    & \msd{0.874}{0.003} & \msd{0.874}{0.003} & \msd{\underline{0.878}}{0.003} & \msd{0.870}{0.003} & \msd{0.876}{0.003} & \msd{0.573}{0.005} & \msd{0.878}{0.003} & \msd{0.868}{0.003} & \msd{0.840}{0.001} & \msd{0.887}{0.002} & \msd{0.891}{0.003} & \msd{\textbf{0.893}}{0.002} & \msd{0.887}{0.002} & \msd{0.877}{0.002} \\[15pt] \midrule
\multicolumn{1}{l}{\multirow{25}{*}{REG}} & car          & \msd{0.293}{0.124} & \msd{0.343}{0.064} & \msd{\underline{\textbf{0.262}}}{0.084} & \msd{0.422}{0.241} & \msd{0.366}{0.145} & \msd{1.689}{0.613} & \msd{0.345}{0.166} & \msd{0.326}{0.097} & \msd{0.346}{0.102} & \msd{0.309}{0.093} & \msd{0.291}{0.077} & \msd{0.286}{0.086} & \msd{0.330}{0.087} & \msd{0.687}{0.458} \\[15pt]
 & student      & \msd{0.862}{0.136} & \msd{0.856}{0.132} & \msd{\underline{0.843}}{0.134} & \msd{0.850}{0.134} & \msd{0.873}{0.113} & \msd{1.675}{0.251} & \msd{0.861}{0.123} & \msd{0.868}{0.147} & \msd{0.885}{0.136} & \msd{\textbf{0.840}}{0.140} & \msd{0.853}{0.142} & \msd{0.841}{0.153} & \msd{0.915}{0.147} & \msd{0.927}{0.169} \\[15pt]
 & productivity & \msd{0.832}{0.096} & \msd{0.768}{0.060} & \msd{\underline{0.716}}{0.037} & \msd{0.859}{0.030} & \msd{0.746}{0.053} & \msd{1.023}{0.049} & \msd{0.775}{0.050} & \msd{0.846}{0.031} & \msd{0.842}{0.046} & \msd{0.705}{0.054} & \msd{0.693}{0.034} & \msd{\textbf{0.689}}{0.033} & \msd{0.773}{0.041} & \msd{0.848}{0.055} \\[15pt]
 & insurance    & \msd{0.501}{0.019} & \msd{0.499}{0.018} & \msd{\underline{0.377}}{0.026} & \msd{0.509}{0.023} & \msd{0.381}{0.023} & \msd{0.618}{0.021} & \msd{0.381}{0.021} & \msd{0.503}{0.018} & \msd{0.405}{0.016} & \msd{0.378}{0.025} & \msd{\textbf{0.373}}{0.027} & \msd{0.373}{0.033} & \msd{0.397}{0.020} & \msd{0.443}{0.049} \\[15pt]
 & crimes       & \msd{0.575}{0.056} & \msd{\underline{\textbf{0.572}}}{0.061} & \msd{0.585}{0.065} & \msd{0.607}{0.052} & \msd{0.585}{0.055} & \msd{0.597}{0.081} & \msd{0.581}{0.056} & \msd{0.587}{0.062} & \msd{0.676}{0.047} & \msd{0.595}{0.062} & \msd{0.592}{0.064} & \msd{0.577}{0.058} & \msd{0.596}{0.060} & \msd{0.629}{0.061} \\[15pt]
 & farming      & \msd{0.665}{0.043} & \msd{0.667}{0.042} & \msd{0.685}{0.044} & \msd{0.708}{0.041} & \msd{0.657}{0.043} & \msd{0.666}{0.031} & \msd{\underline{0.656}}{0.044} & \msd{0.693}{0.045} & \msd{0.729}{0.059} & \msd{0.674}{0.038} & \msd{0.678}{0.039} & \msd{0.677}{0.045} & \msd{\textbf{0.650}}{0.042} & \msd{0.666}{0.050} \\[15pt]
 & wine         & \msd{0.828}{0.038} & \msd{0.898}{0.194} & \msd{\underline{0.765}}{0.021} & \msd{0.822}{0.019} & \msd{0.801}{0.022} & \msd{0.813}{0.031} & \msd{0.795}{0.026} & \msd{0.852}{0.027} & \msd{0.845}{0.029} & \msd{0.678}{0.032} & \msd{0.686}{0.031} & \msd{\textbf{0.669}}{0.028} & \msd{0.779}{0.021} & \msd{0.814}{0.026} \\[15pt]
 & bike         & \msd{0.551}{0.009} & \msd{0.553}{0.010} & \msd{\underline{0.296}}{0.006} & \msd{0.570}{0.014} & \msd{0.297}{0.007} & \msd{0.914}{0.017} & \msd{0.408}{0.009} & \msd{0.776}{0.017} & \msd{0.325}{0.006} & \msd{0.234}{0.008} & \msd{0.214}{0.006} & \msd{\textbf{0.199}}{0.006} & \msd{0.224}{0.009} & \msd{0.320}{0.041} \\[15pt]
 & housing      & \msd{0.521}{0.016} & \msd{0.522}{0.016} & \msd{\underline{0.439}}{0.012} & \msd{0.555}{0.013} & \msd{0.478}{0.016} & \msd{0.501}{0.024} & \msd{0.490}{0.015} & \msd{0.603}{0.020} & \msd{0.539}{0.017} & \msd{0.422}{0.014} & \msd{0.397}{0.011} & \msd{\textbf{0.387}}{0.010} & \msd{0.442}{0.009} & \msd{0.461}{0.012} \\[15pt]
 & diamond      & \msd{0.433}{0.289} & \msd{2.122}{3.271} & \msd{0.207}{0.012} & \msd{0.268}{0.009} & \msd{0.217}{0.005} & \msd{0.980}{0.013} & \msd{0.212}{0.007} & \msd{0.284}{0.007} & \msd{\underline{0.182}}{0.004} & \msd{0.137}{0.003} & \msd{\textbf{0.134}}{0.002} & \msd{0.134}{0.004} & \msd{0.136}{0.003} & \msd{0.168}{0.031} \\[15pt] \midrule
\multicolumn{2}{l}{Average rank CLS}  & \multicolumn{1}{c}{7.15}  & \multicolumn{1}{c}{6.30}   & \multicolumn{1}{c}{3.35}   & \multicolumn{1}{c}{8.95}    & \multicolumn{1}{c}{6.75}   & \multicolumn{1}{c}{11.85}   & \multicolumn{1}{c}{5.95}   & \multicolumn{1}{c}{11.35} & \multicolumn{1}{c}{12.60} & \multicolumn{1}{c}{5.35}  & \multicolumn{1}{c}{4.80}  & \multicolumn{1}{c}{3.40}  & \multicolumn{1}{c}{6.95}  & \multicolumn{1}{c}{10.25}   \\
\multicolumn{2}{l}{Average rank REG}  & \multicolumn{1}{c}{8.10}  & \multicolumn{1}{c}{8.80}   & \multicolumn{1}{c}{4.75}   & \multicolumn{1}{c}{11.20}   & \multicolumn{1}{c}{6.90}   & \multicolumn{1}{c}{11.65}   & \multicolumn{1}{c}{6.65}   & \multicolumn{1}{c}{10.80} & \multicolumn{1}{c}{10.50} & \multicolumn{1}{c}{4.30}  & \multicolumn{1}{c}{3.80}  & \multicolumn{1}{c}{2.30}  & \multicolumn{1}{c}{6.00}  & \multicolumn{1}{c}{9.25}   \\
 \midrule
\multicolumn{2}{l}{Average rank total}  & \multicolumn{1}{c}{7.63}  & \multicolumn{1}{c}{7.55}   & \multicolumn{1}{c}{4.05}   & \multicolumn{1}{c}{10.08}    & \multicolumn{1}{c}{6.83}   & \multicolumn{1}{c}{11.75}   & \multicolumn{1}{c}{6.30}   & \multicolumn{1}{c}{11.08} & \multicolumn{1}{c}{11.55} & \multicolumn{1}{c}{4.83}  & \multicolumn{1}{c}{4.30}  & \multicolumn{1}{c}{2.85}  & \multicolumn{1}{c}{6.48}  & \multicolumn{1}{c}{9.75}  \\ \bottomrule
\end{tabular}%
}
\end{table}

\noindent \textbf{Tuned setting.} Next, we focus on the prediction results of the hyperparameter-tuned models, obtained via a grid search.\footnote{The corresponding hyperparameter configurations for each dataset and model can be found in Appendix~\ref{app:hyperparameter}.} Table~\ref{tab:hpo_predictive_performance} summarizes the results of the evaluation.
Again, we can see that several intrinsically interpretable models, such as \gls{ebm}, \gls{igann}, and \gls{gaminet}, are characterized by strong prediction qualities. Similar to the default setting, the interpretable models perform best in 6 out of 20 datasets, whereas the black-box models show better results in 14 out of 20 datasets.

Although the black-box models achieve some performance gains in the tuned setting, the difference in performance between the best models from both groups remains comparatively small. For the classification tasks, the smallest difference in performance is 0.001~(income: \gls{ebm} 0.928 vs. \gls{xgb}/CatBoost 0.929) and the largest difference is 0.015 (weather: \gls{ebm} 0.878 vs. CatBoost 0.893). For the regression tasks, the smallest difference is 0.003 (student: \gls{ebm} 0.843 vs. \gls{rf} 0.840), whereas the largest difference is 0.097 (bike: \gls{ebm} 0.296 vs. CatBoost 0.199).

Looking at the individual models, it is remarkable that \gls{ebm} still remains the second best-performing model overall with an average rank of 4.05, just behind CatBoost with an average rank of 2.85. For the classification datasets, the interpretable model even slightly outperforms CatBoost by achieving the best average rank across all models with a score of 3.35. Furthermore, \gls{ebm} shows the highest predictive performance across all models in 4 out of 20 datasets and outperforms its interpretable competitors in 14 out of 20 datasets.
The next best models after CatBoost and \gls{ebm} are \gls{xgb} and \gls{rf} with average ranks of 4.30 and 4.83, respectively. This is followed by \gls{igann} as another interpretable model with an average rank of 6.30, outperforming even the two black-box models \gls{mlp} and TabNet with average ranks of 6.48 and 9.75, respectively. \gls{gaminet}, \gls{tpsplines}, and \gls{psplines} rank in the middle with average ranks of 6.83, 7.55, and 7.63, respectively, while the remaining models \gls{nam}, \gls{exnn}, \gls{lr}, and \gls{dt} again exhibit inferior performance.

Moreover, if we compare the best interpretable model \gls{ebm} with the best black-box models CatBoost, \gls{xgb}, and \gls{rf}, the difference in predictive performance is mostly in the range of <~0.01. Only for some larger regression datasets, the three black-box models have slight advantages with differences in the range of <~0.099. Overall, however, the results show that the performance difference between the interpretable \glspl{gam} and the black-box models is marginally small.

\noindent \textbf{Predictive performance difference between default and tuned models.} To shift the perspective towards a better understanding of the change in predictive performance resulting from the hyperparameter tuning, we summarize the performance difference between the default setting and the tuned setting in Table~\ref{tab:diff_predictive_performance}. For better readability, we only report values that are greater than the standard deviation of the tuned models. 

\begin{table}[htp]
\caption{Difference in models' predictive performance between tuned and default hyperparameters. Only changes that deviate beyond the standard deviation are reported. Note that a positive value for classification tasks represents a performance improvement as the difference is measured via \gls{auroc}, whereas for regression tasks a negative value indicates an improvement as the difference is measured via \gls{rmse}.}
\label{tab:diff_predictive_performance}
\resizebox{1\textwidth}{!}{%
\begin{tabular}{@{}ll
S[table-format=-2.3]
S[table-format=-2.3]
S[table-format=-2.3]
S[table-format=-2.3]
S[table-format=-2.3]
S[table-format=-2.3]
S[table-format=-3.3]
S[table-format=-2.3]
S[table-format=-2.3]
S[table-format=-2.3]
S[table-format=-2.3]
S[table-format=-2.3]
S[table-format=-2.3]
S[table-format=-2.3]
 @{}}
\toprule
& & \multicolumn{9}{c}{\textbf{Interpretable Models}} & \multicolumn{5}{c}{\textbf{Black-box Models}} \\ \cmidrule(l){3-11} \cmidrule(l){12-16} 
& & \multicolumn{7}{c}{\textbf{GAMs}} & \multicolumn{2}{c}{\textbf{Traditional}} & & & \\ \cmidrule(l){3-9} \cmidrule(l){10-11}

\textbf{Task} & \textbf{Dataset} & \rotatebox{45}{\textbf{P-Splines}} & \rotatebox{45}{\textbf{TP-Splines}} & \rotatebox{45}{\textbf{EBM}} & \rotatebox{45}{\textbf{NAM}} & \rotatebox{45}{\textbf{GAMI-Net}} & \rotatebox{45}{\textbf{ExNN}} & \rotatebox{45}{\textbf{IGANN}} & \rotatebox{45}{\textbf{LR}} & \rotatebox{45}{\textbf{DT}} & \rotatebox{45}{\textbf{RF}} & \rotatebox{45}{\textbf{XGB}}  & \rotatebox{45}{\textbf{CatBoost}} & \rotatebox{45}{\textbf{MLP}} & \rotatebox{45}{\textbf{TabNet}} \\ \midrule

\multicolumn{1}{l}{\multirow{10}{*}{CLS}}  
 & college    &  &       &       & 0.046 &       &        &       &  &       &       &       &       &       &        \\
 & water      &  &       &       &       &       &        & 0.091 &  &       &       &       &       &       &        \\
 & stroke     &  &       &       &       &       &        &       &  & 0.255 & 0.036 &       & 0.024 &       &        \\
 & churn      &  &       &       & 0.019 &       &        &       &  & 0.161 & 0.020 & 0.020 &       &       &        \\
 & recidivism &  &       &       & 0.037 &       &        &       &  & 0.107 & 0.062 & 0.020 &       &       &        \\
 & credit     &  &       &       & 0.011 &       &        &       &  & 0.132 &       & 0.021 &       &       &       \\
 & income     &  & \textit{0.004} &       & 0.019 &       &        &       &  & 0.154 & 0.014 & 0.002 &       & 0.002 & -0.002 \\
 & bank       &  &       &       & 0.014 &       &        &       &  & 0.106 & 0.014 &       &       &       & -0.028  \\
 & airline    &  &       & 0.002 & 0.013 & 0.004 & 0.001  & 0.009 &  & 0.014 & 0.001 &       &       &       & 0.001  \\
 & weather    &  &       & 0.003 & 0.018 &       & -0.006 & 0.005 &  & 0.144 & 0.004 & 0.005 & 0.006 & 0.002 &       \\ \midrule
\multicolumn{1}{l}{\multirow{10}{*}{REG}} 
 & car          &        &         &        & -0.293 &        &        &        & -0.144 &        &  &        &        &        &  \\
 & student      &        &         &        &        &        &        &        &        & -0.353 &  &        &        &        &  \\
 & productivity &        &         &        & -0.050 &        &        &        & -0.059 & -0.118 &  & -0.051 &        &        &  \\
 & insurance    &        &         &        & -0.174 &        &        & -0.114 & -0.360 & -0.118 &  & -0.062 &        &        &  \\
 & crimes       & -0.162 &         &        & -0.167 &        & -0.087 &        &        & -0.168 &  &        &        &        &  \\
 & farming      &        &         &        & -0.085 &        &        &        & -0.105 & -0.202 &  & -0.047 &        &        &  \\
 & wine         &        &         &        & -0.058 &        &        &        & -0.041 & -0.121 &  &        & -0.060 &        &  \\
 & bike         &        & -0.028  & -0.019 & -0.178 & -0.009 &        & -0.228 & -0.054 &        &  & -0.013 & -0.031 & -0.031 &  \\
 & housing      &        &         &        & -0.149 &        &        & -0.057 & -0.168 & -0.060 &  & -0.018 & -0.024 & -0.023 &  \\
 & diamond      &        &         & -0.023 & -0.145 &        &        & -0.056 & -0.187 & -0.007 &  & -0.004 & -0.005 & -0.011 &  \\ \midrule
\multicolumn{2}{l}{Default average rank}  & \multicolumn{1}{S[table-format=-2.3]}{7.68}  & \multicolumn{1}{S[table-format=-2.3]}{7.00}   & \multicolumn{1}{S[table-format=-2.3]}{4.08}   & \multicolumn{1}{S[table-format=-2.3]}{11.75}   & \multicolumn{1}{S[table-format=-2.3]}{5.63}   & \multicolumn{1}{S[table-format=-2.3]}{11.40}   & \multicolumn{1}{S[table-format=-2.3]}{7.13}   & \multicolumn{1}{S[table-format=-2.3]}{10.60} & \multicolumn{1}{S[table-format=-2.3]}{12.30} & \multicolumn{1}{S[table-format=-2.3]}{5.45}  & \multicolumn{1}{S[table-format=-2.3]}{5.73}  & \multicolumn{1}{S[table-format=-2.3]}{2.93}  & \multicolumn{1}{S[table-format=-2.3]}{5.25}  & \multicolumn{1}{S[table-format=-2.3]}{8.10}   \\
\multicolumn{2}{l}{Tuned average rank}  & \multicolumn{1}{S[table-format=-2.3]}{7.63}  & \multicolumn{1}{S[table-format=-2.3]}{7.55}   & \multicolumn{1}{S[table-format=-2.3]}{4.05}   & \multicolumn{1}{S[table-format=-2.3]}{10.08}    & \multicolumn{1}{S[table-format=-2.3]}{6.83}   & \multicolumn{1}{S[table-format=-2.3]}{11.75}   & \multicolumn{1}{S[table-format=-2.3]}{6.30}   & \multicolumn{1}{S[table-format=-2.3]}{11.08} & \multicolumn{1}{S[table-format=-2.3]}{11.55} & \multicolumn{1}{S[table-format=-2.3]}{4.83}  & \multicolumn{1}{S[table-format=-2.3]}{4.30}  & \multicolumn{1}{S[table-format=-2.3]}{2.85}  & \multicolumn{1}{S[table-format=-2.3]}{6.48}  & \multicolumn{1}{S[table-format=-2.3]}{9.75}   \\
\midrule
\multicolumn{2}{l}{Rank difference}  & \multicolumn{1}{S[table-format=-2.3]}{$\uparrow$ 0.05}  & \multicolumn{1}{S[table-format=-2.3]}{$\downarrow$ 0.55}  & \multicolumn{1}{S[table-format=-2.3]}{$\uparrow$ 0.03}  & \multicolumn{1}{S[table-format=-2.3]}{$\uparrow$ 1.68}    & \multicolumn{1}{S[table-format=-2.3]}{$\downarrow$ 1.20}  & \multicolumn{1}{S[table-format=-2.3]}{$\downarrow$ 0.35}   & \multicolumn{1}{S[table-format=-2.3]}{$\uparrow$ 0.83}   & \multicolumn{1}{S[table-format=-2.3]}{$\downarrow$ 0.48} & \multicolumn{1}{S[table-format=-2.3]}{$\uparrow$ 0.75}  & \multicolumn{1}{S[table-format=-2.3]}{$\uparrow$ 0.63}  & \multicolumn{1}{S[table-format=-2.3]}{$\uparrow$ 1.43}  & \multicolumn{1}{S[table-format=-2.3]}{$\uparrow$ 0.07}  & \multicolumn{1}{S[table-format=-2.3]}{$\downarrow$ 1.23} & \multicolumn{1}{S[table-format=-2.3]}{$\downarrow$ 1.65} \\
\bottomrule
\end{tabular}%
}
\end{table}

The results reveal several insightful observations. For example, the majority of \glspl{gam}, including \gls{ebm}, \gls{psplines}, \gls{tpsplines}, \gls{gaminet}, and \gls{exnn}, generally show little or no improvement. This illustrates that the models in their default configurations already produce fairly robust results without the need for extensive hyperparameter tuning. In the case of \gls{igann}, we can see some improvements, especially for larger datasets. In contrast, the largest differences can be noted for \gls{nam}. Here, we find improvements in almost all datasets, indicating that \gls{nam} is very tuning-intensive and must be adapted to the particular circumstances of the datasets.

Similarly, several of the traditional models, including \gls{lr}, \gls{dt}, \gls{rf}, and \gls{xgb}, require additional hyperparameter tuning to unfold better prediction qualities. While \gls{lr} shows notable improvements only for regression tasks, the situation is reversed for \gls{rf}, which shows improvements only for classification tasks. In contrast, \gls{dt} improves the most among all four models, showing performance gains on almost all datasets. This is not surprising, as decision trees generally tend to overfit without explicit model tuning. Similarly, \gls{xgb} shows improvements on almost all datasets. As a result, the average ranks of \gls{rf} and \gls{xgb} move much closer to the average ranks of the best-performing models CatBoost and \gls{ebm}. Nevertheless, it is striking to observe that especially the intrinsically interpretable model \gls{ebm} still dominates the majority of black-box models even without the need of additional performance gains through extensive hyperparameter tuning.

In summary, we find that the results remain relatively consistent in both settings. Looking at the rank differences at the bottom of Table~\ref{tab:diff_predictive_performance}, we see a slight improvement of about 1.5 average ranks for \gls{xgb} and \gls{nam}, while \gls{gaminet}, \gls{mlp}, and TabNet decline by approximately 1.5 average ranks. Despite these variations, the overall performance ratios remain relatively stable. Therefore, we do not observe a remarkable performance advantage in favor of the black-box models, which is often a widespread belief according to the assumption of the performance-interpretability trade-off \citep{rudin2019stop}. In fact, the opposite seems to be true, as the interpretable models demonstrate competitive performance levels while offering transparent model structures. We delve deeper into this aspect in the next section.

\subsection{Assessment of Model Interpretability}
\label{sec:results_interpret}

As described in Section~\ref{sec:method_interpret}, the interpretability assessment is divided into three parts. First, we consider the \glspl{gam}' ability to externalize their inner workings by offering locally and globally interpretable results. Then, we compare the visual output of the different models to highlight similarities and differences between them. Finally, we evaluate the models' level of interpretability using the objective evaluation criteria introduced above in Table~\ref{tab:interpret_metrics}.

\vspace{0.25cm}
\noindent \textbf{Global and local interpretability.} For the first assessment, we consider an \gls{ebm} model and a reduced version of the \textit{bike} dataset to illustrate the expressiveness of the feature plots in terms of their global and local interpretability. Although we specifically use \gls{ebm} for our demonstration (because it provides the best results among all interpretable models for this particular dataset), it is important to emphasize that it only serves as one representative example of the entire \gls{gam} family. The demonstration provides insight into the inherent properties of \glspl{gam} as a whole, and is broadly applicable to the other models as well. An exception, however, are the \gls{exnn} and \gls{nam} models, which show a different behavior and therefore are discussed separately.

The \textit{bike} dataset contains about $17,000$ samples of hourly counts of rented bikes from a bike-sharing system, along with 12 features that include details about weather and seasonal information \citep{fanaee-t_event_2014}. As such, we consider a regression task with the goal of predicting the number of bikes rented per hour. For our demonstration, we trained a model with a filtered feature set to provide a holistic discussion of the different global and local interpretation perspectives. Specifically, we consider two numerical features (perceived temperature: \textit{atemp}, hour of the day: \textit{hr}), one categorical feature (weather situation: \textit{weathersit}), and two pairwise interaction terms that are automatically identified by the \gls{ebm} model (\textit{atemp} $\times$ \textit{hr}, \textit{weathersit\_3} $\times$ \textit{hr}).

\begin{figure}[ht]
\centering
    \includegraphics[width=1\textwidth]{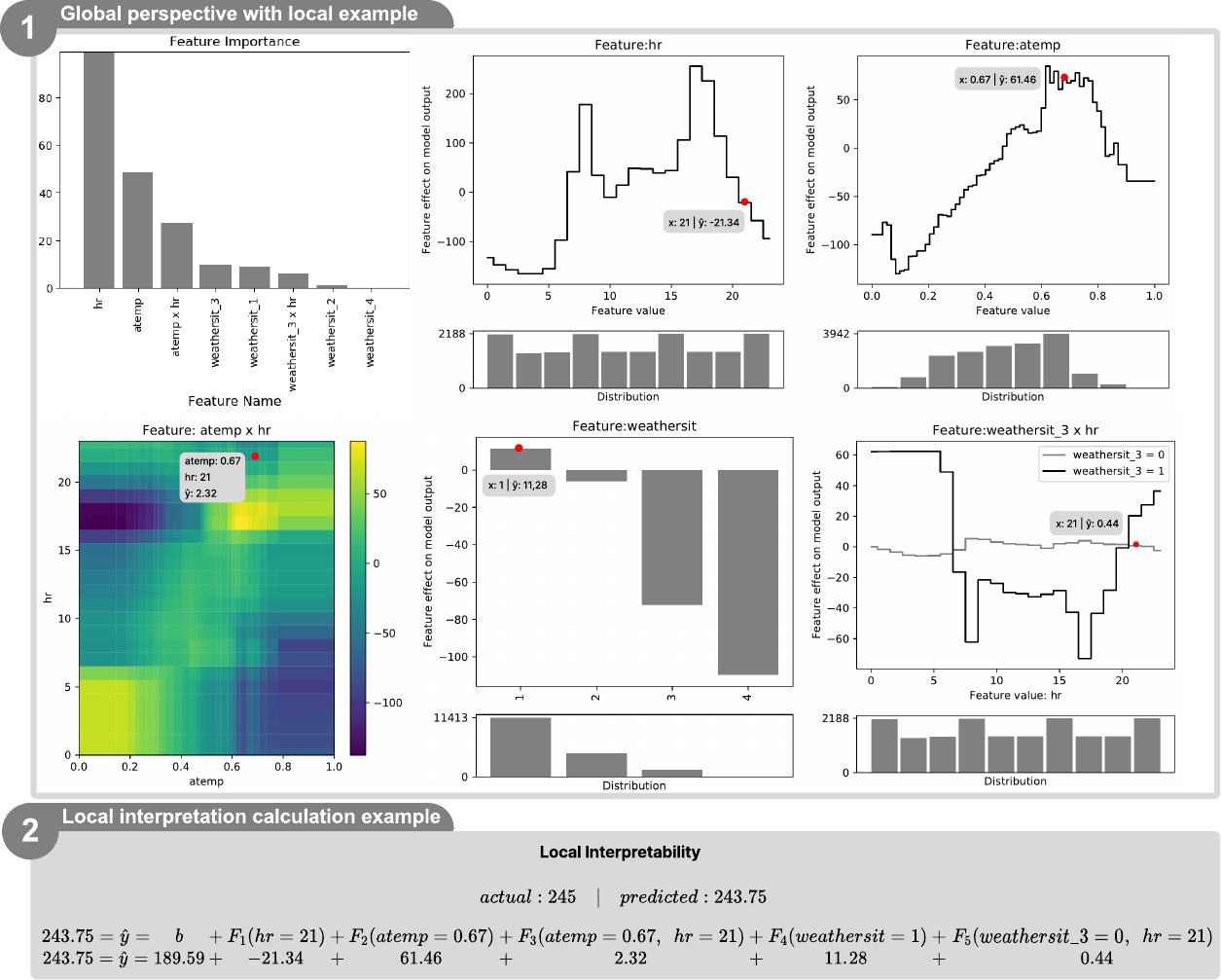}
    \caption{Global and local interpretability of an \gls{ebm} model. (1) The summary plots visualize the overall feature importance and the global relationships between the input features and the prediction target. The red dots illustrate the possibility of reading local values of individual feature effects for an exemplary sample. (2) The corresponding calculation example shows the additive contributions of each feature to the model output to retrieve the final prediction for the selected sample.}
    \label{fig:ebm_plots}
\end{figure}

After fitting the model for the regression task, it offers several feature plots that visualize the captured relationships between the input features and the predicted number of rented bikes, as summarized in Figure~\ref{fig:ebm_plots}. The first plot displays the model's \textit{feature importance}, indicating that \textit{hr} is the most important feature and \textit{weathersit\_4} is the least important feature. The remaining plots show the learned \textit{shape functions} of the model, revealing how each value of the input features affects the predicted output. 

Depending on the feature types, different shape plots are generated. The effects of numerical features are shown by line plots with curved shapes (cf. \textit{hr} and \textit{atemp}), whereas the effects of categorical features are shown by bar plots (cf. \textit{weathersit}). The $x$-axes represent the feature values and the $y$-axes represent their impact on the target variable compared to the average prediction value (i.e., positive impact for $y>0$, negative impact for $y<0$). For example, it can be seen that the most important feature \textit{hr} has a strong positive impact with x-values around 9 and 17, whereas feature values below 7 generally have a negative impact on the predicted output. In other words, the model captures the likely relationship that many bikes will be rented during the morning and afternoon rush hours, whereas at night the number strongly declines.

The shape plots for pairwise interactions are slightly different. For interaction terms between a numerical and categorical feature, a line plot is used with different lines representing the distinct groups of the categorical feature (cf. \textit{weathersit\_3} $\times$ \textit{hr}). For interaction terms between two numerical features, both the $x$-axis and the $y$-axis represent the feature values, whereas the impact on the target is highlighted with a corresponding color scheme, resulting in a two-dimensional heatmap (cf. \textit{atemp} $\times$ \textit{hr}). For example, it can be seen that low temperatures during the early evening hours have a strong negative impact on the predicted outcome (i.e., dark blue area), whereas medium to high temperatures during the same time period have a strong positive impact on the predicted outcome (i.e., light yellow area).

Taken together, the shape plots collectively externalize the model's inner decision logic and visualize how each feature affects the prediction target over its entire range of values. Thus, they address the demand of providing \textit{intrinsic global interpretability}, which is necessary to understand the model's behavior as a whole and to identify potential biases and other pitfalls \citep{du_techniques_2019, rudin2019stop}.

Apart from that, it is also possible to replicate and verify the prediction scores for individual samples by extracting instance-based computations. This procedure is indicated by the red dots in Figure~\ref{fig:ebm_plots}. The sum based on the $x$ and $y$ values is displayed in the gray box in the lower part of Figure~\ref{fig:ebm_plots}, highlighting the model's functionality to provide \textit{intrinsic local interpretability}. This is useful to understand the reasoning behind individual predictions, or why a particular prediction was made for a given input. Furthermore, the example shows that the shape plots are not just an approximate explanation of relevant relationships, but an \textit{exact} description of how the intrinsically interpretable model computes a prediction. This is a unique property of the \glspl{gam}, which, unlike complex black-box models, add up each feature effect to obtain a transparently derived prediction value for comprehensible decision support. 

\vspace{0.25cm}
\noindent \textbf{Comparison of shape plots between \glspl{gam}.} After verifying the models' global and local interpretability, we continue to examine similarities and variations between the visual outputs of the different \glspl{gam}, which are mostly reflected in the appearance of their shape plots. On this basis, we can derive statements regarding their merits and limitations for different interpretability contexts.

\begin{figure}[hb!]
\centering
    \includegraphics[width=1\textwidth]{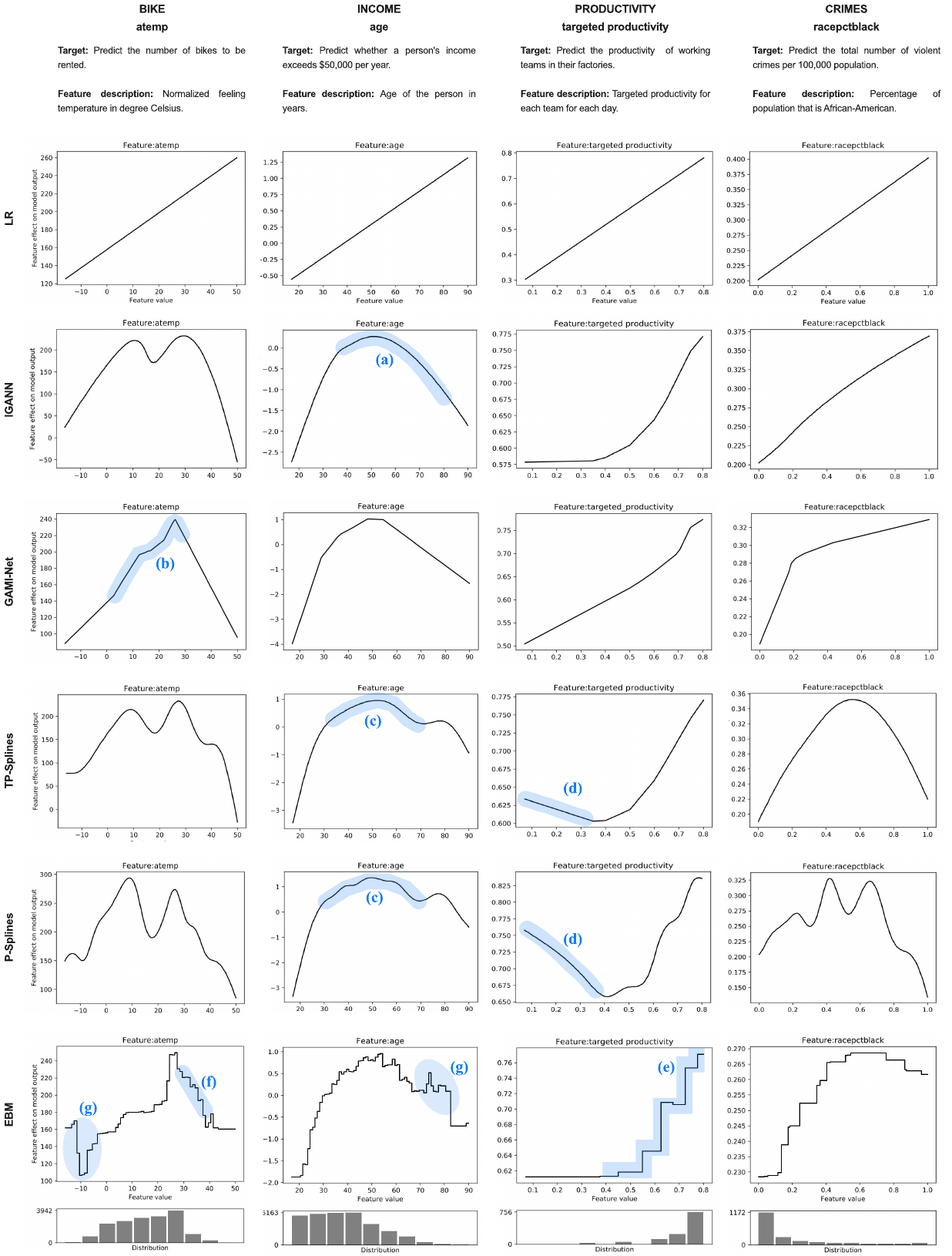}
    \caption{Comparison of shape plots learned by five different \glspl{gam} and a linear model.}
    \label{fig:shape_comparison}
\end{figure}

Figure~\ref{fig:shape_comparison} summarizes exemplary results for five \glspl{gam} that are roughly similar but differ in detail, including \gls{igann}, \gls{gaminet}, \gls{tpsplines}, \gls{psplines}, and \gls{ebm}.\footnote{Note that most \glspl{gam} are capable of producing confidence bands to provide insight into model uncertainty/variability in certain feature regions. However, this functionality was neglected in Figure~\ref{fig:shape_comparison} to focus on the visual properties of the different shape functions. For completeness, another version of the figure with confidence bands is provided in Appendix~\ref{app:visual_output}.} The results of the different models are shown in the rows of the grid. Additionally, we have included the output of the linear model (i.e., LR), which serves as a reference line. The columns of the grid in Figure~\ref{fig:shape_comparison} represent selected features from four different datasets, with a brief description of the features and the prediction targets.

As expected, the linear model yields the simplest relationships that are easy to comprehend. At the same time, however, the linearity constraint impairs the model's ability to capture more complex feature characteristics, resulting in inferior predictive performance. In contrast, all four \glspl{gam} generally agree on non-linear relationships, with roughly comparable feature representations. However, their different model structures also lead to observable differences. 

For example, \gls{igann} produces fairly smooth shape functions that promote interpretability by avoiding sharp and erratic step functions and minor details, as is the case with other models (e.g., see the light blue area (a) in the \textit{income} dataset for \gls{igann} compared to the same area of all other \glspl{gam}). This behavior is due to \gls{igann}'s principle of initializing the training procedure with linear functions and then gradually adapting to non-linearities if the underlying data requires it \citep{kraus_interpretable_2023}. Interestingly, despite neglecting minor details in the shape functions, \gls{igann} achieves very good performance scores with a ranking at least among the top five tuned models, as the results of the previous section indicate. Thus, it can be concluded that \gls{igann} is less prone to overfitting as it does not capture too much noise.

A similar behavior can also be observed with \gls{gaminet}, which captures smooth shape functions that avoid sharp jumps and small details, while positioning among the best-performing (interpretable) models. Besides that, \gls{gaminet} is able to produce piecewise linear shapes using a ReLU activation function, as highlighted in area (b) of the \textit{bike} dataset. This ensures a high degree of interpretability as it is easy to comprehend how changes in the feature values affect the target variable in different feature regions. Another strength of \gls{gaminet} is that it focuses on sparsity and regularization constraints \citep{yang_gami-net_2021}, which makes it particularly interesting for high-dimensional datasets with many features involved. 

In contrast, the shape functions learned by the spline-based \glspl{gam} are generally characterized by more detail and a lower degree of smoothness. For example, while \gls{tpsplines} use additional penalization and smoothing techniques, \gls{psplines} are just a combination of piecewise polynomial basis functions (i.e., B-splines) with a discrete penalty on the basis coefficients \citep{eilers_flexible_1996}. Therefore, \gls{tpsplines} generally result in smoother shape functions compared to \gls{psplines} (cf. highlighted areas (c)). Nevertheless, the degree of smoothness can also be controlled by several hyperparameters, such as the number of splines. That is, more splines generally mean curvier shape functions (see Appendix~\ref{app:visual_output}). Moreover, it can be observed that both \gls{psplines} and \gls{tpsplines} tend to produce overly confident extrapolations in feature regions with small sample sizes \citep{chang_how_2021}. Some examples of such extrapolations can be seen in the highlighted areas (d) of the \textit{productivity} dataset, where both models capture the misleading relationship that a lower level of targeted productivity increases the likelihood of a higher level of actual productivity, which may lead to erroneous conclusions.

The \gls{ebm} is located at the other end of the spectrum. It generally produces detailed shape functions that are most similar to those of the spline-based models. However, because \gls{ebm} is based on tree ensembles, the shapes are characterized by piecewise constant step functions \citep{lou_intelligible_2012}. These functions undergo abrupt changes and sharp jumps at discrete values, which is highly favorable in situations where features inherently exhibit step-like patterns. Such an effect can be seen in the highlighted area (e), which shows the feature effects for different discrete productivity levels. In other situations, however, such sharp jumps are hard to justify for decision-making purposes (cf. highlighted area (f) that shows an erratic downward trend for the feature effect). Similarly, the model tends to capture even subtle nuances in the data (cf. highlighted areas (g)), which further impairs the model's interpretability. Nevertheless, such detailed patterns can also have positive implications, such as detecting anomalies that aid in removing data quality issues or support model debugging purposes. Furthermore, this comprehensive approach appears to be effective, as the \gls{ebm} model generally shows the strongest predictive performance in comparison to CatBoost, \gls{xgb}, and other opaque models.

In contrast to the five \glspl{gam} discussed above, we excluded \gls{nam} and \gls{exnn} from our previous considerations. This is because they show a fundamentally different behavior. \gls{nam} suffers from strong overfitting in all our experiments, which could not be resolved even with extensive hyperparameter tuning and additional model implementations. As a result, the model exhibits below-average performance and produces extremely jagged shape functions that are hardly able to convey a comprehensible decision logic. Further details as well as exemplary shape plots for the \textit{bike} and the \textit{productivity} datasets can be found in Appendix~\ref{app:visual_output}.

The \gls{exnn}, on the other hand, is a special case, as it is based on the structure of an additive index model. For the interpretation of the model, this means that not only a single feature is covered by a corresponding shape function, but that a whole set of features may provide partial contributions to that shape function. Consequently, the model output is barely interpretable in a meaningful way when many features are involved. In Appendix~\ref{app:visual_output}, we provide additional details and an exemplary model output. 

\vspace{0.25cm}
\noindent \textbf{Model assessment based on interpretability criteria.} Based on the detailed model inspections above, it is possible to evaluate all models regarding the various interpretability criteria described in Section~\ref{sec:method_interpret}. The results of our evaluation process are summarized in Table~\ref{tab:interpretability-scoring}.

\begin{table}[ht]
\centering
\caption{Results of the interpretability assessment.}
\label{tab:interpretability-scoring}
\resizebox{\columnwidth}{!}{
\begin{tabular}{lcccccccccccccc}
\toprule
& \multicolumn{9}{c}{\textbf{Interpretable Models}} & \multicolumn{5}{c}{\textbf{Black-box Models}} \\ 
\cmidrule(l){2-10} \cmidrule(l){11-15} 
& \multicolumn{7}{c}{\textbf{GAMs}} & \multicolumn{2}{c}{\textbf{Traditional}} & & & \\ \cmidrule(l){2-8} \cmidrule(l){9-10}

\textbf{Task} & \rotatebox{45}{\textbf{P-Splines}} & \rotatebox{45}{\textbf{TP-Splines}} & \rotatebox{45}{\textbf{EBM}} & \rotatebox{45}{\textbf{NAM}} & \rotatebox{45}{\textbf{GAMI-Net}} & \rotatebox{45}{\textbf{ExNN}} & \rotatebox{45}{\textbf{IGANN}} & \rotatebox{45}{\textbf{LR}} & \rotatebox{45}{\textbf{DT}} & \rotatebox{45}{\textbf{RF}} & \rotatebox{45}{\textbf{XGB}}  & \rotatebox{45}{\textbf{CatBoost}} & \rotatebox{45}{\textbf{MLP}} & \rotatebox{45}{\textbf{TabNet}} \\ \midrule

Additivity      & 2              & 2                       & 2            & 2            & 2                 & 0             & 2              & 2                   & 0                  & 0           & 0            & 0                            & 0            & 0                          \\
Sparsity        & 0              & 1                       & 0            & 1            & 2                 & 1             & 1              & 2                   & 1                  & 1           & 1            & 1    & 0            & 2                          \\
Linearity       & 1              & 1                       & 0            & 0            & 2                 & 0             & 2              & 2                   & 0                  & 0           & 0            & 0                            & 0            & 0                          \\
Smoothness      & 1              & 2                       & 1            & 0            & 2                 & 2             & 2              & 2                   & 0                  & 0           & 0            & 0                            & 0            & 0                          \\
Monotonicity    & 1              & 0                       & 1            & 0            & 1                 & 0             & 0              & 2                   & 0                  & 0           & 1            & 1                            & 0            & 0                          \\
Visualizability & 2              & 2                       & 2            & 0            & 2                 & 0             & 2              & 2                   & 2                  & 0           & 0            & 0                            & 0            & 0                          \\ \hline
Sum  & 7              & 8   & 6            & 3            & 11                & 3             & 9              & 12                  & 3                  & 1           & 2            & 2        & 0            & 2      \\ \bottomrule     
\end{tabular}%
}
\end{table}

As expected, the linear model (i.e., \gls{lr}) receives the highest interpretability score (12 points) because all criteria are fully satisfied by design.\footnote{It could be argued that \gls{lr} is not sparse by default, since regularization constraints must be explicitly considered. However, in our experiments, we use LogisticRegression and ElasticNet from the sklearn (Python) package, which explicitly consider L1/L2 regularization.} In contrast, the five black-box models receive the lowest scores, with the neural network (i.e.,~\gls{mlp}) not fulfilling any of the criteria (0 points) and TabNet and the tree ensemble models (i.e., \gls{rf}, \gls{xgb}, CatBoost) having at least the option of incorporating sparsity and/or monotonicity constraints through certain hyperparameters (1 and 2 point(s), respectively).

The scores obtained among the different \glspl{gam} are highly scattered. \gls{gaminet} and \gls{igann} achieve the best scores (11 and 9 points, respectively), since they satisfy most of the criteria by default or at least have additional hyperparameters to address them. \gls{nam} and \gls{exnn}, on the other hand, are ranked at the lower end of the spectrum with 3 points each. The main deficiencies are the inability to capture linear relationships, missing monotonicity constraints, and a lack of visualizability to clearly understand how a change in the input affects the output, which can also be seen in the examples in Appendix~\ref{app:visual_output}. The remaining three \glspl{gam} (i.e., \gls{psplines}, \gls{tpsplines}, and \gls{ebm}) receive medium scores of 7, 8, and 6 points, respectively. This is due to partially fulfilled interpretability criteria or the need to set additional configurations. Thus, depending on the model, suitable hyperparameters must be set in order to meet the corresponding requirements for sparsity, linearity, smoothness, and/or monotonicity (see Appendix~\ref{app:interpret_criteria}).

A final comment should be made on the result of the decision tree. The model is usually considered highly interpretable, but receives low scores in our evaluation procedure. This is due to the fact that shallow decision trees can usually be converted into easily understandable decision rules --- an ability that is not appropriately reflected in the current assessment framework adapted from \cite{Sudjianto2021DesigningII}. We discuss this limitation in Section~\ref{sec:discussion}.

\subsection{Summary of the Performance-Interpretability Evaluation}
\label{sec:results_summary}

\begin{figure}[ht]
\centering
    \includegraphics[width=0.65\textwidth]{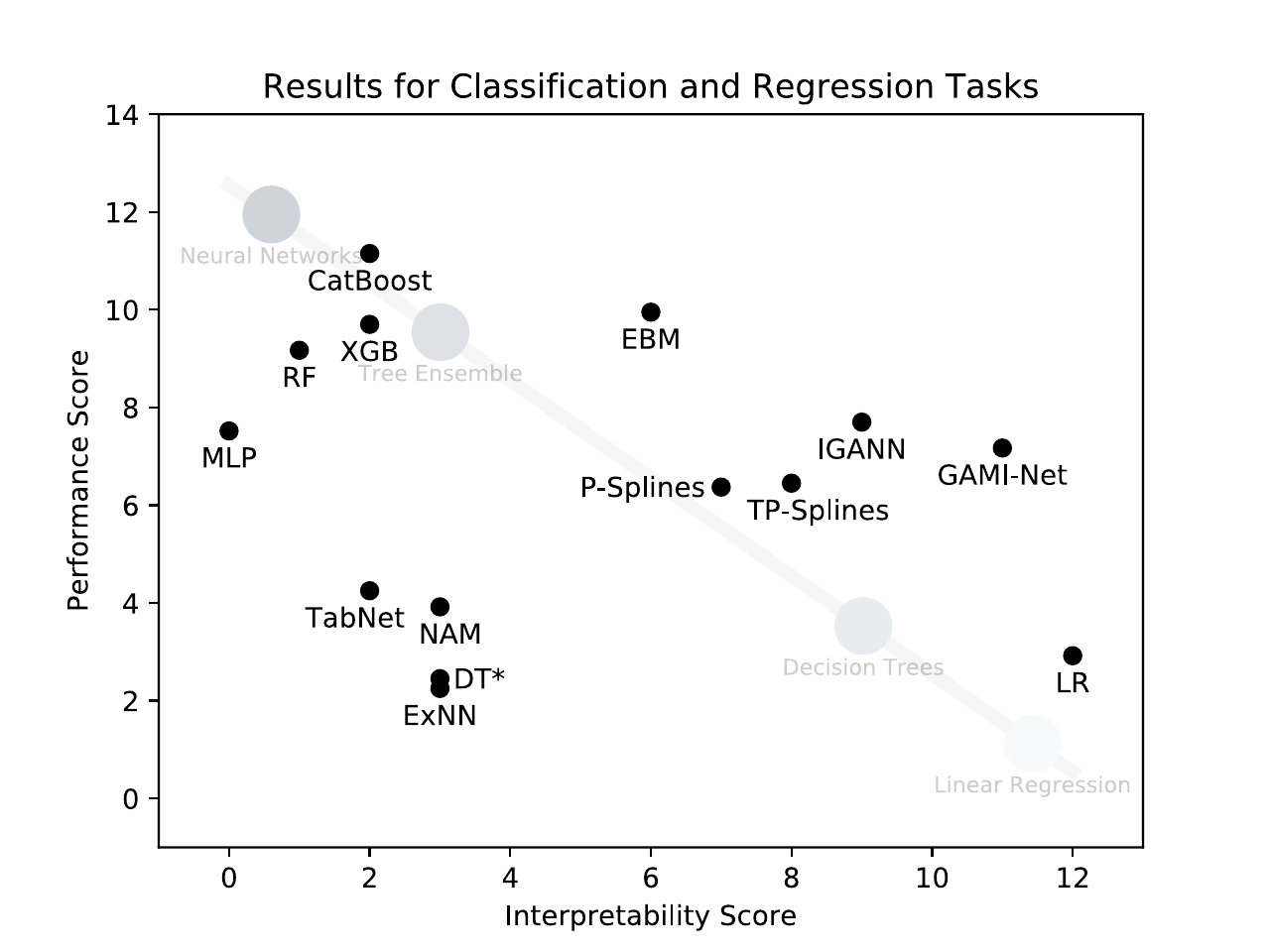}
    \caption{Summary of the performance-interpretability evaluation. The x-axis represents the interpretability score from the interpretability assessment. The y-axis represents the performance score, computed by inverting the average rank of each model.}
    \label{fig:perf_interpret_trade-off}
\end{figure}

In this section, we summarize the results of our previous considerations in order to combine the different evaluation perspectives into a single view. Specifically, we plot the results of the average performance evaluation based on all tuned models on the $y$-axis, and the results of the interpretability evaluation on the $x$-axis, resulting in the performance-interpretability diagram in Figure~\ref{fig:perf_interpret_trade-off}. 

Such performance-interpretability diagrams have been widely used in recent years to emphasize the trade-off between models with high performance but low interpretability (e.g., neural networks, tree ensembles) versus models with high interpretability but low performance (e.g., linear regression, decision trees) \citep[e.g.,][]{barredo_arrieta_explainable_2020, gunning_darpas_2019}. This assumed trade-off, highlighted in gray in Figure~\ref{fig:perf_interpret_trade-off}, has often been used as an argument in academia and industry to justify the need for complex black-box models. Thus, model developers and researchers have argued that only complex black-box models can achieve high predictive performance, which subsequently require post-hoc explanation methods like \gls{shap} or \gls{lime} to simplify their decision logic in a human-digestible way~\citep{rudin2019stop, rudin_why_2019}. However, this trend is problematic since post-hoc explanation methods cannot represent the full, highly complex functioning of black-box models and therefore may provide unreliable approximations or may lead to misleading conclusions~\citep{rudin_why_2019, babic_beware_2021}.

The results of our evaluation study contribute important insights to this central debate. In particular, they challenge the conventional belief and show that there is no strict trade-off between predictive performance and model interpretability for tabular data. Instead, we see that advanced models like \gls{ebm}, \gls{igann}, and \gls{gaminet} are indeed able to achieve high predictive performance while remaining fully interpretable without the need for additional explanation methods. On average, \gls{ebm} even surpasses the performance of most complex black-box models, including \gls{xgb}, \gls{rf}, and \gls{mlp}, providing excellent prediction qualities balanced with decent interpretability. Concurrently, \gls{gaminet} and \gls{igann} demonstrate excellent interpretability scores while maintaining competitive performance metrics. These relations also hold true if the results for regression and classification tasks are considered separately (see Appendix~\ref{app:trade-off}).

\section{Discussion of the Results}
\label{sec:discussion}

In recent years, we have seen a tremendous amount of \gls{ml} applications across a wide range of application domains, with a primary focus on advanced black-box models due to their expected performance benefits. To mitigate the limitation of such model's opaque behavior, the \gls{ml} community has brought forth various post-hoc explanation methods to turn the model's complex decision logic into human-understandable insights. For this reason, post-hoc explanation methods are now widely used in a broad variety of disciplines, which can also be observed in our \gls{is} community \citep[e.g.,][]{wanner_white_2020, bauer_explain_2021, brasse_explainable_2023, zacharias_designing_2022, bauer_explained_2023}. 

However, although post-hoc explanation methods like \gls{shap} and \gls{lime} can offer informative insights, they have to be regarded with caution for two critical reasons. First, the complexity of black-box models can never be fully explained by simple approximations without losing information. As such, they rather provide snapshots of the models' inner decision logic based on a few instances, which may yield incomplete insights. Second, post-hoc explanations can only be provided \emph{post-hoc}, making it impossible to fully validate the functioning of the model for all inputs before model deployment. This becomes particularly critical if the distribution of the input data shifts over time and the model potentially handles value ranges of input features that have not been observed in the training data. As a result, post-hoc explanations have led to misleading conclusions \citep{rudin_why_2019, babic_beware_2021}. Because of these reasons, the option of choosing a complex black-box model that requires post-hoc explanations should only be considered if there is no better alternative \citep{rudin2019stop}.

With our study, we show that black-box models are indeed not necessarily required in the first place to achieve strong prediction qualities for tabular data. By conducting a comprehensive and systematic evaluation study, we shed light on the benefits and limitations of advanced intrinsically interpretable \gls{ml} models. Specifically, this included seven state-of-the-art \glspl{gam}, each built upon diverse foundations such as splines, tree-based algorithms, and tailored neural networks. The results of our experiments clearly show that models like \gls{ebm}, \gls{tpsplines}, \gls{gaminet}, and \gls{igann} do not have to hide behind their opaque counterparts CatBoost, \gls{xgb}, \gls{rf}, and \gls{mlp} for tabular prediction tasks. Therefore, we argue that such interpretable models should be firmly established as first-choice models in predictive modeling projects as we see a large potential for applying them in research and industry alike.

\subsection{Implications for Research and Practice}

Our research has several implications for \gls{is} research and practice. First and foremost, we hope that our results will promote the use of intrinsically interpretable models in our community as a socio-technical discipline. Because of their flexibility to capture complex patterns and their simplicity to produce easily understandable results, they offer a technically equivalent but ethically more acceptable alternative to black-box models. For example, we have seen many examples in the past where bias and fairness issues have been reported in \gls{ml} applications. Prominent examples include hiring applications that have discriminated against women, or criminal justice applications that have made unfair pretrial detention and release decisions for African Americans due to biased predictions \citep{janiesch_machine_2021, rudin_why_2019}. While detecting bias in training data is generally considered a tedious task, with advanced \glspl{gam}, such biases can be better detected using the shape plots of the final model to avoid various types of discrimination. For example, using the shape plot of the \textit{crimes} dataset in Figure~\ref{fig:shape_comparison} as an example, one can quickly see the extent to which the different \glspl{gam} capture racial bias effects that may need to be corrected in order to provide fair decision support.

Likewise, model users can better interact with such interpretable \gls{ml} models because they can directly see how individual features affect the outcome of a model. Thus, domain knowledge can be incorporated into the model by removing or adding features, adjusting certain model properties (e.g., smoothness), or adding other model constraints (e.g., upper and lower bounds on shape functions) to better account for desired patterns. Because of all these characteristics, \glspl{gam} can be considered promising models for critical applications in research and practice.

Furthermore, we see great potential in using such models as a tool for theory development. The inherent interpretability of the models facilitates the identification of meaningful predictors, the assessment of their effect, and the generation of hypotheses about causal relationships. Thus, although the observed feature effects are still based on correlations, and therefore it is not possible to say with certainty why some of the effects shown in the feature shape plots are present, they can still be used as indications for further investigation. Therefore, \glspl{gam} can serve as a bridge between data-driven exploration and theory-building in our socio-technical field, helping researchers to uncover hidden phenomena and refine theoretical frameworks.

In addition to providing important insights and implications for our community, our study also provides a versatile evaluation pipeline for systematically evaluating and comparing a variety of \gls{ml} models.   This pipeline encapsulates all of our models, datasets, metrics, and evaluation procedures, thus promoting both reproducibility and reusability of our experiments. The novelty of our pipeline lies in its broad applicability, which is not limited to the models we studied. Instead, it is tailored to be easily adapted to a range of regression or classification datasets for benchmarking a variety of models, including \glspl{gam}, traditional interpretable models, and black-box models. Moreover, it allows for the inclusion of custom or proprietary models, which are either implemented in Python or R, to meet individual needs in research and practice.

Ultimately, our study provides a broad and lightweight introduction to \glspl{gam} as an important family of interpretable \gls{ml} models. This introduction shows that interpretable machine learning is not just a research area, but a serious, applicable domain. Therefore, we hope that our work will attract the attention of policy and decision makers to promote the relevance of interpretable machine learning beyond the research community. In this way, they might reflect on the responsible and ethical use of AI so that it can guide decisions about the regulation of AI and \gls{ml} technologies in a reasonable way.

\subsection{Limitations and Outlook}

As with any research, our work is not without limitations. By acknowledging them, we derive a basis for future work to guide promising research avenues in this field.

First, we limited our evaluation study to medium-sized datasets to keep the computational cost manageable. This was necessary because we could see that especially the \glspl{gam} based on deep neural networks require high computational resources with long training times, leading to a total training time of more than $279$ hours (see Appendix~\ref{app:training_times}). However, for future work, it is worth investigating datasets with different characteristics (e.g., large collections with several million samples and many more features) in order to provide more evidence for our findings.

Second, we explicitly focused on prediction tasks with tabular data, which typically contain naturally meaningful features for interpretation purposes. In domains with higher-dimensional data, such as images and text, the results of this study may not be directly applicable. For this purpose, some upstream feature engineering methods are required that transform raw input data such as pixels or text snippets into higher-level features before feeding them into \glspl{gam} to produce interpretable shape functions.

Third, we considered only a limited set of hyperparameters and refrained from using a fully exhaustive tuning grid, which is also computationally infeasible. Thus, it is conceivable that both the interpretable as well as the black-box models may achieve even higher predictive performance for certain configurations and datasets. However, since we took care to vary influential parameters in all models and followed the recommendations of the respective model developers and authors, we are confident that our results based on $68,500$ training runs provide a fair and representative reflection of the true predictive qualities.

Fourth, the selection of interpretable models in our study was explicitly limited to the family of \glspl{gam}. This was necessary to define a standardized evaluation procedure and to ensure the comparability of the results. However, we would like to emphasize that there are even more types of intrinsically interpretable models, such as Bayesian models and rule-based learners \citep{barredo_arrieta_explainable_2020}, but also more advanced (hybrid) approaches, such as logit leaf models \citep{de_caigny_new_2018} and spline-rule ensembles \citep{de_bock_spline_rule_2021}. Extending our systematic model evaluation to such model families would be a promising direction to get an even more comprehensive picture of how the individual models are positioned within the machine learning landscape. Furthermore, it would be interesting to use ensembles of different \gls{gam} variants \citep{de_bock_ensemble_2010} to investigate how their predictive performance can be further improved while investigating the impact on interpretability aspects.

A final limitation concerns the evaluation of the models' degree of interpretability. For this purpose, we adopted a qualitative assessment framework based on six evaluation criteria, and scored each model by three experienced developers of this study in three consecutive cycles. This type of "internal" evaluation was necessary because it required several weeks of engagement with each individual model in order to fully grasp their particularities. As a result, it was possible to thoroughly examine and discuss the merits and limitations of each model, and to report our findings in a transparent and reproducible manner. At the same time, however, an external perspective is currently lacking that considers how external model users would assess the level of interpretability of the different models under realistic circumstances \citep{doshi2017towards}. To this end, it is planned to conduct case studies and field experiments with data science experts and decision makers from different domains and to evaluate the usefulness of the different models using real prediction problems.

In addition, we plan to expand the evaluation criteria of the qualitative evaluation framework, as it may not currently provide a fully representative picture of all necessary dimensions of interpretability. For example, it was noticeable that the decision tree received a low interpretability score even though it is largely classified as a highly interpretable model. At this point, other dimensions are conceivable, such as the convertibility of the model into easily understandable decision rules. Likewise, our evaluation framework currently assumes that all six interpretability criteria are equally important. However, depending on the specifics of the application domain, individual criteria may have a much higher weight, such as the requirement for sparsity in use cases where hundreds of features are involved (e.g., manufacturing, e-commerce) or the requirement for monotonicity in specific applications like credit scoring (see Appendix~\ref{app:trade-off}). Future work should therefore focus on the extension and context-specific weighting of interpretability criteria for different contexts and application domains.

Furthermore, we believe that it is worthwhile to invest more effort in the identification and examination of \gls{gam}-specific interpretability metrics (e.g., number of visual chunks, degree of smoothness, number of discontinuities), so that a more fine-grained interpretability assessment of different \gls{gam} variants is possible. Such metrics could also be combined with user-centric evaluation studies to find out which factors have the greatest impact on comprehensibility and usability.

\section{Conclusion}
\label{sec:conclusion}

The expanding influence of machine learning across various domains makes the adoption of intrinsically interpretable models mandatory, which is particularly true for high-stakes sectors such as healthcare, finance, and judiciary. Our research aimed to evaluate and compare modern variants of such intrinsically interpretable models, namely generalized additive models, with traditional interpretable as well as black-box models to dispel the myth that only black-box models offer strong prediction qualities.

Our systematic study involved an extensive analysis of seven \glspl{gam} and seven commonly used \gls{ml} models based on a large collection of twenty benchmark datasets and an extensive hyperparameter search. Our analysis revealed that advanced \glspl{gam} provide a promising alternative to black-box models, offering competitive predictive performance and inherent interpretability. Among the \glspl{gam} we evaluated, \gls{ebm} emerged as the leading model. It outperformed both conventional interpretable models, such as \gls{lr} and \gls{dt}, as well as most of the commonly used black-box models, including \gls{xgb}, \gls{rf}, and \gls{mlp}. This finding highlights the potential of \glspl{gam} as a preferable choice for high-stakes decision scenarios, in which transparency plays a pivotal role.

Despite these promising results, we acknowledge the limitations of our study, including the focus on medium-sized tabular datasets, a specific set of \gls{ml} models, and a manageable choice of hyperparameters. In addition, the assessment of interpretability was primarily based on a qualitative evaluation framework that may not capture all dimensions of interpretability. Other criteria may be relevant in different contexts or for different stakeholders. Therefore, future research should address these limitations to provide more nuanced perspectives on this highly relevant topic. Nevertheless, we are confident that our results provide important insights for the IS community, paving the way for a deeper understanding of the predictive performance, interpretability, and applicability of intrinsically interpretable models.


\setstretch{1.2} 
\bibliographystyle{apacite}
\bibliography{references}


\clearpage
\appendix
\gdef\thesection{\Alph{section}} 
\makeatletter
\renewcommand\@seccntformat[1]{Appendix \csname the#1\endcsname.\hspace{0.5em}}
\makeatother

\setcounter{page}{1}

\vspace{1cm}
\begin{center}
\Large Online Appendix
\end{center}
\vspace{0.5cm}


\section{Model Implementations}
\label{app:models}

We implement and evaluate all models in a Python environment. Decision tree, random forest, multi-layer perceptron, and linear/logistic regression are implemented using scikit-learn. Depending on the prediction task, we either use the classifier or the regressor version. In the case of the linear regression model for the regression task, we use the ElasticNet implementation with different regularization settings. For \gls{psplines}, we use the Python package pyGAM \citep{serven_pygam_2021}. For \gls{tpsplines}, we adopt models from the mgcv package implemented in R \citep{wood_mgcv_2023} using a Python wrapper \citep{chang_how_2021}. For \gls{igann}, a custom interaction selection method based on decision trees is used. For the remaining models, we use the implementations provided by the respective authors and/or developers that are available in public repositories. Table~\ref{tab:models_overview} provides an overview of the applied models and their implementations. 

To ensure a fair comparison, all models are treated equally and receive the same data input. This means, for example, that no model-specific feature encodings or individual feature transformations are applied beyond the preprocessing steps that are already carried out internally by the respective model implementations by default. An overview of the generically used preprocessing steps, which constitute the basis for all datasets and models, can be found in Appendix~\ref{app:datasets}.

Furthermore, all models are applied both with default configurations and with varied hyperparameter configurations. An overview of the default values and the selected hyperparameter values can be found in Appendix~\ref{app:hyperparameter}. 
While some models like \gls{ebm} or \gls{gaminet} implement early stopping by default, \gls{mlp} and TabNet do not. Therefore, we enable early stopping for both models, using 10\% of the training set as an internal validation set to avoid overfitting and unnecessarily long training times.

\begin{table}[ht]
\caption{Overview of applied models with implementations.}
\label{tab:models_overview}
\resizebox{\textwidth}{!}{%
\begin{tabular}{lll}
\toprule
\textbf{Model} & \textbf{Python \& R Implementation} & \textbf{Repository/URL}\\ \midrule
P-Splines & pyGAM[LogisticGAM/LinearGAM] (version 0.8.0) & https://github.com/dswah/pyGAM \\[5pt]
\begin{tabular}[c]{@{}l@{}}TP-Splines\\ \\ \end{tabular} & \begin{tabular}[c]{@{}l@{}}MyRSpline[MyRSplineClassifier/MyRSplineRegressor] (version n.d.)\\ R-Package: mgcv (version 1.9.1)\end{tabular} & \begin{tabular}[c]{@{}l@{}}https://github.com/zzzace2000/GAMs\_models/tree/master\\ https://cran.r-project.org/web/packages/mgcv/index.html\end{tabular} \\[10pt]
EBM      & {interpret.ExplainableBoosting[Classifier/Regressor]} (version 0.2.7) & https://github.com/interpretml/interpret \\[5pt]
NAM      & {nam pytorch version} (version 0.0.0) &  https://github.com/lemeln/nam\\[5pt]
GAMI-Net & {gaminet} (version 0.6.0) & https://github.com/SelfExplainML/PiML-Toolbox\\[5pt]
ExNN     & {exnn} (version 0.1) & https://github.com/SelfExplainML/ExNN\\[5pt]
IGANN    & {igann} (version pre-release) & https://github.com/MathiasKraus/igann\\[5pt]   
LR       & {sklearn.linear\_models[LogisticRegression/ElasticNet]} (version 1.1.2) & https://scikit-learn.org/stable/\\[5pt]
DT       & {sklearn.tree.DecisionTree[Classifier/Regressor]} (version 1.1.2) & https://scikit-learn.org/stable/\\[5pt]
RF       & {sklearn.ensemble.RandomForest[Classifier/Regressor]} (version 1.1.2) & https://scikit-learn.org/stable/\\[5pt]
XGB      & {xgboost.XGB[Classifier/Regressor]} (version 1.6.2) & https://github.com/dmlc/xgboost\\[5pt]
CatBoost & {catboost[CatBoostClassifier/CatBoostRegressor]} (version 1.2.3) & https://catboost.ai/en/docs/ \\[5pt]
MLP      & {sklearn.neural\_network.MLP[Classifier/Regressor]} (version 1.1.2) & https://scikit-learn.org/stable/\\[5pt]
TabNet & pytorch\_tabnet.tab\_model[TabNetClassifier/TabNetRegressor] (version 4.1.0) & https://github.com/dreamquark-ai/tabnet  \\
\bottomrule
\end{tabular}}
\end{table}


\newpage

\section{Datasets and Data Preprocessing}
\label{app:datasets}

\begin{table}[ht]
\caption{References to the repositories of the applied benchmark datasets.}
\label{tab:datasets_sources}

\sisetup{group-digits=true, group-separator={,}, group-minimum-digits=4, detect-all}
\resizebox{\textwidth}{!}{%
\begin{tabular}{lll}
\toprule
\multicolumn{1}{l}{\textbf{Type}} & \multicolumn{1}{l}{\textbf{Dataset}}& \multicolumn{1}{l}{\textbf{Repository}}\\
\hline
\multirow{10}{*}{CLS}
 & College \citep{college_source} & https://www.kaggle.com/datasets/saddamazyazy/go-to-college-dataset \\
 & Water potability \citep{water_source} & https://kaggle.com/adityakadiwal/water-potability \\
 & Stroke \citep{stroke_source} & https://kaggle.com/fedesoriano/stroke-prediction-dataset\\
 & Customer churn \citep{IBM} & https://kaggle.com/blastchar/telco-customer-churn\\
 & Recidivism \citep{larson_angwin_kirchner_mattu_2016}& https://www.kaggle.com/datasets/danofer/compass\\
 & Credit scoring \citep{fico_source}& https://community.fico.com/s/explainable-machine-learning-challenge \\
 & Income adults \citep{Kohavi.1996}& https://archive.ics.uci.edu/ml/datasets/adult\\
 & Bank marketing \citep{moro2014-driven_2014}& https://archive.ics.uci.edu/ml/datasets/Bank+Marketing \\
 & Airline satisfaction \citep{tj_klein_airline}& https://kaggle.com/teejmahal20/airline-passenger-satisfaction \\
 & Weather forecast \citep{weather_australia}& https://www.kaggle.com/datasets/jsphyg/weather-dataset-rattle-package\\ \\
\multirow{10}{*}{REG} 
 & Car price \citep{kibler_instance-based_1989}& https://archive.ics.uci.edu/ml/datasets/automobile \\
 & Student grade \citep{cortez_using_2008}& https://archive.ics.uci.edu/ml/datasets/Student+Performance\\
 & Productivity \citep{Imran_2019}& https://archive.ics.uci.edu/ml/datasets/Productivity+Prediction+of+Garment+Employees\\
 & Medical insurance \citep{lantz_2015}& https://www.kaggle.com/datasets/mirichoi0218/insurance\\
 & Violent Crimes \citep{redmond_data-driven_2002} & https://archive.ics.uci.edu/ml/datasets/Communities+and+Crime\\
 & Crab farming \citep{gursewak_singh_sidhu_2021}& https://www.kaggle.com/datasets/sidhus/crab-age-prediction\\
 & Wine quality \citep{cortez_2009}& https://archive.ics.uci.edu/ml/datasets/wine+quality\\
 & Bike rental \citep{fanaee-t_event_2014}& https://archive.ics.uci.edu/ml/datasets/Bike+Sharing+Dataset \\
 & House price \citep{pace_sparse_1997}& https://www.kaggle.com/datasets/camnugent/california-housing-prices\\
 & Diamond price \citep{garside_2021}& https://www.kaggle.com/datasets/nancyalaswad90/diamonds-prices \\ \bottomrule
 \end{tabular}%
}
\end{table}

\noindent All 20 datasets are loaded into a shared computing environment using a unified evaluation pipeline. After loading the datasets, we apply standard approaches for data preprocessing. More specifically, we apply the following steps:

\begin{enumerate}[(1)]
    \item Remove IDs and features that cause obvious data leakage.
    \item Replace empty strings in the numerical features with "np.nan" to ensure consistency in handling missing values.
    \item Remove samples that have no target variable or consist only of missing feature values.
    \item Remove features with more than $50 \%$ missing values.
    \item Clean remaining missing values via median value imputation for numerical features and NA-encoding for categorical features.
    \item Remove categorical features with more than $25$ distinct values to reduce computational complexity arising from high cardinality.
    \item Apply one-hot encoding for categorical features.
    \item Standardize continuous features as well as numerical target variables by removing the mean and scaling to unit variance.
    \item Convert binary target variables into a standardized format ($y \in \{1,0\}$). 
\end{enumerate}

\noindent Apart from that, we keep the datasets in their default structure without further cleaning and feature transformations to ensure a fair model comparison. Table~\ref{tab:preprocessing-impact-table} provides an overview of the dataset characteristics after applying the described preprocessing steps and one-hot encoding for categorical features.

\begin{table}[ht]
\caption{Overview of dataset characteristics with and without preprocessing and encoding.}
\centering
\label{tab:preprocessing-impact-table}
\sisetup{group-digits=true, group-separator={,}, group-minimum-digits=4}
\resizebox{0.8\columnwidth}{!}{%
\begin{tabular}{llS[table-format=6.0]rrS[table-format=6.0]rrrrrrrr}
\toprule
               & & \multicolumn{3}{l}{\textbf{No Preprocessing}} & \multicolumn{6}{l}{\textbf{Preprocessing}} & \multicolumn{3}{l}{\textbf{Encoded}} \\
               \cmidrule(l){3-5} \cmidrule(l){6-11} \cmidrule(l){12-14}
               
          \textbf{Task} & \textbf{Dataset Name} &          \textbf{Samples} & \multicolumn{2}{l}{\textbf{Columns}} &       \multicolumn{2}{l}{\textbf{Samples}} & \multicolumn{4}{l}{\textbf{Columns}} & \multicolumn{3}{l}{\textbf{Columns}} \\
          \cmidrule(l){3-3} \cmidrule(l){4-5} \cmidrule(l){6-7} \cmidrule(l){8-11} \cmidrule(l){12-14}
          
               & & &     \textbf{num} & \multicolumn{2}{l}{\textbf{cat}} & &       \multicolumn{2}{l}{\textbf{num}} &     \multicolumn{2}{l}{\textbf{cat}}  &     \textbf{num} &      \multicolumn{2}{l}{\textbf{cat}}\\
\midrule
{\multirow{10}{*}{CLS}}
 &      college &             1000 &       4 &   6 &      1000 & (0) &     4 &(0) &   6 &(0) &       4 &  10 &(+4) \\
 &        water &             3276 &       9 &   0 &      3276  &(0) &     9 & (0) &   0 &(0) &       9 &    0 &(0) \\
 &       stroke &             5110 &       4 &   7 &      5110 & (0) &    3 &(-1) &   7 &(0) &       3 &  16 &(+9) \\
 &        churn &             7043 &       3 &  17 &      7043 & (0) &     3 &(0) & 16 &(-1) &       3 & 37 &(+21) \\
 &   recidivism &             7214 &      31 &  22 &      7214 & (0) &   7 &(-24) & 4 &(-18) &       7 &  11 &(+7) \\
 &       credit &            10459 &      21 &   2 &   9871 & (-588) &   21 &(0) &  2 &(0) &      21 & 16 &(+14) \\
 &       income &            32561 &       6 &   8 &     32561 &  (0) &     6 & (0) &  7& (-1) &       6 & 59 &(+52) \\
 &         bank &            45211 &      10 &  10 &     45211 & (0) &    6 &(-4) &  9 &(-1) &       6 & 41 &(+32) \\
 &      airline &           103904 &      20 &   4 &    103904 & (0) &   18 &(-2) &   4 &(0) &      18 &   6 &(+2) \\
 &      weather &           142193 &      18 &   5 &    142193 & (0) &   16 &(-2) &  4 &(-1) &      16 & 54 &(+50) \\
 \\
 {\multirow{10}{*}{REG}}
     &          car &              205 &      14 &  11 &       201 & (-4) &   13 & (-1) &  11 &(0) &      13 & 63& (+52) \\
     &      student &              649 &      13 &  17 &       649 & (0) &    13 & (0) &  17 &(0) &      13 & 30 &(+13) \\
     & productivity &             1197 &      10 &   4 &      1197 & (0) &    9 & (-1) &   4 &(0) &       9 & 26& (+22) \\
     &    insurance &             1338 &       3 &   3 &      1338 & (0) &     3 &(0) &   3 & (0) &       3 &   6 &(+3) \\
     &       crimes &             1994 &     125 &   2 &      1994 & (0) & 100 &(-25) &  0 &(-2) &     100 &    0 &(0) \\
     &      farming &             3893 &       7 &   1 &      3893 & (0) &     7 &(0) &   1 &(0) &       7 &   3 &(+2) \\
     &         wine &             4898 &      11 &   0 &      4898 & (0) &    11 &(0) &   0 &(0) &      11 &    0 &(0) \\
     &         bike &            17379 &      11 &   5 &     17379 & (0) &    2 &(-9) &  1 &(-4) &       2 &   4 &(+3) \\
     &        house &            20640 &       8 &   0 &     20640 & (0) &     8 &(0) &   0 & (0) &       8 &    0 &(0) \\
     &      diamond &            53943 &       7 &   3 &     53943&  (0) &    6 &(-1) &   3&  (0) &       6 & 20 &(+17) \\
\bottomrule
\end{tabular}
}
\end{table}

\newpage
\clearpage

\section{Hyperparameter Tuning}
\label{app:hyperparameter}

\begin{table}[htp!]
\caption{Hyperparameter tuning grid used for the model evaluation.}
\label{tab:hyperparameter_grid}
\centering
\resizebox{0.95\linewidth}{!}{%
\begin{tabular}{llll}
\hline
\textbf{Model} & \textbf{Hyperparameter} & \textbf{Hyperparameter Values for Grid Search} & \textbf{Default Value}           \\ \hline
\multirow{3}{*}{P-Splines}  & \textit{basis} & "ps" & "ps"\\       
                     & \textit{n\_splines} & 5, 10, 15, 20, 25 & 20\\
                     & \textit{lam}                            & 0.2, 0.4, 0.6, 0.9 & 0.6\\ \hline
\multirow{4}{*}{TP-Splines}  & \textit{spline\_type} & "ts" & "ts"                      \\
                     & \textit{maxk} & 5, 10, 20 & 10                     \\
                     & \textit{m} & 2, 3 & 2                        \\
                     & \textit{gamma} & 1, 1.2, 1.4 & 1                         \\ \hline
\multirow{4}{*}{EBM}                  & \textit{max\_bins} & 256, 512 & 256                      \\
                     & \textit{interactions} & 0, 10, 20 & 10                     \\
                     & \textit{outer\_bags} & 8,16 & 8                         \\
                     & \textit{inner\_bags} & 0, 4 & 0                         \\ \hline
\multirow{4}{*}{NAM}                  & \textit{lr} & 1e-2, 2.082e-2  & 2.082e-2 \\
                     & \textit{num\_learners} & 1, 5, 8 & 1 \\
                     & \textit{dropout}                        & 0, 0.1      &0.1                  \\
                     & \textit{num\_basis\_functions} & 1, 64   & 64              \\ \hline
\multirow{3}{*}{GAMI-Net}              & \textit{interact\_num} & 0, 10, 20 & 20                     \\
                     & \textit{activation\_func} & "relu", "tanh", "sigmoid" & "relu"                \\
                     & \textit{reg\_clarity} & 0.01, 0.1, 0.2 & 0.1                \\ \hline
\multirow{3}{*}{EXNN}                 & \textit{subnet\_num} & 5, 10 & 5                         \\
                     & \textit{l1\_proj} & 0.01, 0.001, 0.0001 & 0.001           \\
                     & \textit{l1\_subnet} & 0.01, 0.001, 0.0001 & 0.001           \\ \hline
\multirow{3}{*}{IGANN} & \textit{boost\_rate} & 0.025, 0.1 & 0.1      \\
                     & \textit{elm\_scale} & 1, 2, 5 & 1\\
                     & \textit{interactions} & 0, 10, 20 & 0\\ \hline
\multirow{6}{*}{\begin{tabular}{@{}l@{}}LR - Logistic \\ (Classification)\end{tabular}} & \textit{C} & 0.001, 0.01, 0.1, 1, 10, 100, 1000  & 1.0 \\
                     & \textit{penalty} & "l1", "l2", "elasticnet", None & "l2"               \\ 
                     & \textit{class\_weight}                      & "balanced", None & None \\
                     & \textit{solver} & "lbfgs", "liblinear", "saga" & "lbfgs" \\
                     & \textit{l1\_ratio} & 0.25, 0.5, 0.75, None & None \\
                     & \textit{max\_iter} & 100, 300 & 100 \\
                     \hline
\multirow{2}{*}{\begin{tabular}{@{}l@{}}LR - ElasticNet\\ (Regression)\end{tabular}} & \textit{alpha} & 0.001, 0.01, 0.1, 1, 10, 100, 1000 & 1 \\
                    & \textit{l1\_ratio} & 0, 0.1, 0.2, 0.3, 0.4, 0.5, 0.6, 0.7, 0.8, 0.9, 1 & 0\\ \hline
\multirow{4}{*}{DT} & \textit{max\_depth} & 5, 10, 20, 40, None & None\\
                     & \textit{max\_leaf\_nodes} & None, 5, 10, 20, 40 & None\\
                     & \textit{class\_weight} & "balanced", None & None\\
                     & \textit{splitter} & "best", "random" & "best" \\ \hline
\multirow{3}{*}{RF} & \textit{n\_estimators} & 50, 100, 200, 500, 1000 & 100 \\
                     & \textit{max\_depth} & 5, 10, 20, 40, None & None\\
                     & class\_weight                  & "balanced", None & None \\ \hline
\multirow{3}{*}{XGB} & \textit{max\_depth} & 3, 6, 9, 12, None & 6             \\
                     & \textit{learning\_rate} & 0.01, 0.1, 0.3 & 0.3 \\
                     & \textit{n\_estimators} & 50, 100, 200, 500, 1000, 2000 & 100\\ \hline
\multirow{3}{*}{CatBoost} & \textit{max\_depth} & 3, 6, 9, 12 & 6             \\
                     & \textit{eta} & 0.01, 0.03, 0.1, 0.3 & 0.03 \\
                     & \textit{n\_estimators} & 50, 100, 200, 500, 1000 & 1000 \\ \hline
\multirow{4}{*}{MLP} &
  \textit{hidden\_layer\_sizes} &
  \begin{tabular}[c]{@{}l@{}}{[}50{]}, {[}100{]}, {[}25, 25{]}, {[}50, 50{]}, {[}75, 75{]}, {[}100, 100{]}, \\ {[}25, 25, 25{]}, {[}50, 50, 50{]}, {[}75, 75, 75{]}, {[}100, 100, 100{]}, \\ {[}50, 50, 50, 50{]}, {[}100, 100, 100, 100{]}\end{tabular} &  {[}100{]}\\
                     & \textit{alpha} & 0.0001, 0.001, 0.01 & 0.0001\\
                     & \textit{activation} & "relu", "tanh" & "relu" \\ \hline
\multirow{3}{*}{TabNet} & \textit{n\_a\_and\_d} & 8, 16, 32 & 8             \\
                     & \textit{n\_steps} & 3, 5, 10 & 3 \\
                     & \textit{gamma} & 1.3, 1.5, 1.8 & 1.3 \\ \hline
\end{tabular}
}
\end{table}
\clearpage
\newpage


\newpage
\clearpage

\noindent Table~\ref{tab:hpo_best_config_fold_majority_gams} and Table~\ref{tab:hpo_best_config_fold_majority_iml_blackbox} summarize the best hyperparameter configurations per dataset and model.

\begin{table}[htp!]
\centering
\caption{Best hyperparameter configurations across all five folds of the 5-fold cross-validation (determined by majority voting) for the generalized additive models. Parameters for which the majority vote resulted in a tie are marked with *. For these, the first emerging parameter value is given.}
\label{tab:hpo_best_config_fold_majority_gams}
\resizebox*{!}{0.85\textheight}{%
\begin{tabular}{@{}llllllll@{}}
\toprule
& \multicolumn{7}{c}{\textbf{Generalized Additive Models}} \\ \cmidrule(l){2-8}
\textbf{Task}& \textbf{P-Splines} & \textbf{TP-Splines} & \textbf{EBM} & \textbf{NAM} & \textbf{GAMI-Net} & \textbf{ExNN} & \textbf{IGANN} \\ \midrule
college       & \begin{tabular}[c]{@{}l@{}}'n\_splines*': 5,\\ 'lam': 0.2\end{tabular}  & \begin{tabular}[c]{@{}l@{}}'spline\_type': 'ts',\\ 'maxk': 5,\\ 'm': 3,\\ 'gamma': 1\end{tabular}  & \begin{tabular}[c]{@{}l@{}}'max\_bins': 512,\\ 'interactions': 20,\\ 'outer\_bags': 8,\\ 'inner\_bags': 4\end{tabular}  & \begin{tabular}[c]{@{}l@{}}'lr': 0.01,\\ 'num\_learners': 5,\\ 'dropout': 0,\\ 'num\_basis\_functions': 64\end{tabular}    & \begin{tabular}[c]{@{}l@{}}'interact\_num': 20,\\ 'activation\_func': 'ReLU',\\ 'reg\_clarity*': 0.01\end{tabular}    & \begin{tabular}[c]{@{}l@{}}'subnet\_num': 10,\\ 'l1\_proj': 0.01,\\ 'l1\_subnet': 0.001\end{tabular}    & \begin{tabular}[c]{@{}l@{}}'boost\_rate': 0.1,\\ 'elm\_scale': 1,\\ 'interactions': 20\end{tabular}    \\[35pt]
water         & \begin{tabular}[c]{@{}l@{}}'n\_splines': 5,\\ 'lam': 0.2\end{tabular}   & \begin{tabular}[c]{@{}l@{}}'spline\_type': 'ts',\\ 'maxk': 5,\\ 'm': 3,\\ 'gamma': 1.2\end{tabular} & \begin{tabular}[c]{@{}l@{}}'max\_bins': 512,\\ 'interactions': 10,\\ 'outer\_bags': 8,\\ 'inner\_bags': 4\end{tabular}  & \begin{tabular}[c]{@{}l@{}}'lr': 0.01,\\ 'num\_learners*': 1,\\ 'dropout': 0,\\ 'num\_basis\_functions': 64\end{tabular}   & \begin{tabular}[c]{@{}l@{}}'interact\_num': 20,\\ 'activation\_func': 'ReLU',\\ 'reg\_clarity*': 0.1\end{tabular}     & \begin{tabular}[c]{@{}l@{}}'subnet\_num': 5,\\ 'l1\_proj': 0.01,\\ 'l1\_subnet*': 0.001\end{tabular}    & \begin{tabular}[c]{@{}l@{}}'boost\_rate': 0.1,\\ 'elm\_scale': 1,\\ 'interactions': 20\end{tabular}    \\[35pt]
stroke        & \begin{tabular}[c]{@{}l@{}}'n\_splines': 5,\\ 'lam': 0.2\end{tabular}   & \begin{tabular}[c]{@{}l@{}}'spline\_type': 'ts',\\ 'maxk': 5,\\ 'm': 2,\\ 'gamma': 1\end{tabular}  & \begin{tabular}[c]{@{}l@{}}'max\_bins': 256,\\ 'interactions': 0,\\ 'outer\_bags': 8,\\ 'inner\_bags': 0\end{tabular}   & \begin{tabular}[c]{@{}l@{}}'lr': 0.01,\\ 'num\_learners': 1,\\ 'dropout': 0,\\ 'num\_basis\_functions': 1\end{tabular}     & \begin{tabular}[c]{@{}l@{}}'interact\_num': 0,\\ 'activation\_func': 'ReLU',\\ 'reg\_clarity': 0.01\end{tabular}      & \begin{tabular}[c]{@{}l@{}}'subnet\_num': 5,\\ 'l1\_proj': 0.01,\\ 'l1\_subnet': 0.01\end{tabular}      & \begin{tabular}[c]{@{}l@{}}'boost\_rate': 0.025,\\ 'elm\_scale': 1,\\ 'interactions': 0\end{tabular}   \\[35pt]
churn         & \begin{tabular}[c]{@{}l@{}}'n\_splines': 5,\\ 'lam': 0.4\end{tabular}   & \begin{tabular}[c]{@{}l@{}}'spline\_type': 'ts',\\ 'maxk*': 5,\\ 'm': 2,\\ 'gamma': 1.4\end{tabular} & \begin{tabular}[c]{@{}l@{}}'max\_bins': 512,\\ 'interactions': 20,\\ 'outer\_bags': 16,\\ 'inner\_bags': 4\end{tabular} & \begin{tabular}[c]{@{}l@{}}'lr': 0.01,\\ 'num\_learners*': 5,\\ 'dropout': 0,\\ 'num\_basis\_functions': 64\end{tabular}   & \begin{tabular}[c]{@{}l@{}}'interact\_num*': 20,\\ 'activation\_func': 'ReLU',\\ 'reg\_clarity*': 0.01\end{tabular}   & \begin{tabular}[c]{@{}l@{}}'subnet\_num': 5,\\ 'l1\_proj*': 0.0001,\\ 'l1\_subnet': 0.0001\end{tabular} & \begin{tabular}[c]{@{}l@{}}'boost\_rate': 0.025,\\ 'elm\_scale': 2,\\ 'interactions': 10\end{tabular}  \\[35pt]
recidivism    & \begin{tabular}[c]{@{}l@{}}'n\_splines': 5,\\ 'lam': 0.2\end{tabular}   & \begin{tabular}[c]{@{}l@{}}'spline\_type': 'ts',\\ 'maxk*': 10,\\ 'm': 3,\\ 'gamma*': 1.4\end{tabular}  & \begin{tabular}[c]{@{}l@{}}'max\_bins': 256,\\ 'interactions': 10,\\ 'outer\_bags': 8,\\ 'inner\_bags': 4\end{tabular}  & \begin{tabular}[c]{@{}l@{}}'lr': 0.01,\\ 'num\_learners': 8,\\ 'dropout': 0,\\ 'num\_basis\_functions': 64\end{tabular}    & \begin{tabular}[c]{@{}l@{}}'interact\_num*': 0,\\ 'activation\_func': 'ReLU',\\ 'reg\_clarity': 0.01\end{tabular}     & \begin{tabular}[c]{@{}l@{}}'subnet\_num': 5,\\ 'l1\_proj': 0.0001,\\ 'l1\_subnet': 0.001\end{tabular}   & \begin{tabular}[c]{@{}l@{}}'boost\_rate': 0.1,\\ 'elm\_scale': 5,\\ 'interactions': 0\end{tabular}     \\[35pt]
credit        & \begin{tabular}[c]{@{}l@{}}'n\_splines': 10,\\ 'lam': 0.6\end{tabular}  & \begin{tabular}[c]{@{}l@{}}'spline\_type': 'ts',\\ 'maxk': 10,\\ 'm': 3,\\ 'gamma': 1.4\end{tabular} & \begin{tabular}[c]{@{}l@{}}'max\_bins': 512,\\ 'interactions': 0,\\ 'outer\_bags': 8,\\ 'inner\_bags': 4\end{tabular}   & \begin{tabular}[c]{@{}l@{}}'lr': 0.01,\\ 'num\_learners*': 5,\\ 'dropout': 0.1,\\ 'num\_basis\_functions': 1\end{tabular}  & \begin{tabular}[c]{@{}l@{}}'interact\_num': 20,\\ 'activation\_func': 'ReLU',\\ 'reg\_clarity*': 0.01\end{tabular}    & \begin{tabular}[c]{@{}l@{}}'subnet\_num': 10,\\ 'l1\_proj': 0.01,\\ 'l1\_subnet*': 0.0001\end{tabular}  & \begin{tabular}[c]{@{}l@{}}'boost\_rate': 0.025,\\ 'elm\_scale': 1,\\ 'interactions': 0\end{tabular}   \\[35pt]
income        & \begin{tabular}[c]{@{}l@{}}'n\_splines': 25,\\ 'lam': 0.2\end{tabular}  & \begin{tabular}[c]{@{}l@{}}'spline\_type': 'ts',\\ 'maxk': 20,\\ 'm': 2,\\ 'gamma*': 1\end{tabular} & \begin{tabular}[c]{@{}l@{}}'max\_bins': 512,\\ 'interactions': 20,\\ 'outer\_bags': 8,\\ 'inner\_bags': 4\end{tabular}  & \begin{tabular}[c]{@{}l@{}}'lr': 0.01,\\ 'num\_learners': 8,\\ 'dropout': 0,\\ 'num\_basis\_functions': 64\end{tabular}    & \begin{tabular}[c]{@{}l@{}}'interact\_num': 10,\\ 'activation\_func': 'ReLU',\\ 'reg\_clarity': 0.01\end{tabular}     & \begin{tabular}[c]{@{}l@{}}'subnet\_num': 10,\\ 'l1\_proj': 0.01,\\ 'l1\_subnet': 0.01\end{tabular}     & \begin{tabular}[c]{@{}l@{}}'boost\_rate': 0.1,\\ 'elm\_scale': 2,\\ 'interactions': 10\end{tabular}    \\[35pt]
bank          & \begin{tabular}[c]{@{}l@{}}'n\_splines*': 10,\\ 'lam': 0.6\end{tabular} & \begin{tabular}[c]{@{}l@{}}'spline\_type': 'ts',\\ 'maxk*': 20,\\ 'm': 2,\\ 'gamma': 1.2\end{tabular} & \begin{tabular}[c]{@{}l@{}}'max\_bins': 256,\\ 'interactions': 10,\\ 'outer\_bags': 8,\\ 'inner\_bags': 0\end{tabular}  & \begin{tabular}[c]{@{}l@{}}'lr': 0.01,\\ 'num\_learners*': 1,\\ 'dropout': 0,\\ 'num\_basis\_functions': 64\end{tabular}   & \begin{tabular}[c]{@{}l@{}}'interact\_num': 10,\\ 'activation\_func': 'Sigmoid',\\ 'reg\_clarity': 0.01\end{tabular}  & \begin{tabular}[c]{@{}l@{}}'subnet\_num': 5,\\ 'l1\_proj': 0.01,\\ 'l1\_subnet': 0.01\end{tabular}      & \begin{tabular}[c]{@{}l@{}}'boost\_rate': 0.025,\\ 'elm\_scale': 1,\\ 'interactions*': 20\end{tabular} \\[35pt]
airline       & \begin{tabular}[c]{@{}l@{}}'n\_splines*': 15,\\ 'lam': 0.4\end{tabular} & \begin{tabular}[c]{@{}l@{}}'spline\_type': 'ts',\\ 'maxk': 10,\\ 'm': 2,\\ 'gamma': 1.4\end{tabular} & \begin{tabular}[c]{@{}l@{}}'max\_bins': 512,\\ 'interactions': 20,\\ 'outer\_bags': 8,\\ 'inner\_bags': 4\end{tabular}  & \begin{tabular}[c]{@{}l@{}}'lr': 0.02082,\\ 'num\_learners': 8,\\ 'dropout': 0,\\ 'num\_basis\_functions': 64\end{tabular} & \begin{tabular}[c]{@{}l@{}}'interact\_num': 20,\\ 'activation\_func': 'ReLU',\\ 'reg\_clarity': 0.01\end{tabular}     & \begin{tabular}[c]{@{}l@{}}'subnet\_num': 10,\\ 'l1\_proj': 0.001,\\ 'l1\_subnet': 0.001\end{tabular}   & \begin{tabular}[c]{@{}l@{}}'boost\_rate': 0.1,\\ 'elm\_scale*': 2,\\ 'interactions': 20\end{tabular}   \\[35pt]
weather       & \begin{tabular}[c]{@{}l@{}}'n\_splines': 20,\\ 'lam': 0.2\end{tabular}  & \begin{tabular}[c]{@{}l@{}}'spline\_type': 'ts',\\ 'maxk': 20,\\ 'm': 3,\\ 'gamma': 1.2\end{tabular} & \begin{tabular}[c]{@{}l@{}}'max\_bins': 256,\\ 'interactions': 20,\\ 'outer\_bags': 16,\\ 'inner\_bags': 0\end{tabular} & \begin{tabular}[c]{@{}l@{}}'lr': 0.01,\\ 'num\_learners': 8,\\ 'dropout': 0,\\ 'num\_basis\_functions': 64\end{tabular}    & \begin{tabular}[c]{@{}l@{}}'interact\_num': 20,\\ 'activation\_func': 'ReLU',\\ 'reg\_clarity': 0.01\end{tabular}     & \begin{tabular}[c]{@{}l@{}}'subnet\_num': 5,\\ 'l1\_proj': 0.01,\\ 'l1\_subnet': 0.01\end{tabular}      & \begin{tabular}[c]{@{}l@{}}'boost\_rate': 0.1,\\ 'elm\_scale*': 1,\\ 'interactions': 20\end{tabular}   \\[35pt] \midrule
car           & \begin{tabular}[c]{@{}l@{}}'n\_splines': 5,\\ 'lam': 0.2\end{tabular}   & \begin{tabular}[c]{@{}l@{}}'spline\_type': 'ts',\\ 'maxk*': 10,\\ 'm': 2,\\ 'gamma': 1.4\end{tabular}  & \begin{tabular}[c]{@{}l@{}}'max\_bins': 256,\\ 'interactions': 20,\\ 'outer\_bags': 8,\\ 'inner\_bags': 0\end{tabular}  & \begin{tabular}[c]{@{}l@{}}'lr': 0.01,\\ 'num\_learners': 8,\\ 'dropout': 0,\\ 'num\_basis\_functions': 64\end{tabular}    & \begin{tabular}[c]{@{}l@{}}'interact\_num*': 10,\\ 'activation\_func*': 'Tanh',\\ 'reg\_clarity': 0.01\end{tabular}   & \begin{tabular}[c]{@{}l@{}}'subnet\_num': 5,\\ 'l1\_proj': 0.001,\\ 'l1\_subnet*': 0.0001\end{tabular}  & \begin{tabular}[c]{@{}l@{}}'boost\_rate': 0.1,\\ 'elm\_scale*': 1,\\ 'interactions*': 10\end{tabular}  \\[35pt]
student       & \begin{tabular}[c]{@{}l@{}}'n\_splines': 5,\\ 'lam': 0.9\end{tabular}   & \begin{tabular}[c]{@{}l@{}}'spline\_type': 'ts',\\ 'maxk': 5,\\ 'm': 2,\\ 'gamma': 1.4\end{tabular}  & \begin{tabular}[c]{@{}l@{}}'max\_bins': 256,\\ 'interactions': 0,\\ 'outer\_bags': 8,\\ 'inner\_bags': 4\end{tabular} & \begin{tabular}[c]{@{}l@{}}'lr': 0.02082,\\ 'num\_learners': 8,\\ 'dropout': 0,\\ 'num\_basis\_functions': 64\end{tabular} & \begin{tabular}[c]{@{}l@{}}'interact\_num': 0,\\ 'activation\_func*': 'ReLU',\\ 'reg\_clarity': 0.01\end{tabular} & \begin{tabular}[c]{@{}l@{}}'subnet\_num': 5,\\ 'l1\_proj': 0.001,\\ 'l1\_subnet': 0.001\end{tabular}     & \begin{tabular}[c]{@{}l@{}}'boost\_rate': 0.025,\\ 'elm\_scale': 1,\\ 'interactions*': 20\end{tabular}     \\[35pt]
productivity  & \begin{tabular}[c]{@{}l@{}}'n\_splines': 20,\\ 'lam': 0.4\end{tabular}   & \begin{tabular}[c]{@{}l@{}}'spline\_type': 'ts',\\ 'maxk': 20,\\ 'm': 2,\\ 'gamma': 1.4\end{tabular} & \begin{tabular}[c]{@{}l@{}}'max\_bins': 512,\\ 'interactions': 10,\\ 'outer\_bags': 16,\\ 'inner\_bags': 4\end{tabular}   & \begin{tabular}[c]{@{}l@{}}'lr': 0.02082,\\ 'num\_learners': 8,\\ 'dropout': 0,\\ 'num\_basis\_functions': 64\end{tabular}    & \begin{tabular}[c]{@{}l@{}}'interact\_num': 20,\\ 'activation\_func': 'Tanh',\\ 'reg\_clarity': 0.01\end{tabular}     & \begin{tabular}[c]{@{}l@{}}'subnet\_num': 5,\\ 'l1\_proj': 0.01,\\ 'l1\_subnet*': 0.0001\end{tabular}    & \begin{tabular}[c]{@{}l@{}}'boost\_rate': 0.1,\\ 'elm\_scale': 5,\\ 'interactions': 20\end{tabular} \\[35pt]
insurance     & \begin{tabular}[c]{@{}l@{}}'n\_splines': 5,\\ 'lam*': 0.9\end{tabular}  & \begin{tabular}[c]{@{}l@{}}'spline\_type': 'ts',\\ 'maxk': 20,\\ 'm': 2,\\ 'gamma': 1\end{tabular} & \begin{tabular}[c]{@{}l@{}}'max\_bins': 256,\\ 'interactions': 10,\\ 'outer\_bags': 16,\\ 'inner\_bags': 0\end{tabular} & \begin{tabular}[c]{@{}l@{}}'lr': 0.01,\\ 'num\_learners': 8,\\ 'dropout': 0,\\ 'num\_basis\_functions': 64\end{tabular} & \begin{tabular}[c]{@{}l@{}}'interact\_num': 10,\\ 'activation\_func': 'ReLU',\\ 'reg\_clarity*': 0.2\end{tabular}     & \begin{tabular}[c]{@{}l@{}}'subnet\_num': 5,\\ 'l1\_proj': 0.01,\\ 'l1\_subnet': 0.001\end{tabular}   & \begin{tabular}[c]{@{}l@{}}'boost\_rate': 0.1,\\ 'elm\_scale*': 2,\\ 'interactions': 10\end{tabular}    \\[35pt]
crimes        & \begin{tabular}[c]{@{}l@{}}'n\_splines': 5,\\ 'lam*': 0.9\end{tabular} & \begin{tabular}[c]{@{}l@{}}'spline\_type': 'ts',\\ 'maxk': 5,\\ 'm': 3,\\ 'gamma': 1.4\end{tabular}  & \begin{tabular}[c]{@{}l@{}}'max\_bins': 512,\\ 'interactions': 20,\\ 'outer\_bags': 16,\\ 'inner\_bags': 4\end{tabular} & \begin{tabular}[c]{@{}l@{}}'lr': 0.02082,\\ 'num\_learners': 8,\\ 'dropout': 0,\\ 'num\_basis\_functions': 64\end{tabular}    & \begin{tabular}[c]{@{}l@{}}'interact\_num*': 10,\\ 'activation\_func': 'Sigmoid',\\ 'reg\_clarity': 0.01\end{tabular}     & \begin{tabular}[c]{@{}l@{}}'subnet\_num': 5,\\ 'l1\_proj': 0.01,\\ 'l1\_subnet': 0.001\end{tabular}    & \begin{tabular}[c]{@{}l@{}}'boost\_rate': 0.1,\\ 'elm\_scale': 1,\\ 'interactions': 0\end{tabular}    \\[35pt]
farming       & \begin{tabular}[c]{@{}l@{}}'n\_splines': 5,\\ 'lam': 0.9\end{tabular}   & \begin{tabular}[c]{@{}l@{}}'spline\_type': 'ts',\\ 'maxk': 5,\\ 'm': 2,\\ 'gamma*': 1.4\end{tabular}  & \begin{tabular}[c]{@{}l@{}}'max\_bins': 512,\\ 'interactions': 10,\\ 'outer\_bags': 16,\\ 'inner\_bags': 4\end{tabular} & \begin{tabular}[c]{@{}l@{}}'lr': 0.01,\\ 'num\_learners': 8,\\ 'dropout': 0,\\ 'num\_basis\_functions': 64\end{tabular}    & \begin{tabular}[c]{@{}l@{}}'interact\_num': 10,\\ 'activation\_func': 'ReLU',\\ 'reg\_clarity': 0.01\end{tabular}     & \begin{tabular}[c]{@{}l@{}}'subnet\_num': 5,\\ 'l1\_proj': 0.001,\\ 'l1\_subnet*': 0.01\end{tabular}    & \begin{tabular}[c]{@{}l@{}}'boost\_rate': 0.1,\\ 'elm\_scale': 1,\\ 'interactions': 20\end{tabular}    \\[35pt]
wine          & \begin{tabular}[c]{@{}l@{}}'n\_splines': 25,\\ 'lam': 0.2\end{tabular}  & \begin{tabular}[c]{@{}l@{}}'spline\_type': 'ts',\\ 'maxk': 20,\\ 'm': 2,\\ 'gamma': 1\end{tabular} & \begin{tabular}[c]{@{}l@{}}'max\_bins': 256,\\ 'interactions': 20,\\ 'outer\_bags': 16,\\ 'inner\_bags': 0\end{tabular} & \begin{tabular}[c]{@{}l@{}}'lr': 0.02082,\\ 'num\_learners': 8,\\ 'dropout': 0,\\ 'num\_basis\_functions': 64\end{tabular} & \begin{tabular}[c]{@{}l@{}}'interact\_num': 20,\\ 'activation\_func': 'ReLU',\\ 'reg\_clarity': 0.01\end{tabular}     & \begin{tabular}[c]{@{}l@{}}'subnet\_num': 10,\\ 'l1\_proj': 0.01,\\ 'l1\_subnet': 0.001\end{tabular}    & \begin{tabular}[c]{@{}l@{}}'boost\_rate': 0.1,\\ 'elm\_scale*': 5,\\ 'interactions': 20\end{tabular}   \\[35pt]
bike          & \begin{tabular}[c]{@{}l@{}}'n\_splines': 25,\\ 'lam*': 0.2\end{tabular}  & \begin{tabular}[c]{@{}l@{}}'spline\_type': 'ts',\\ 'maxk': 20,\\ 'm': 2,\\ 'gamma': 1\end{tabular} & \begin{tabular}[c]{@{}l@{}}'max\_bins': 256,\\ 'interactions': 20,\\ 'outer\_bags': 16,\\ 'inner\_bags': 4\end{tabular} & \begin{tabular}[c]{@{}l@{}}'lr': 0.01,\\ 'num\_learners': 8,\\ 'dropout': 0,\\ 'num\_basis\_functions': 64\end{tabular}    & \begin{tabular}[c]{@{}l@{}}'interact\_num': 20,\\ 'activation\_func': 'ReLU',\\ 'reg\_clarity': 0.01\end{tabular}     & \begin{tabular}[c]{@{}l@{}}'subnet\_num': 5,\\ 'l1\_proj': 0.0001,\\ 'l1\_subnet': 0.01\end{tabular} & \begin{tabular}[c]{@{}l@{}}'boost\_rate': 0.1,\\ 'elm\_scale': 5,\\ 'interactions': 20\end{tabular}    \\[35pt]
housing       & \begin{tabular}[c]{@{}l@{}}'n\_splines': 25,\\ 'lam': 0.9\end{tabular}  & \begin{tabular}[c]{@{}l@{}}'spline\_type': 'ts',\\ 'maxk': 20,\\ 'm': 2,\\ 'gamma': 1.4\end{tabular} & \begin{tabular}[c]{@{}l@{}}'max\_bins': 256,\\ 'interactions': 20,\\ 'outer\_bags': 16,\\ 'inner\_bags': 4\end{tabular} & \begin{tabular}[c]{@{}l@{}}'lr': 0.01,\\ 'num\_learners': 8,\\ 'dropout': 0,\\ 'num\_basis\_functions': 64\end{tabular}    & \begin{tabular}[c]{@{}l@{}}'interact\_num': 20,\\ 'activation\_func': 'ReLU',\\ 'reg\_clarity*': 0.01\end{tabular}    & \begin{tabular}[c]{@{}l@{}}'subnet\_num': 10,\\ 'l1\_proj': 0.0001,\\ 'l1\_subnet*': 0.001\end{tabular}  & \begin{tabular}[c]{@{}l@{}}'boost\_rate': 0.1,\\ 'elm\_scale': 5,\\ 'interactions': 20\end{tabular}    \\[35pt]
diamond       & \begin{tabular}[c]{@{}l@{}}'n\_splines': 25,\\ 'lam': 0.2\end{tabular}  & \begin{tabular}[c]{@{}l@{}}'spline\_type': 'ts',\\ 'maxk*': 10,\\ 'm': 2,\\ 'gamma': 1\end{tabular} & \begin{tabular}[c]{@{}l@{}}'max\_bins': 256,\\ 'interactions': 20,\\ 'outer\_bags': 16,\\ 'inner\_bags': 4\end{tabular} & \begin{tabular}[c]{@{}l@{}}'lr': 0.01,\\ 'num\_learners': 8,\\ 'dropout': 0,\\ 'num\_basis\_functions': 64\end{tabular}    & \begin{tabular}[c]{@{}l@{}}'interact\_num': 20,\\ 'activation\_func': 'ReLU',\\ 'reg\_clarity*': 0.01\end{tabular}     & \begin{tabular}[c]{@{}l@{}}'subnet\_num': 10,\\ 'l1\_proj': 0.001,\\ 'l1\_subnet': 0.0001\end{tabular}     & \begin{tabular}[c]{@{}l@{}}'boost\_rate': 0.1,\\ 'elm\_scale*': 2,\\ 'interactions': 10\end{tabular}   \\ \bottomrule
\end{tabular}%
}
\end{table}

\begin{table}[ht]
\centering
\caption{Best hyperparameter configurations across all five folds of the 5-fold cross-validation (determined by majority voting) for the traditional interpretable models and the black-box models. Parameters for which the majority vote resulted in a tie are marked with *. For these, the first emerging parameter value is given.}
\label{tab:hpo_best_config_fold_majority_iml_blackbox}
\resizebox*{!}{0.9\textheight}{%
\begin{tabular}{@{}llllllll@{}}
\toprule
& \multicolumn{2}{c}{\textbf{Traditional Interpretable Models}} & \multicolumn{5}{c}{\textbf{Black-box Models}} \\ \cmidrule(l){2-3} \cmidrule(l){4-8}
\textbf{Task} & \textbf{LR} & \textbf{DT} & \textbf{RF} & \textbf{XGB} & \textbf{CatBoost} & \textbf{MLP} & \textbf{TabNet}  \\ \midrule
college       & \begin{tabular}[c]{@{}l@{}}'C': 0.1,\\ 'penalty': 'l2',\\ 'class\_weight': 'balanced',\\ 'solver': 'lbfgs',\\ 'l1\_ratio': 'None',\\ 'max\_iter': 100\end{tabular}       & \begin{tabular}[c]{@{}l@{}}'max\_depth*': 20,\\ 'max\_leaf\_nodes*': 40,\\ 'class\_weight': 'balanced',\\ 'splitter': 'best'\end{tabular} & \begin{tabular}[c]{@{}l@{}}'n\_estimators': 100,\\ 'max\_depth': 10,\\ 'class\_weight': 'balanced'\end{tabular}   & \begin{tabular}[c]{@{}l@{}}'max\_depth*': 9,\\ 'learning\_rate*': 0.01,\\ 'n\_estimators': 2000\end{tabular} & \begin{tabular}[c]{@{}l@{}}'n\_estimators': 500,\\ 'max\_depth': 6,\\ 'eta': 0.3\end{tabular}      & \begin{tabular}[c]{@{}l@{}}'hidden\_layer\_sizes': '{[}50 50{]}',\\ 'alpha': 0.0001,\\ 'activation': 'relu'\end{tabular}        & \begin{tabular}[c]{@{}l@{}}'n\_a\_and\_d*': 32,\\ 'n\_steps*': 5,\\ 'gamma*': 1.5\end{tabular}   \\[35pt]
water         & \begin{tabular}[c]{@{}l@{}}'C': 0.001,\\ 'penalty': 'l1',\\ 'class\_weight': 'None',\\ 'solver': 'liblinear',\\ 'l1\_ratio': 'None',\\ 'max\_iter': 100\end{tabular}     & \begin{tabular}[c]{@{}l@{}}'max\_depth': 5,\\ 'max\_leaf\_nodes': 20,\\ 'class\_weight': 'None',\\ 'splitter': 'best'\end{tabular}        & \begin{tabular}[c]{@{}l@{}}'n\_estimators': 1000,\\ 'max\_depth': 20,\\ 'class\_weight': 'balanced'\end{tabular}  & \begin{tabular}[c]{@{}l@{}}'max\_depth*': 3,\\ 'learning\_rate': 0.01,\\ 'n\_estimators*': 200\end{tabular}  & \begin{tabular}[c]{@{}l@{}}'n\_estimators': 100,\\ 'max\_depth': 6,\\ 'eta*': 0.1\end{tabular}     & \begin{tabular}[c]{@{}l@{}}'hidden\_layer\_sizes*': '{[}50 50 50 50{]}',\\ 'alpha': 0.0001,\\ 'activation': 'tanh'\end{tabular}   & \begin{tabular}[c]{@{}l@{}}'n\_a\_and\_d*': 8,\\ 'n\_steps': 3,\\ 'gamma': 1.5\end{tabular}    \\[35pt]
stroke        & \begin{tabular}[c]{@{}l@{}}'C': 0.001,\\ 'penalty': 'l1',\\ 'class\_weight': 'balanced',\\ 'solver': 'liblinear',\\ 'l1\_ratio': 'None',\\ 'max\_iter': 100\end{tabular} & \begin{tabular}[c]{@{}l@{}}'max\_depth': 5,\\ 'max\_leaf\_nodes': 5,\\ 'class\_weight': 'None',\\ 'splitter': 'random'\end{tabular}       & \begin{tabular}[c]{@{}l@{}}'n\_estimators': 50,\\ 'max\_depth': 5,\\ 'class\_weight': 'None'\end{tabular}         & \begin{tabular}[c]{@{}l@{}}'max\_depth': 3,\\ 'learning\_rate': 0.01,\\ 'n\_estimators': 50\end{tabular}     & \begin{tabular}[c]{@{}l@{}}'n\_estimators': 50,\\ 'max\_depth': 3,\\ 'eta': 0.01\end{tabular}      & \begin{tabular}[c]{@{}l@{}}'hidden\_layer\_sizes': '{[}100{]}',\\ 'alpha': 0.0001,\\ 'activation': 'relu'\end{tabular}            & \begin{tabular}[c]{@{}l@{}}'n\_a\_and\_d*': 16,\\ 'n\_steps*': 10,\\ 'gamma': 1.3\end{tabular}    \\[35pt]
churn         & \begin{tabular}[c]{@{}l@{}}'C*': 0.01,\\ 'penalty': 'l1',\\ 'class\_weight': 'None',\\ 'solver': 'liblinear',\\ 'l1\_ratio': 'None',\\ 'max\_iter': 100\end{tabular}     & \begin{tabular}[c]{@{}l@{}}'max\_depth*': 5,\\ 'max\_leaf\_nodes': 20,\\ 'class\_weight': 'None',\\ 'splitter': 'best'\end{tabular}       & \begin{tabular}[c]{@{}l@{}}'n\_estimators': 200,\\ 'max\_depth': 10,\\ 'class\_weight': 'None'\end{tabular}       & \begin{tabular}[c]{@{}l@{}}'max\_depth': 3,\\ 'learning\_rate': 0.01,\\ 'n\_estimators': 500\end{tabular}    & \begin{tabular}[c]{@{}l@{}}'n\_estimators': 500,\\ 'max\_depth': 3,\\ 'eta': 0.01\end{tabular}     & \begin{tabular}[c]{@{}l@{}}'hidden\_layer\_sizes*': '{[}25 25 25{]}',\\ 'alpha': 0.0001,\\ 'activation': 'tanh'\end{tabular}      & \begin{tabular}[c]{@{}l@{}}'n\_a\_and\_d': 8,\\ 'n\_steps': 5,\\ 'gamma*': 1.8\end{tabular}   \\[35pt]
recidivism    & \begin{tabular}[c]{@{}l@{}}'C': 0.01,\\ 'penalty*': 'l1',\\ 'class\_weight': 'None',\\ 'solver': 'saga',\\ 'l1\_ratio': 'None',\\ 'max\_iter': 100\end{tabular}          & \begin{tabular}[c]{@{}l@{}}'max\_depth': 10,\\ 'max\_leaf\_nodes': 40,\\ 'class\_weight': 'None',\\ 'splitter': 'best'\end{tabular}       & \begin{tabular}[c]{@{}l@{}}'n\_estimators': 1000,\\ 'max\_depth': 10,\\ 'class\_weight': 'balanced'\end{tabular}  & \begin{tabular}[c]{@{}l@{}}'max\_depth': 3,\\ 'learning\_rate': 0.1,\\ 'n\_estimators': 50\end{tabular}      & \begin{tabular}[c]{@{}l@{}}'n\_estimators': 100,\\ 'max\_depth': 3,\\ 'eta': 0.3\end{tabular}      & \begin{tabular}[c]{@{}l@{}}'hidden\_layer\_sizes': '{[}100 100{]}',\\ 'alpha*': 0.001,\\ 'activation': 'tanh'\end{tabular}        & \begin{tabular}[c]{@{}l@{}}'n\_a\_and\_d': 8,\\ 'n\_steps': 3,\\ 'gamma': 1.8\end{tabular} \\[35pt]
credit        & \begin{tabular}[c]{@{}l@{}}'C*': 0.001,\\ 'penalty': 'l2',\\ 'class\_weight': 'None',\\ 'solver': 'lbfgs',\\ 'l1\_ratio': 'None',\\ 'max\_iter': 100\end{tabular}        & \begin{tabular}[c]{@{}l@{}}'max\_depth': 5,\\ 'max\_leaf\_nodes*': 10,\\ 'class\_weight': 'None',\\ 'splitter': 'best'\end{tabular}       & \begin{tabular}[c]{@{}l@{}}'n\_estimators*': 200,\\ 'max\_depth*': 40,\\ 'class\_weight': 'balanced'\end{tabular} & \begin{tabular}[c]{@{}l@{}}'max\_depth': 3,\\ 'learning\_rate*': 0.1,\\ 'n\_estimators': 50\end{tabular}     & \begin{tabular}[c]{@{}l@{}}'n\_estimators*': 1000,\\ 'max\_depth*': 6,\\ 'eta': 0.01\end{tabular}  & \begin{tabular}[c]{@{}l@{}}'hidden\_layer\_sizes': '{[}100 100 100{]}',\\ 'alpha': 0.0001,\\ 'activation': 'tanh'\end{tabular}    & \begin{tabular}[c]{@{}l@{}}'n\_a\_and\_d': 32,\\ 'n\_steps': 3,\\ 'gamma': 1.3\end{tabular}   \\[35pt]
income        & \begin{tabular}[c]{@{}l@{}}'C': 0.1,\\ 'penalty': 'l2',\\ 'class\_weight': 'None',\\ 'solver*': 'lbfgs',\\ 'l1\_ratio': 'None',\\ 'max\_iter': 100\end{tabular}          & \begin{tabular}[c]{@{}l@{}}'max\_depth': 20,\\ 'max\_leaf\_nodes': 40,\\ 'class\_weight': 'None',\\ 'splitter': 'best'\end{tabular}       & \begin{tabular}[c]{@{}l@{}}'n\_estimators': 200,\\ 'max\_depth': 20,\\ 'class\_weight': 'None'\end{tabular}       & \begin{tabular}[c]{@{}l@{}}'max\_depth': 3,\\ 'learning\_rate': 0.3,\\ 'n\_estimators': 100\end{tabular}     & \begin{tabular}[c]{@{}l@{}}'n\_estimators': 500,\\ 'max\_depth*': 6,\\ 'eta': 0.03\end{tabular}    & \begin{tabular}[c]{@{}l@{}}'hidden\_layer\_sizes': '{[}100 100 100{]}',\\ 'alpha': 0.01,\\ 'activation': 'tanh'\end{tabular}      & \begin{tabular}[c]{@{}l@{}}'n\_a\_and\_d*': 32,\\ 'n\_steps': 3,\\ 'gamma': 1.5\end{tabular}     \\[35pt]
bank          & \begin{tabular}[c]{@{}l@{}}'C': 0.1,\\ 'penalty*': 'l2',\\ 'class\_weight': 'None',\\ 'solver': 'saga',\\ 'l1\_ratio': 'None',\\ 'max\_iter': 100\end{tabular}           & \begin{tabular}[c]{@{}l@{}}'max\_depth': 5,\\ 'max\_leaf\_nodes': 20,\\ 'class\_weight': 'None',\\ 'splitter': 'random'\end{tabular}      & \begin{tabular}[c]{@{}l@{}}'n\_estimators': 500,\\ 'max\_depth': 20,\\ 'class\_weight': 'None'\end{tabular}       & \begin{tabular}[c]{@{}l@{}}'max\_depth': 9,\\ 'learning\_rate*': 0.1,\\ 'n\_estimators': 50\end{tabular}     & \begin{tabular}[c]{@{}l@{}}'n\_estimators*': 200,\\ 'max\_depth*': 12,\\ 'eta*': 0.03\end{tabular} & \begin{tabular}[c]{@{}l@{}}'hidden\_layer\_sizes': '{[}50{]}',\\ 'alpha': 0.0001,\\ 'activation': 'relu'\end{tabular}             & \begin{tabular}[c]{@{}l@{}}'n\_a\_and\_d*': 8,\\ 'n\_steps*': 5,\\ 'gamma*': 1.8\end{tabular}   \\[35pt]
airline       & \begin{tabular}[c]{@{}l@{}}'C': 0.01,\\ 'penalty': 'l1',\\ 'class\_weight': 'None',\\ 'solver': 'liblinear',\\ 'l1\_ratio': 'None',\\ 'max\_iter': 100\end{tabular}      & \begin{tabular}[c]{@{}l@{}}'max\_depth': 20,\\ 'max\_leaf\_nodes': 'None',\\ 'class\_weight': 'None',\\ 'splitter': 'random'\end{tabular} & \begin{tabular}[c]{@{}l@{}}'n\_estimators*': 500,\\ 'max\_depth': 40,\\ 'class\_weight': 'balanced'\end{tabular}  & \begin{tabular}[c]{@{}l@{}}'max\_depth': 12,\\ 'learning\_rate': 0.01,\\ 'n\_estimators*': 2000\end{tabular} & \begin{tabular}[c]{@{}l@{}}'n\_estimators': 1000,\\ 'max\_depth': 9,\\ 'eta': 0.03\end{tabular}    & \begin{tabular}[c]{@{}l@{}}'hidden\_layer\_sizes*': '{[}100 100{]}',\\ 'alpha': 0.001,\\ 'activation': 'tanh'\end{tabular}        & \begin{tabular}[c]{@{}l@{}}'n\_a\_and\_d': 32,\\ 'n\_steps': 3,\\ 'gamma': 1.3\end{tabular}    \\[35pt]

weather       & \begin{tabular}[c]{@{}l@{}}'C*': 1,\\ 'penalty*': 'l1',\\ 'class\_weight': 'None',\\ 'solver': 'saga',\\ 'l1\_ratio': 'None',\\ 'max\_iter': 300\end{tabular}            & \begin{tabular}[c]{@{}l@{}}'max\_depth': 10,\\ 'max\_leaf\_nodes': 'None',\\ 'class\_weight': 'None',\\ 'splitter': 'best'\end{tabular}   & \begin{tabular}[c]{@{}l@{}}'n\_estimators': 1000,\\ 'max\_depth': 'None',\\ 'class\_weight': 'None'\end{tabular}  & \begin{tabular}[c]{@{}l@{}}'max\_depth': 12,\\ 'learning\_rate': 0.01,\\ 'n\_estimators': 2000\end{tabular}  & \begin{tabular}[c]{@{}l@{}}'n\_estimators': 1000,\\ 'max\_depth': 12,\\ 'eta': 0.03\end{tabular}   & \begin{tabular}[c]{@{}l@{}}'hidden\_layer\_sizes': '{[}100 100{]}',\\ 'alpha': 0.01,\\ 'activation': 'tanh'\end{tabular}          & \begin{tabular}[c]{@{}l@{}}'n\_a\_and\_d': 16,\\ 'n\_steps': 3,\\ 'gamma': 1.8\end{tabular}     \\[35pt] \midrule
car           & \begin{tabular}[c]{@{}l@{}}'alpha*': 0.01,\\ 'l1\_ratio': 0\end{tabular}                                                                                                 &
\begin{tabular}[c]{@{}l@{}}'max\_depth': 5,\\ 'max\_leaf\_nodes': 20,\\ 'class\_weight': 'None',\\ 'splitter': 'best'\end{tabular}        & \begin{tabular}[c]{@{}l@{}}'n\_estimators*': 100,\\ 'max\_depth*': 40,\\ 'class\_weight': 'None'\end{tabular}     & \begin{tabular}[c]{@{}l@{}}'max\_depth*': 6,\\ 'learning\_rate*': 0.3,\\ 'n\_estimators*': 50\end{tabular}   & \begin{tabular}[c]{@{}l@{}}'n\_estimators': 500,\\ 'max\_depth': 3,\\ 'eta*': 0.1\end{tabular}     & \begin{tabular}[c]{@{}l@{}}'hidden\_layer\_sizes': '{[}100 100 100{]}',\\ 'alpha': 0.01,\\ 'activation': 'relu'\end{tabular}      & \begin{tabular}[c]{@{}l@{}}'n\_a\_and\_d': 32,\\ 'n\_steps': 3,\\ 'gamma': 1.3\end{tabular}     \\[35pt]
student       & \begin{tabular}[c]{@{}l@{}}'alpha*': 0.01,\\ 'l1\_ratio': 0\end{tabular}                                                                                                &
\begin{tabular}[c]{@{}l@{}}'max\_depth': 5,\\ 'max\_leaf\_nodes*': 5,\\ 'class\_weight': 'None',\\ 'splitter': 'random'\end{tabular}   & \begin{tabular}[c]{@{}l@{}}'n\_estimators*': 100,\\ 'max\_depth': 5,\\ 'class\_weight': 'None'\end{tabular}      & \begin{tabular}[c]{@{}l@{}}'max\_depth': 3,\\ 'learning\_rate': 0.1,\\ 'n\_estimators': 50\end{tabular}   & \begin{tabular}[c]{@{}l@{}}'n\_estimators': 100,\\ 'max\_depth': 6,\\ 'eta*': 0.1\end{tabular}     & \begin{tabular}[c]{@{}l@{}}'hidden\_layer\_sizes*': '{[}100 100{]}',\\ 'alpha': 0.01,\\ 'activation': 'tanh'\end{tabular}        & \begin{tabular}[c]{@{}l@{}}'n\_a\_and\_d': 16,\\ 'n\_steps': 3,\\ 'gamma': 1.5\end{tabular}     \\[35pt]
productivity  & \begin{tabular}[c]{@{}l@{}}'alpha': 0.01,\\ 'l1\_ratio': 0\end{tabular}                                                                                                 &
\begin{tabular}[c]{@{}l@{}}'max\_depth': 5,\\ 'max\_leaf\_nodes': 40,\\ 'class\_weight': 'None',\\ 'splitter': 'best'\end{tabular}       & \begin{tabular}[c]{@{}l@{}}'n\_estimators*': 50,\\ 'max\_depth': 10,\\ 'class\_weight': 'None'\end{tabular}       & \begin{tabular}[c]{@{}l@{}}'max\_depth': 3,\\ 'learning\_rate': 0.1,\\ 'n\_estimators*': 500\end{tabular}      & \begin{tabular}[c]{@{}l@{}}'n\_estimators': 1000,\\ 'max\_depth*': 6,\\ 'eta*': 0.03\end{tabular}  & \begin{tabular}[c]{@{}l@{}}'hidden\_layer\_sizes*': '{[}75 75{]}',\\ 'alpha': 0.01,\\ 'activation': 'tanh'\end{tabular}         & \begin{tabular}[c]{@{}l@{}}'n\_a\_and\_d*': 32,\\ 'n\_steps*': 3,\\ 'gamma': 1.8\end{tabular} \\[35pt]
insurance     & \begin{tabular}[c]{@{}l@{}}'alpha': 0.001,\\ 'l1\_ratio': 1\end{tabular}                                                                                                  &
\begin{tabular}[c]{@{}l@{}}'max\_depth': 5,\\ 'max\_leaf\_nodes': 10,\\ 'class\_weight': 'None',\\ 'splitter': 'best'\end{tabular}        & \begin{tabular}[c]{@{}l@{}}'n\_estimators': 1000,\\ 'max\_depth': 5,\\ 'class\_weight': 'None'\end{tabular}       & \begin{tabular}[c]{@{}l@{}}'max\_depth': 3,\\ 'learning\_rate': 0.1,\\ 'n\_estimators': 50\end{tabular}    & \begin{tabular}[c]{@{}l@{}}n\_estimators': 1000,\\ 'max\_depth': 3,\\ 'eta': 0.01\end{tabular}     & \begin{tabular}[c]{@{}l@{}}'hidden\_layer\_sizes*': '{[}50{]}',\\ 'alpha': 0.0001,\\ 'activation': 'relu'\end{tabular}           & \begin{tabular}[c]{@{}l@{}}'n\_a\_and\_d': 16,\\ 'n\_steps*': 10,\\ 'gamma': 1.5\end{tabular}    \\[35pt]
crimes        & \begin{tabular}[c]{@{}l@{}}'alpha': 0.1,\\ 'l1\_ratio*': 0.1\end{tabular}                                                                                                 & 
\begin{tabular}[c]{@{}l@{}}'max\_depth': 5,\\ 'max\_leaf\_nodes*': 'None',\\ 'class\_weight': 'None',\\ 'splitter': 'best'\end{tabular}  & \begin{tabular}[c]{@{}l@{}}'n\_estimators': 1000,\\ 'max\_depth': 10,\\ 'class\_weight': 'None'\end{tabular}      & \begin{tabular}[c]{@{}l@{}}'max\_depth': 3,\\ 'learning\_rate': 0.01,\\ 'n\_estimators*': 100\end{tabular}   & \begin{tabular}[c]{@{}l@{}}'n\_estimators*': 500,\\ 'max\_depth': 6,\\ 'eta*': 0.1\end{tabular}    & \begin{tabular}[c]{@{}l@{}}'hidden\_layer\_sizes': '{[}50 50 50{]}',\\ 'alpha': 0.001,\\ 'activation': 'relu'\end{tabular}          & \begin{tabular}[c]{@{}l@{}}'n\_a\_and\_d*': 32,\\ 'n\_steps*': 10,\\ 'gamma*': 1.8\end{tabular}  \\[35pt]
farming       & \begin{tabular}[c]{@{}l@{}}'alpha': 0.001,\\ 'l1\_ratio': 1\end{tabular}                                                                                                 & 
\begin{tabular}[c]{@{}l@{}}'max\_depth': 10,\\ 'max\_leaf\_nodes': 20,\\ 'class\_weight': 'None',\\ 'splitter': 'best'\end{tabular}       & \begin{tabular}[c]{@{}l@{}}'n\_estimators': 1000,\\ 'max\_depth': 10,\\ 'class\_weight': 'None'\end{tabular}      & \begin{tabular}[c]{@{}l@{}}'max\_depth': 3,\\ 'learning\_rate': 0.1,\\ 'n\_estimators': 200\end{tabular}     & \begin{tabular}[c]{@{}l@{}}'n\_estimators': 200,\\ 'max\_depth*': 9,\\ 'eta*': 0.1\end{tabular}      & \begin{tabular}[c]{@{}l@{}}'hidden\_layer\_sizes*': '{[}50 50 50 50{]}',\\ 'alpha': 0.01,\\ 'activation': 'tanh'\end{tabular}     & \begin{tabular}[c]{@{}l@{}}'n\_a\_and\_d*': 8,\\ 'n\_steps': 3,\\ 'gamma': 1.8\end{tabular}     \\[35pt]
wine          & \begin{tabular}[c]{@{}l@{}}'alpha': 0.001,\\ 'l1\_ratio': 1\end{tabular}                                                                                                 & 
\begin{tabular}[c]{@{}l@{}}'max\_depth': 10,\\ 'max\_leaf\_nodes*': 40,\\ 'class\_weight': 'None',\\ 'splitter': 'best'\end{tabular}      & \begin{tabular}[c]{@{}l@{}}'n\_estimators': 1000,\\ 'max\_depth': 40,\\ 'class\_weight': 'None'\end{tabular}      & \begin{tabular}[c]{@{}l@{}}'max\_depth': 6,\\ 'learning\_rate': 0.1,\\ 'n\_estimators': 500\end{tabular}     & \begin{tabular}[c]{@{}l@{}}'n\_estimators': 1000,\\ 'max\_depth': 12,\\ 'eta': 0.03\end{tabular}   & \begin{tabular}[c]{@{}l@{}}'hidden\_layer\_sizes': '{[}100 100 100 100{]}',\\ 'alpha': 0.01,\\ 'activation': 'relu'\end{tabular}  & \begin{tabular}[c]{@{}l@{}}'n\_a\_and\_d': 32,\\ 'n\_steps': 3,\\ 'gamma': 1.3\end{tabular}  \\[35pt]
bike          & \begin{tabular}[c]{@{}l@{}}'alpha': 0.001,\\ 'l1\_ratio': 0\end{tabular}                                                                                                 & 
\begin{tabular}[c]{@{}l@{}}'max\_depth': 10,\\ 'max\_leaf\_nodes': 'None',\\ 'class\_weight': 'None',\\ 'splitter': 'best'\end{tabular}   & \begin{tabular}[c]{@{}l@{}}'n\_estimators': 1000,\\ 'max\_depth': 40,\\ 'class\_weight': 'None'\end{tabular}       & \begin{tabular}[c]{@{}l@{}}'max\_depth': 9,\\ 'learning\_rate': 0.01,\\ 'n\_estimators': 2000\end{tabular}    & \begin{tabular}[c]{@{}l@{}}'n\_estimators': 1000,\\ 'max\_depth': 12,\\ 'eta': 0.03\end{tabular}   & \begin{tabular}[c]{@{}l@{}}'hidden\_layer\_sizes*': '{[}100 100{]}',\\ 'alpha': 0.01,\\ 'activation': 'tanh'\end{tabular} & \begin{tabular}[c]{@{}l@{}}'n\_a\_and\_d*': 32,\\ 'n\_steps': 3,\\ 'gamma': 1.3\end{tabular}   \\[35pt]
housing       & \begin{tabular}[c]{@{}l@{}}'alpha': 0.001,\\ 'l1\_ratio': 0\end{tabular}                                                                                                 & 
\begin{tabular}[c]{@{}l@{}}'max\_depth': 10,\\ 'max\_leaf\_nodes': 'None',\\ 'class\_weight': 'None',\\ 'splitter': 'best'\end{tabular}   & \begin{tabular}[c]{@{}l@{}}'n\_estimators': 500,\\ 'max\_depth': 40,\\ 'class\_weight': 'None'\end{tabular}      & \begin{tabular}[c]{@{}l@{}}'max\_depth': 6,\\ 'learning\_rate': 0.1,\\ 'n\_estimators': 2000\end{tabular}   & \begin{tabular}[c]{@{}l@{}}'n\_estimators': 1000,\\ 'max\_depth': 9,\\ 'eta': 0.1\end{tabular}     & \begin{tabular}[c]{@{}l@{}}'hidden\_layer\_sizes*': '{[}100 100 100 100{]}',\\ 'alpha': 0.01,\\ 'activation': 'tanh'\end{tabular}     & \begin{tabular}[c]{@{}l@{}}'n\_a\_and\_d': 16,\\ 'n\_steps': 3,\\ 'gamma*': 1.5\end{tabular}   \\[35pt]
diamond       & \begin{tabular}[c]{@{}l@{}}'alpha': 0.001,\\ 'l1\_ratio': 0\end{tabular}                                                                                                 & 
\begin{tabular}[c]{@{}l@{}}'max\_depth': 20,\\ 'max\_leaf\_nodes': 'None',\\ 'class\_weight': 'None',\\ 'splitter': 'best'\end{tabular}        & \begin{tabular}[c]{@{}l@{}}'n\_estimators': 1000,\\ 'max\_depth': 20,\\ 'class\_weight': 'None'\end{tabular}       & \begin{tabular}[c]{@{}l@{}}'max\_depth': 9,\\ 'learning\_rate': 0.01,\\ 'n\_estimators': 2000\end{tabular}      & \begin{tabular}[c]{@{}l@{}}'n\_estimators': 1000,\\ 'max\_depth*': 6,\\ 'eta': 0.03\end{tabular}    & \begin{tabular}[c]{@{}l@{}}'hidden\_layer\_sizes*': '{[}50 50 50 50{]}',\\ 'alpha': 0.01,\\ 'activation': 'tanh'\end{tabular}            & \begin{tabular}[c]{@{}l@{}}'n\_a\_and\_d': 16,\\ 'n\_steps': 3,\\ 'gamma': 1.3\end{tabular}    \\ \bottomrule
\end{tabular}%
}
\end{table}

\clearpage
\newpage

\noindent Table \ref{tab:tuning_overview} shows the number of tuning candidates. 
The grid for linear regression theoretically has $1,344$ candidates by combining the parameter lists from the hyperparameter grid in Table \ref{tab:hyperparameter_grid}. However, this number is constrained by the possible combinations between $solver$ and $penalty$, as well as between $l1\_ratio$ and $penalty$. The scikit-learn documentation specifies which penalties are supported by each solver. For example, the linear/logistic regression candidate grid yields $232$ candidates for classification and $77$ for regression. For the decision tree and the random forest, the $class\_weight$ parameter becomes obsolete for the regression task, resulting in $50$ and $25$ runs, compared to $100$ and $50$ runs for the classification task.

\begin{table}[htp!]
\sisetup{group-digits=true, group-separator={,}, group-minimum-digits=4}
\caption{Overview of the number of hyperparameter tuning runs.}
\label{tab:tuning_overview}
\resizebox{\textwidth}{!}{%
\begin{tabular}{
l
S[table-format=4.3]
S[table-format=4.3]
S[table-format=4.3]
S[table-format=4.3]
S[table-format=4.3]
S[table-format=6.3]
S[table-format=4.3]
S[table-format=4.3]
S[table-format=4.3]
S[table-format=4.3]
S[table-format=4.3]
S[table-format=4.3]
S[table-format=4.3]
S[table-format=4.3]@{\hspace{0.5cm}}
S[table-format=7.3]<{\hspace{0pt}}
}
\toprule
 & \multicolumn{9}{c}{\textbf{Interpretable Models}} & \multicolumn{4}{c}{\textbf{Black-box Models}} &  \\ \cmidrule(l){2-10} \cmidrule(l){11-15} 
& \multicolumn{7}{c}{\textbf{GAMs}} & \multicolumn{2}{c}{\textbf{Traditional}} & & & \\ \cmidrule(l){2-8} \cmidrule(l){9-10}
 & \textbf{P-Splines} & \textbf{TP-Splines} & \textbf{EBM} & \textbf{NAM} & \textbf{GAMI-Net} & \textbf{ExNN} & \textbf{IGANN} & \textbf{LR} & \textbf{DT} & \textbf{RF} & \textbf{XGB} & \textbf{CatBoost} & \textbf{MLP} & \textbf{TabNet} & \textbf{Sum} \\ \cmidrule(l){1-15} \cmidrule(l){16-16} 
Number of tuning candidates (CLS) & 20 & 18 & 24 & 24 & 27 & 18	& 18 & 232 & 100 & 50 & 90 & 80 & 72 & 27 & \\ \cmidrule(l){1-15} \cmidrule(l){16-16} 
Number of tuning candidates (REG) & 20 & 18 & 24 & 24 & 27 & 18	& 18 & 77 & 50 & 25 & 90 & 80 & 72 & 27 & \\ \cmidrule(l){1-15} \cmidrule(l){16-16} 
 \begin{tabular}{@{}l@{}} Total tuning candidate runs \\ for all 20 datasets and 5 folds \end{tabular}& 2000 & 1800 & 2400 & 2400 &	2700 &	1800 &	1800 &	15450 &	7500  &	3750 &	9000 & 8000 & 7200 & 2700 & 68500 \\ \bottomrule
\end{tabular}%
}
\end{table}

\clearpage
\newpage

\section{Training Runtimes}
\label{app:training_times}

In addition to evaluating the predictive performance, we also examine the training times of the models to assess their usability. For this purpose, we measure the training times in average seconds per fold for the models with default and tuned hyperparameters. The results of the default setting are summarized in Table~\ref{tab:default_hp_training_time} and the results of the tuned setting are summarized in Table~\ref{tab:tuned_hp_training_time}. Furthermore, we report the mean runtimes across all hyperparameters in Table~\ref{tab:tuning_runtime_mean_model} and provide a summary of aggregated tuning runtimes across all datasets and models in Table~\ref{tab:tuning_runtime_overview}.

Considering the results of the default setting (Table~\ref{tab:default_hp_training_time}), it can be seen that the simpler interpretable models \gls{dt} and \gls{lr} unsurprisingly yield the shortest training times, mostly under 1 second, due to their basic model structures. Similarly, the tree-based ensemble models are also characterized by very short runtimes in their default configuration (\gls{rf}: 0.25, \gls{xgb}: 0.31, CatBoost: 1.05). The remaining two black-box models \gls{mlp} and TabNet, on the other hand, require much longer training, with average runtimes of 3.55 and 12.1 seconds, respectively.
The runtimes of the \glspl{gam} show mixed results. \gls{igann}, \gls{ebm}, and \gls{psplines} are the fastest variants, with average training times of 3.1, 4.64, and 6.13 seconds, respectively. Thus, their computational complexity is comparable to that of \gls{mlp}. In contrast, the \glspl{gam} based on deep neural networks show the longest training times. In particular, \gls{nam} and \gls{exnn}, with average runtimes of 35.89 and 179.9 seconds, are characterized by computationally intensive operations, whereas \gls{gaminet} shows medium runtimes with 22.69 seconds. With an average value of 41.88 seconds, the runtime of \gls{tpsplines} is also relatively long. Although the model generally has similar training times to \gls{psplines}, the results are biased by some exceptional cases where the model is unable to converge in a short time (e.g., \textit{airline}, \textit{weather}, and \textit{crimes}).

In the setting with the best hyperparameters (Table~\ref{tab:tuned_hp_training_time}), the majority of models show longer training times. In the case of \gls{ebm}, \gls{tpsplines}, \gls{exnn}, and \gls{mlp}, training times are increased by a small factor of about 1.5 to 2.5, whereas \gls{igann}, \gls{nam}, \gls{rf}, and CatBoost show increased training times by a factor of about 3 to 7. The biggest difference, however, is observed with \gls{xgb}, where the increase in training time is about a factor of 15 on average.
By contrast, for some models, the training times remain more or less unchanged compared to the non-tuned variants. These include \gls{gaminet}, \gls{psplines}, \gls{lr}, and TabNet. In the case of \gls{dt}, the training time is even reduced by half on average, since the tuned hyperparameters limit the flexibility/complexity of the trees and thus require less computation.

Overall, the results show that the interpretability constraints of the \glspl{gam} do not necessarily lead to impractically long training times. Especially \gls{ebm}, \gls{gaminet}, and \gls{igann}, as the best-performing \glspl{gam}, are characterized by both strong prediction qualities and moderate training times, making them promising candidates for data science projects where both criteria must be satisfied. 

\begin{table}[htp!]
\caption{Mean training runtimes per fold, measured in seconds, using the default hyperparameter set. Models employing GPU acceleration are denoted with `**'.}
\label{tab:default_hp_training_time}
\resizebox{1\textwidth}{!}{%
\begin{tabular}{@{}l@{\hspace{0.1cm}}lllllllllllllll@{}}
\toprule
& & \multicolumn{9}{c}{\textbf{Interpretable Models}} & \multicolumn{5}{c}{\textbf{Black-box Models}} \\ \cmidrule(l){3-11} \cmidrule(l){12-16} 
& & \multicolumn{7}{c}{\textbf{GAMs}} & \multicolumn{2}{c}{\textbf{Traditional}} & & & \\ \cmidrule(l){3-9} \cmidrule(l){10-11}
\textbf{Task} & \textbf{Dataset} & \rotatebox{45}{\textbf{P-Splines}} & \rotatebox{45}{\textbf{TP-Splines}} & \rotatebox{45}{\textbf{EBM}} & \rotatebox{45}{\textbf{NAM}$^{**}$} & \rotatebox{45}{\textbf{GAMI-Net}} & \rotatebox{45}{\textbf{ExNN}} & \rotatebox{45}{\textbf{IGANN}$^{**}$} & \rotatebox{45}{\textbf{LR}} & \rotatebox{45}{\textbf{DT}} & \rotatebox{45}{\textbf{RF}} & \rotatebox{45}{\textbf{XGB}}  & \rotatebox{45}{\textbf{CatBoost}$^{**}$} & \rotatebox{45}{\textbf{MLP}} & \rotatebox{45}{\textbf{TabNet}$^{**}$} \\ \midrule
\multicolumn{1}{l}{\multirow{25}{*}{CLS}} 
& college & \msd{0.140}{0.016}  & \msd{0.082}{0.051}    & \msd{0.396}{0.029}  & \msd{2.826}{1.102}    & \msd{5.712}{2.720}    & \msd{4.328}{0.343}      & \msd{0.106}{0.048}   & \msd{0.000}{0.000} & \msd{0.000}{0.000} & \msd{0.198}{0.008} & \msd{0.048}{0.004} & \msd{0.480}{0.000} & \msd{0.256}{0.049}  & \msd{4.626}{1.571}   \\[15pt]
 & water   & \msd{0.104}{0.005}  & \msd{0.910}{0.313}    & \msd{0.516}{0.051}  & \msd{2.096}{0.593}    & \msd{2.462}{0.451}    & \msd{38.852}{14.416}    & \msd{0.056}{0.009}   & \msd{0.000}{0.000} & \msd{0.012}{0.004} & \msd{0.196}{0.054} & \msd{0.080}{0.000} & \msd{0.658}{0.004} & \msd{0.348}{0.037}  & \msd{4.564}{1.028}   \\[15pt]
 & stroke  & \msd{1.118}{0.096}  & \msd{0.352}{0.099}    & \msd{0.478}{0.065}  & \msd{6.630}{2.735}    & \msd{5.326}{2.464}    & \msd{75.246}{8.068}     & \msd{0.100}{0.050}   & \msd{0.132}{0.004} & \msd{0.002}{0.004} & \msd{0.204}{0.024} & \msd{0.080}{0.000} & \msd{0.740}{0.035} & \msd{0.480}{0.145}  & \msd{4.764}{1.149}   \\[15pt]
 & churn   & \msd{1.714}{0.123}  & \msd{0.530}{0.094}    & \msd{0.998}{0.165}  & \msd{17.650}{6.678}   & \msd{15.230}{8.198}   & \msd{63.300}{8.717}     & \msd{0.384}{0.185}   & \msd{1.020}{0.000} & \msd{0.010}{0.000} & \msd{0.174}{0.036} & \msd{0.130}{0.012} & \msd{1.010}{0.010} & \msd{0.956}{0.430}  & \msd{3.942}{1.074}   \\[15pt]
 & recidivism  & \msd{0.596}{0.066}  & \msd{1.448}{0.384}    & \msd{0.644}{0.046}  & \msd{11.150}{3.318}   & \msd{4.992}{1.642}    & \msd{83.576}{42.462}    & \msd{1.156}{1.211}   & \msd{0.010}{0.000} & \msd{0.010}{0.000} & \msd{0.180}{0.048} & \msd{0.080}{0.007} & \msd{0.832}{0.008} & \msd{1.012}{0.356}  & \msd{4.642}{0.817}   \\[15pt]
 & credit    & \msd{3.272}{1.596}  & \textit{\msd{15.590}{2.948}}   & \msd{1.110}{0.117}  & \msd{16.244}{2.161}   & \msd{12.050}{5.308}   & \msd{71.520}{6.022}     & \msd{0.306}{0.304}   & \msd{1.030}{0.000} & \msd{0.040}{0.000} & \msd{0.180}{0.025} & \msd{0.136}{0.009} & \msd{1.026}{0.009} & \msd{0.986}{0.100}  & \msd{5.284}{1.758}   \\[15pt]
 & income   & \msd{32.424}{0.941} & \textit{\msd{32.377}{19.628}}  & \msd{13.312}{1.060} & \msd{67.090}{18.596}  & \msd{37.458}{22.962}  & \msd{244.040}{72.620}   & \msd{5.630}{6.196}   & \msd{1.078}{0.004} & \msd{0.070}{0.000} & \msd{0.212}{0.011} & \msd{0.980}{0.229} & \msd{1.994}{0.011} & \msd{3.750}{0.338}  & \msd{14.092}{1.893}  \\[15pt]
 & bank    & \msd{11.142}{3.348} & \msd{11.974}{3.128}   & \msd{4.972}{0.374}  & \msd{96.118}{25.529}  & \msd{39.210}{17.473}  & \msd{268.004}{48.053}   & \msd{1.876}{1.167}   & \msd{1.080}{0.000} & \msd{0.134}{0.005} & \msd{0.216}{0.009} & \msd{0.948}{0.097} & \msd{2.298}{0.029} & \msd{4.688}{0.227}  & \msd{22.634}{1.359}  \\[15pt]
 & airline & \msd{10.956}{0.105} & \msd{144.894}{25.058} & \msd{17.642}{0.455} & \msd{81.136}{23.471}  & \msd{127.908}{37.400} & \msd{1,141.652}{302.074} & \msd{12.190}{1.856}  & \msd{1.076}{0.005} & \msd{0.186}{0.005} & \msd{0.328}{0.015} & \msd{0.680}{0.027} & \msd{2.842}{0.033} & \msd{17.038}{1.271} & \msd{40.450}{6.052}  \\[15pt]
 & weather & \msd{54.490}{6.291} & \msd{271.994}{52.868} & \msd{36.320}{1.506} & \msd{228.432}{81.803} & \msd{101.056}{31.648} & \msd{792.080}{223.089}  & \msd{26.478}{12.834} & \msd{0.774}{0.193} & \msd{0.822}{0.018} & \msd{0.854}{0.011} & \msd{1.710}{0.014} & \msd{3.414}{0.056} & \msd{15.872}{0.828} & \msd{52.226}{9.737} \\[15pt] \midrule
\multicolumn{1}{l}{\multirow{25}{*}{REG}} 
 & car          & \msd{0.682}{0.035} & \textit{\msd{0.424}{0.251}}     & \msd{1.398}{0.858} & \msd{9.302}{2.039}   & \msd{5.192}{2.601}  & \msd{3.224}{0.445}     & \msd{0.726}{0.860} & \msd{0.010}{0.000} & \msd{0.000}{0.000} & \msd{0.134}{0.022} & \msd{0.046}{0.005} & \msd{0.332}{0.070} & \msd{0.086}{0.036} & \msd{2.910}{0.288} \\[15pt]
 & student      & \msd{0.302}{0.004} & \msd{0.154}{0.083}     & \msd{0.368}{0.080} & \msd{7.370}{3.378}   & \msd{3.596}{0.450}  & \msd{4.300}{0.213}     & \msd{0.182}{0.111} & \msd{0.010}{0.000} & \msd{0.000}{0.000} & \msd{0.120}{0.017} & \msd{0.070}{0.000} & \msd{0.476}{0.005} & \msd{0.094}{0.017} & \msd{2.600}{1.055}  \\[15pt]
 & productivity & \msd{0.308}{0.013} & \msd{0.136}{0.032}     & \msd{0.672}{0.098} & \msd{4.804}{1.294}   & \msd{8.862}{7.388}  & \msd{11.510}{1.002}    & \msd{3.166}{4.355} & \msd{0.034}{0.005} & \msd{0.000}{0.000} & \msd{0.144}{0.024} & \msd{0.070}{0.000} & \msd{0.338}{0.004} & \msd{0.208}{0.040} & \msd{3.706}{1.752}  \\[15pt]
 & insurance      & \msd{0.030}{0.000} & \msd{0.020}{0.000}     & \msd{0.378}{0.032} & \msd{2.438}{0.712}   & \msd{3.712}{0.736}  & \msd{12.586}{2.168}    & \msd{0.228}{0.366} & \msd{0.010}{0.000} & \msd{0.000}{0.000} & \msd{0.176}{0.024} & \msd{0.060}{0.000} & \msd{0.310}{0.000} & \msd{0.322}{0.140} & \msd{4.590}{0.900}  \\[15pt]
 & crimes       & \msd{2.904}{0.015} & \msd{354.870}{137.310} & \msd{1.078}{0.343} & \msd{16.644}{2.787}  & \msd{11.310}{2.104} & \msd{18.396}{0.395}    & \msd{0.462}{0.273} & \msd{0.080}{0.007} & \msd{0.040}{0.000} & \msd{0.216}{0.005} & \msd{0.146}{0.009} & \msd{0.714}{0.005} & \msd{0.304}{0.043} & \msd{3.692}{0.854}  \\[15pt]
 & farming         & \msd{0.060}{0.000} & \msd{0.094}{0.005}     & \msd{0.774}{0.061} & \msd{5.956}{2.159}   & \msd{5.300}{1.218}  & \msd{52.742}{24.775}   & \msd{0.058}{0.019} & \msd{0.038}{0.004} & \msd{0.010}{0.000} & \msd{0.168}{0.015} & \msd{0.092}{0.004} & \msd{0.468}{0.004} & \msd{0.876}{0.473} & \msd{4.414}{0.617}  \\[15pt]
 & wine         & \msd{0.072}{0.004} & \msd{0.126}{0.030}     & \msd{1.112}{0.334} & \msd{7.052}{1.212}   & \msd{3.732}{0.630}  & \msd{48.372}{8.541}    & \msd{0.794}{1.389} & \msd{0.042}{0.004} & \msd{0.010}{0.000} & \msd{0.150}{0.026} & \msd{0.096}{0.005} & \msd{0.500}{0.000} & \msd{0.910}{0.043} & \msd{3.960}{1.604}  \\[15pt]
 & bike         & \msd{0.382}{0.004} & \msd{0.236}{0.005}     & \msd{3.270}{0.092} & \msd{38.930}{11.993} & \msd{29.996}{7.638} & \msd{146.224}{70.532}  & \msd{4.976}{0.743} & \msd{1.004}{0.025} & \msd{0.030}{0.000} & \msd{0.204}{0.017} & \msd{0.150}{0.000} & \msd{0.694}{0.011} & \msd{8.200}{0.444} & \msd{13.546}{0.031} \\[15pt]
 & housing      & \msd{0.150}{0.007} & \msd{0.494}{0.009}     & \msd{4.612}{0.278} & \msd{40.334}{6.487}  & \msd{14.344}{4.544} & \msd{198.586}{70.990}  & \msd{1.832}{1.058} & \msd{0.526}{0.021} & \msd{0.060}{0.000} & \msd{0.328}{0.008} & \msd{0.228}{0.019} & \msd{0.844}{0.005} & \msd{7.048}{1.939} & \msd{14.226}{0.552} \\[15pt]
 & diamond      & \msd{1.776}{0.042} & \msd{0.846}{0.011}     & \msd{2.754}{0.155} & \msd{55.600}{9.001}  & \msd{16.248}{6.282} & \msd{319.462}{113.938} & \msd{1.218}{1.083} & \msd{1.820}{0.046} & \msd{0.120}{0.000} & \msd{0.644}{0.011} & \msd{0.396}{0.005} & \msd{1.066}{0.005} & \msd{7.496}{3.824} & \msd{31.076}{1.638}\\[15pt] \midrule
\multicolumn{2}{l}{Average runtime} & \multicolumn{1}{c}{6.131} & \multicolumn{1}{c}{41.878} & \multicolumn{1}{c}{4.640}  & \multicolumn{1}{c}{35.890}  & \multicolumn{1}{c}{22.685} & \multicolumn{1}{c}{179.900} & \multicolumn{1}{c}{3.096}  & \multicolumn{1}{c}{0.489} & \multicolumn{1}{c}{0.078} & \multicolumn{1}{c}{0.251} & \multicolumn{1}{c}{0.311} & \multicolumn{1}{c}{1.052} & \multicolumn{1}{c}{3.547} & \multicolumn{1}{c}{12.097}  \\ \bottomrule
\end{tabular}%
}
\end{table}

\begin{table}[htp!]
\caption{Mean training runtimes per fold, measured in seconds, using the best hyperparameter set identified by hyperparameter grid search. Models employing GPU acceleration are annotated with `**'.}
\label{tab:tuned_hp_training_time}
\resizebox{1\textwidth}{!}{%
\begin{tabular}{@{}l@{\hspace{0.1cm}}lllllllllllllll@{}}
\toprule
& & \multicolumn{9}{c}{\textbf{Interpretable Models}} & \multicolumn{5}{c}{\textbf{Black-box Models}} \\ \cmidrule(l){3-11} \cmidrule(l){12-16} 
& & \multicolumn{7}{c}{\textbf{GAMs}} & \multicolumn{2}{c}{\textbf{Traditional}} & & & \\ \cmidrule(l){3-9} \cmidrule(l){10-11}
\textbf{Task} & \textbf{Dataset} & \rotatebox{45}{\textbf{P-Splines}} & \rotatebox{45}{\textbf{TP-Splines}} & \rotatebox{45}{\textbf{EBM}} & \rotatebox{45}{\textbf{NAM}$^{**}$} & \rotatebox{45}{\textbf{GAMI-Net}} & \rotatebox{45}{\textbf{ExNN}} & \rotatebox{45}{\textbf{IGANN}$^{**}$} & \rotatebox{45}{\textbf{LR}} & \rotatebox{45}{\textbf{DT}} & \rotatebox{45}{\textbf{RF}} & \rotatebox{45}{\textbf{XGB}}  & \rotatebox{45}{\textbf{CatBoost}$^{**}$} & \rotatebox{45}{\textbf{MLP}} & \rotatebox{45}{\textbf{TabNet}$^{**}$} \\ \midrule
\multicolumn{1}{l}{\multirow{25}{*}{CLS}} 
& college      & \msd{0.048}{0.020}   & \msd{0.068}{0.025}     & \msd{0.744}{0.213}   & \msd{6.402}{2.953}     & \msd{4.744}{2.004}   & \msd{4.692}{0.494}      & \msd{3.906}{6.873}   & \msd{0.004}{0.005} & \msd{0.000}{0.000} & \msd{0.458}{0.440} & \msd{0.564}{0.401}   & \msd{0.202}{0.103}   & \msd{0.384}{0.106}   & \msd{9.158}{5.672}  \\[15pt]
 & water        & \msd{0.044}{0.019}   & \msd{0.654}{0.400}     & \msd{0.974}{0.306}   & \msd{4.122}{1.915}     & \msd{2.472}{0.587}   & \msd{39.524}{8.172}     & \msd{0.440}{0.324}   & \msd{0.002}{0.004} & \msd{0.006}{0.005} & \msd{1.822}{1.373} & \msd{0.132}{0.126}   & \msd{0.294}{0.223}   & \msd{0.654}{0.144}   & \msd{5.260}{1.736}   \\[15pt]
 & stroke       & \msd{0.516}{0.598}   & \textit{\msd{0.315}{0.048}}     & \msd{0.334}{0.062}   & \msd{5.234}{4.110}     & \msd{1.494}{0.436}   & \msd{98.268}{26.452}    & \msd{0.104}{0.059}   & \msd{0.000}{0.000} & \msd{0.000}{0.000} & \msd{0.130}{0.037} & \msd{0.028}{0.011}   & \msd{0.200}{0.331}   & \msd{0.664}{0.599}   & \msd{6.062}{1.949} \\[15pt]
 & churn        & \msd{0.622}{0.569}   & \msd{0.622}{0.223}     & \msd{1.466}{0.209}   & \msd{55.044}{25.724}   & \msd{9.240}{6.325}   & \msd{74.236}{27.039}    & \msd{0.934}{1.028}   & \msd{0.212}{0.452} & \msd{0.006}{0.005} & \msd{0.886}{1.140} & \msd{0.440}{0.414}   & \msd{0.306}{0.180}   & \msd{1.254}{0.227}   & \msd{9.290}{4.306}  \\[15pt]
 & recidivism   & \msd{0.468}{0.416}   & \msd{1.682}{1.047}    & \msd{0.928}{0.360}   & \msd{25.956}{5.313}    & \msd{3.722}{1.931}   & \msd{66.182}{17.069}    & \msd{1.772}{1.752}   & \msd{0.016}{0.013} & \msd{0.000}{0.000} & \msd{1.610}{1.272} & \msd{0.120}{0.148}   & \msd{0.214}{0.239}   & \msd{1.982}{0.886}   & \msd{6.380}{2.430}  \\[15pt]
 & credit       & \msd{1.122}{0.685}   & \msd{40.536}{50.561}   & \msd{0.934}{0.450}   & \msd{34.156}{16.588}   & \msd{12.128}{4.288}  & \msd{121.332}{13.374}   & \msd{0.280}{0.208}   & \msd{0.626}{0.535} & \msd{0.010}{0.007} & \msd{1.348}{1.358} & \msd{0.188}{0.226}   & \msd{2.458}{3.216}   & \msd{2.390}{1.033}   & \msd{6.036}{3.838}  \\[15pt]
 & income       & \msd{49.282}{1.340}  & \textit{\msd{194.680}{97.156}}   & \msd{24.608}{6.698}  & \msd{205.312}{16.230}  & \msd{56.752}{5.328}  & \msd{376.532}{89.886}   & \msd{18.514}{21.160} & \msd{0.664}{0.448} & \msd{0.040}{0.000} & \msd{0.724}{0.669} & \msd{0.944}{0.823}   & \msd{1.300}{0.559}   & \msd{9.044}{4.016}   & \msd{14.894}{3.712} \\[15pt]
 & bank         & \msd{5.735}{3.529}   & \msd{11.358}{3.462}    & \msd{5.956}{1.075}   & \msd{154.650}{91.562}  & \msd{29.852}{31.330} & \msd{608.956}{393.782}  & \msd{4.508}{5.522}   & \msd{1.196}{0.824} & \msd{0.034}{0.019} & \msd{0.656}{0.325} & \msd{2.996}{3.689}   & \msd{4.156}{3.082}   & \msd{6.838}{4.232}   & \msd{36.652}{16.486} \\[15pt]
 & airline      & \msd{11.032}{6.639}  & \msd{168.092}{18.842}  & \msd{41.740}{22.439} & \msd{261.484}{69.631}  & \msd{91.750}{31.515} & \msd{2,763.774}{900.049} & \msd{71.860}{1.233}  & \msd{0.176}{0.098} & \msd{0.114}{0.060} & \msd{1.524}{1.037} & \msd{13.852}{13.764} & \msd{4.766}{2.507}   & \msd{27.864}{10.550} & \msd{46.466}{1.229}  \\[15pt]
 & weather      & \msd{56.694}{21.307} & \msd{781.550}{271.845} & \msd{81.090}{43.050} & \msd{760.092}{188.613} & \msd{70.510}{17.655} & \msd{856.648}{395.645}  & \msd{53.626}{47.748} & \msd{7.990}{6.477} & \msd{0.386}{0.065} & \msd{6.258}{3.098} & \msd{59.184}{17.816} & \msd{35.866}{25.969} & \msd{39.212}{29.197} & \msd{58.414}{10.197}  \\[15pt] \midrule
\multicolumn{1}{l}{\multirow{25}{*}{REG}} 
 & car          & \msd{0.100}{0.007}   & \msd{1.006}{1.664}        & \msd{2.084}{0.944}   & \msd{45.140}{11.152}   & \msd{4.844}{2.041}   & \msd{3.342}{0.320}      & \msd{1.038}{1.046}   & \msd{0.004}{0.005} & \msd{0.000}{0.000} & \msd{0.168}{0.111} & \msd{0.096}{0.104}   & \msd{0.090}{0.035}   & \msd{0.222}{0.134}   & \msd{5.612}{1.381}  \\[15pt]
 & student      & \msd{0.216}{0.214}   & \msd{0.198}{0.144}        & \msd{0.288}{0.106}   & \msd{30.134}{10.291}   & \msd{1.392}{0.599}   & \msd{3.608}{0.239}      & \msd{0.212}{0.048}   & \msd{0.006}{0.005} & \msd{0.000}{0.000} & \msd{0.836}{0.729} & \msd{0.070}{0.069}   & \msd{0.198}{0.198}   & \msd{0.146}{0.055}   & \msd{4.156}{1.795} \\[15pt]
 & productivity & \msd{0.358}{0.072}   & \msd{0.612}{0.372}        & \msd{1.198}{0.303}   & \msd{32.564}{8.022}    & \msd{8.236}{3.858}   & \msd{13.344}{2.845}     & \msd{0.860}{0.677}   & \msd{0.014}{0.019} & \msd{0.000}{0.000} & \msd{0.568}{0.512} & \msd{0.124}{0.149}   & \msd{0.332}{0.398}   & \msd{0.438}{0.089}   & \msd{5.906}{3.796}  \\[15pt]
 & insurance    & \msd{0.026}{0.009}   & \msd{0.024}{0.009}        & \msd{0.534}{0.155}   & \msd{7.154}{1.774}     & \msd{3.590}{1.837}   & \msd{20.862}{9.447}     & \msd{0.758}{0.747}   & \msd{0.000}{0.000} & \msd{0.000}{0.000} & \msd{1.298}{0.938} & \msd{0.020}{0.000}   & \msd{0.140}{0.045}   & \msd{0.518}{0.207}   & \msd{10.396}{3.999}  \\[15pt]
 & crimes       & \msd{0.260}{0.000}   & \msd{233.988}{243.168}    & \msd{2.646}{0.923}   & \msd{101.886}{29.574}  & \msd{26.068}{13.399} & \msd{22.480}{2.501}     & \msd{0.392}{0.049}   & \msd{0.014}{0.026} & \msd{0.010}{0.000} & \msd{1.770}{0.532} & \msd{0.364}{0.455}   & \msd{0.454}{0.271}   & \msd{0.600}{0.145}   & \msd{10.548}{5.343}  \\[15pt]
 & farming      & \msd{0.026}{0.009}   & \msd{0.124}{0.064}        & \msd{1.860}{0.951}   & \msd{17.902}{2.796}    & \msd{5.116}{1.281}   & \msd{36.856}{12.188}    & \msd{0.260}{0.105}   & \msd{0.002}{0.004} & \msd{0.000}{0.000} & \msd{1.830}{0.950} & \msd{0.296}{0.339}   & \msd{0.484}{0.570}   & \msd{1.530}{0.539}   & \msd{8.810}{6.285}  \\[15pt]
 & wine         & \msd{0.066}{0.029}   & \msd{0.418}{0.227}        & \msd{1.792}{0.770}   & \msd{26.086}{4.823}    & \msd{3.658}{0.495}   & \msd{81.686}{43.962}    & \msd{0.740}{0.641}   & \msd{0.018}{0.025} & \msd{0.006}{0.005} & \msd{1.140}{0.291} & \msd{1.018}{1.310}   & \msd{15.408}{0.194}  & \msd{1.650}{0.273}   & \msd{6.722}{1.803}  \\[15pt]
 & bike         & \msd{0.524}{0.015}   & \msd{0.320}{0.012}        & \msd{19.240}{7.616}  & \msd{103.890}{28.892}  & \msd{23.290}{2.444}  & \msd{230.246}{139.197}  & \msd{22.878}{0.105}  & \msd{0.604}{0.497} & \msd{0.026}{0.005} & \msd{1.676}{0.408} & \msd{3.740}{2.528}   & \msd{4.684}{3.114}   & \msd{12.924}{1.232}  & \msd{14.108}{0.137}  \\[15pt]
 & housing      & \msd{0.190}{0.007}   & \msd{0.726}{0.030}        & \msd{24.530}{8.950}  & \msd{115.898}{14.997}  & \msd{13.484}{4.659}  & \msd{515.472}{190.438}  & \msd{9.822}{7.356}   & \msd{0.508}{0.024} & \msd{0.040}{0.000} & \msd{2.394}{0.928} & \msd{3.598}{1.234}   & \msd{2.418}{0.103}   & \msd{11.446}{1.589}  & \msd{15.256}{0.066}  \\[15pt]
 & diamond      & \msd{1.964}{0.676}   & \msd{0.912}{0.126}        & \msd{5.526}{2.122}   & \msd{283.598}{64.906}  & \msd{19.702}{3.776}  & \msd{439.006}{107.119}  & \msd{15.034}{14.431} & \msd{1.132}{0.878} & \msd{0.086}{0.033} & \msd{6.336}{0.021} & \msd{8.776}{2.908}   & \msd{8.260}{9.167}   & \msd{28.010}{12.743} & \msd{31.188}{1.786} \\[15pt] \midrule
\multicolumn{2}{l}{Average runtime} & \multicolumn{1}{c}{6.465} & \multicolumn{1}{c}{71.894} & \multicolumn{1}{c}{10.924} & \multicolumn{1}{c}{113.835} & \multicolumn{1}{c}{19.602} & \multicolumn{1}{c}{318.852} & \multicolumn{1}{c}{10.397} & \multicolumn{1}{c}{0.659} & \multicolumn{1}{c}{0.038} & \multicolumn{1}{c}{1.672} & \multicolumn{1}{c}{4.828} & \multicolumn{1}{c}{4.112} & \multicolumn{1}{c}{7.389} & \multicolumn{1}{c}{15.566}  \\ \bottomrule
\end{tabular}%
}
\end{table}

\begin{table}[htp!]
\caption{Mean hyperparameter tuning runtimes per fold, calculated in seconds, based on the average runtime across the defined set of hyperparameter candidates for each model. Models employing GPU acceleration are denoted with `**'.}
\label{tab:tuning_runtime_mean_model}
\sisetup{group-digits=true, group-separator={,}, group-minimum-digits=4}
\resizebox{1\textwidth}{!}{%
\begin{tabular}{@{}l@{\hspace{0.1cm}}lS[table-format=6]lllllllllllllll@{}}
\toprule
& & & & \multicolumn{9}{c}{\textbf{Interpretable Models}} & \multicolumn{5}{c}{\textbf{Black-box Models}} \\ \cmidrule(l){5-13} \cmidrule(l){14-18} 
& & & & \multicolumn{7}{c}{\textbf{GAMs}} & \multicolumn{2}{c}{\textbf{Traditional}} & & & \\ \cmidrule(l){5-11} \cmidrule(l){12-13}
\textbf{Task} & \textbf{Dataset} & \textbf{\# samples} & \textbf{\# feat}  & \rotatebox{45}{\textbf{P-Splines}} & \rotatebox{45}{\textbf{TP-Splines}} & \rotatebox{45}{\textbf{EBM}} & \rotatebox{45}{\textbf{NAM}$^{**}$} & \rotatebox{45}{\textbf{GAMI-Net}} & \rotatebox{45}{\textbf{ExNN}} & \rotatebox{45}{\textbf{IGANN}$^{**}$} & \rotatebox{45}{\textbf{LR}} & \rotatebox{45}{\textbf{DT}} & \rotatebox{45}{\textbf{RF}} & \rotatebox{45}{\textbf{XGB}}  & \rotatebox{45}{\textbf{CatBoost}$^{**}$} & \rotatebox{45}{\textbf{MLP}} & \rotatebox{45}{\textbf{TabNet}$^{**}$} \\ \midrule
\multicolumn{1}{l}{\multirow{25}{*}{CLS}} 
& college      & 1000       & 14      & \msd{0.114}{0.005}  & \msd{0.104}{0.015}     & \msd{0.528}{0.036}  & \msd{6.870}{0.486}    & \msd{4.694}{1.063}   & \msd{4.552}{0.078}      & \msd{1.480}{1.643}   & \msd{0.010}{0.000} & \msd{0.000}{0.000} & \msd{0.898}{0.004} & \msd{0.236}{0.009}  & \msd{1.222}{0.008} & \msd{0.320}{0.014}  & \msd{7.216}{0.640} \\[15pt]
 & water        & 3276       & 9       & \msd{0.080}{0.007}  & \msd{1.532}{0.127}     & \msd{0.746}{0.039}  & \msd{5.434}{0.505}    & \msd{1.716}{0.276}   & \msd{42.838}{1.517}     & \msd{0.530}{0.233}   & \msd{0.010}{0.000} & \msd{0.000}{0.000} & \msd{0.962}{0.035} & \msd{0.508}{0.004}  & \msd{2.354}{0.011} & \msd{0.692}{0.031}  & \msd{6.244}{0.630} \\[15pt]
 & stroke       & 5110       & 19      & \msd{0.774}{0.057}  & \textit{\msd{0.445}{0.030}}     & \msd{0.516}{0.066}  & \msd{16.958}{2.608}   & \msd{3.538}{0.476}   & \msd{114.158}{11.828}   & \msd{0.158}{0.004}   & \msd{0.040}{0.000} & \msd{0.000}{0.000} & \msd{1.010}{0.007} & \msd{0.480}{0.000}  & \msd{1.240}{0.046} & \msd{0.920}{0.085}  & \msd{6.250}{0.296} \\[15pt]
 & churn        & 7043       & 40      & \msd{1.196}{0.096}  & \msd{0.802}{0.016}     & \msd{1.112}{0.073}  & \msd{44.678}{2.596}   & \msd{10.146}{1.757}  & \msd{90.984}{8.814}     & \msd{0.548}{0.242}   & \msd{0.286}{0.005} & \msd{0.010}{0.000} & \msd{1.048}{0.013} & \msd{0.906}{0.005}  & \msd{2.950}{0.007} & \msd{1.566}{0.059}  & \msd{8.482}{0.261} \\[15pt]
 & recidivism   & 7214       & 18      & \msd{0.508}{0.204}  & \msd{2.062}{0.255}     & \msd{0.854}{0.038}  & \msd{20.996}{1.694}   & \msd{5.484}{3.240}   & \msd{84.912}{6.722}     & \msd{2.398}{2.109}   & \msd{0.050}{0.000} & \msd{0.000}{0.000} & \msd{1.070}{0.007} & \msd{0.588}{0.008}  & \msd{1.224}{0.005} & \msd{1.592}{0.054}  & \msd{8.036}{0.595} \\[15pt]
 & credit       & 10459      & 37      & \msd{2.270}{1.015}  & \msd{48.224}{10.944}    & \msd{1.250}{0.080}  & \msd{38.438}{2.218}   & \msd{9.314}{2.731}   & \msd{108.910}{5.073}    & \msd{0.372}{0.113}   & \msd{0.328}{0.004} & \msd{0.010}{0.000} & \msd{0.948}{0.018} & \msd{1.006}{0.005}  & \msd{1.888}{0.013} & \msd{1.864}{0.042}  & \msd{9.414}{0.890} \\[15pt]
 & income       & 32561      & 65      & \msd{24.138}{0.834} & \textit{\msd{53.360}{7.580}}     & \msd{22.456}{1.744} & \msd{156.928}{4.666}  & \msd{31.846}{3.360}  & \msd{360.332}{26.394}   & \msd{12.808}{8.602}  & \msd{0.860}{0.007} & \msd{0.030}{0.000} & \msd{0.676}{0.005} & \msd{6.636}{0.160}  & \msd{3.860}{0.029} & \msd{6.384}{0.149}  & \msd{25.570}{0.614} \\[15pt]
 & bank         & 45211      & 47      & \msd{7.547}{1.113}  & \msd{15.404}{2.211}    & \msd{6.168}{0.440}  & \msd{158.174}{8.927}  & \msd{20.444}{2.540}  & \msd{422.112}{28.803}   & \msd{5.442}{4.047}   & \msd{0.916}{0.202} & \msd{0.040}{0.000} & \msd{0.674}{0.009} & \msd{6.696}{0.060}  & \msd{4.900}{0.056} & \msd{8.426}{0.279}  & \msd{26.616}{1.145} \\[15pt]
 & airline      & 103904     & 24      & \msd{8.366}{0.122}  & \msd{176.934}{6.774}  & \msd{33.612}{0.427} & \msd{184.962}{4.712}  & \msd{73.186}{31.464} & \msd{1,761.734}{152.080} & \msd{38.408}{0.920}  & \msd{0.482}{0.050} & \msd{0.070}{0.000} & \msd{1.042}{0.015} & \msd{5.398}{0.022}  & \msd{4.400}{0.119} & \msd{31.186}{0.501} & \msd{67.586}{2.638} \\[15pt]
 & weather      & 142193     & 70      & \msd{39.462}{3.042} & \msd{450.410}{37.688}   & \msd{73.452}{2.388} & \msd{482.628}{17.371} & \msd{56.066}{6.096}  & \msd{1,092.328}{51.549}  & \msd{43.514}{13.954} & \msd{1.952}{0.043} & \msd{0.226}{0.005} & \msd{2.504}{0.030} & \msd{13.464}{0.021} & \msd{6.326}{0.105} & \msd{43.764}{0.575} & \msd{84.370}{2.931} \\[15pt] \midrule
\multicolumn{1}{l}{\multirow{25}{*}{REG}} 
 & car          & 201        & 76      & \msd{0.502}{0.024}  & \msd{1.024}{0.253}     & \msd{1.596}{0.434}  & \msd{29.478}{3.620}   & \msd{4.222}{1.541}   & \msd{3.314}{0.036}      & \msd{0.732}{0.365}   & \msd{0.000}{0.000} & \msd{0.000}{0.000} & \msd{0.700}{0.000} & \msd{0.218}{0.004}  & \msd{0.980}{0.449} & \msd{0.136}{0.023}  & \msd{6.160}{0.373} \\[15pt]
 & student      & 649        & 43      & \msd{0.230}{0.000}  & \msd{0.144}{0.015}      & \msd{0.416}{0.049}  & \msd{24.002}{1.434}   & \msd{3.064}{1.023}   & \msd{3.786}{0.068}      & \msd{0.466}{0.373}   & \msd{0.000}{0.000} & \msd{0.000}{0.000} & \msd{0.740}{0.000} & \msd{0.318}{0.004}  & \msd{1.922}{0.040} & \msd{0.162}{0.013}  & \msd{5.394}{0.954} \\[15pt]
 & productivity & 1197       & 35      & \msd{0.218}{0.004}  & \msd{0.334}{0.106}      & \msd{0.982}{0.113}  & \msd{23.710}{3.431}   & \msd{7.164}{1.688}   & \msd{14.944}{0.830}     & \msd{2.254}{1.770}   & \msd{0.000}{0.000} & \msd{0.000}{0.000} & \msd{0.804}{0.005} & \msd{0.390}{0.000}  & \msd{0.938}{0.008} & \msd{0.366}{0.027}  & \msd{6.596}{0.865} \\[15pt]
 & insurance    & 1338       & 9       & \msd{0.030}{0.000}  & \msd{0.020}{0.000}     & \msd{0.472}{0.018}  & \msd{5.416}{0.328}    & \msd{3.280}{0.695}   & \msd{14.154}{1.359}     & \msd{0.950}{0.256}   & \msd{0.000}{0.000} & \msd{0.000}{0.000} & \msd{0.766}{0.009} & \msd{0.300}{0.000}  & \msd{0.708}{0.004} & \msd{0.442}{0.052}  & \msd{8.754}{0.289} \\[15pt]
 & crimes       & 1994       & 100     & \msd{2.386}{0.009}  & \msd{377.098}{50.271} & \msd{1.362}{0.124}  & \msd{77.484}{5.881}   & \msd{12.470}{1.901}  & \msd{22.220}{0.946}     & \msd{0.464}{0.030}   & \msd{0.010}{0.000} & \msd{0.010}{0.000} & \msd{0.674}{0.015} & \msd{0.648}{0.004}  & \msd{5.114}{0.046} & \msd{0.434}{0.009}  & \msd{7.362}{0.589} \\[15pt]
 & farming      & 3893       & 10      & \msd{0.050}{0.000}  & \msd{0.122}{0.004}     & \msd{1.148}{0.117}  & \msd{15.644}{1.366}   & \msd{4.078}{1.689}   & \msd{50.018}{7.030}     & \msd{0.398}{0.179}   & \msd{0.000}{0.000} & \msd{0.000}{0.000} & \msd{0.720}{0.010} & \msd{0.554}{0.005}  & \msd{1.502}{0.016} & \msd{1.074}{0.110}  & \msd{8.700}{0.486} \\[15pt]
 & wine         & 4898       & 11      & \msd{0.060}{0.000}  & \msd{0.250}{0.061}      & \msd{1.194}{0.132}  & \msd{20.184}{0.464}   & \msd{2.788}{0.642}   & \msd{67.040}{3.995}     & \msd{0.692}{0.658}   & \msd{0.006}{0.005} & \msd{0.000}{0.000} & \msd{0.454}{0.005} & \msd{0.558}{0.004}  & \msd{1.660}{0.012} & \msd{1.294}{0.059}  & \msd{9.602}{0.832} \\[15pt]
 & bike         & 17379      & 7       & \msd{0.290}{0.000}  & \msd{0.260}{0.000}     & \msd{7.328}{0.306}  & \msd{75.680}{3.939}   & \msd{24.500}{2.099}  & \msd{209.602}{20.033}   & \msd{14.524}{0.102}  & \msd{0.110}{0.000} & \msd{0.010}{0.000} & \msd{0.618}{0.004} & \msd{1.042}{0.004}  & \msd{0.892}{0.004} & \msd{8.462}{0.412}  & \msd{20.786}{0.540} \\[15pt]
 & housing      & 20640      & 8       & \msd{0.120}{0.000}  & \msd{0.570}{0.007}      & \msd{10.048}{0.375} & \msd{74.674}{3.999}   & \msd{7.938}{0.947}   & \msd{331.488}{28.005}   & \msd{8.118}{1.896}   & \msd{0.060}{0.000} & \msd{0.012}{0.004} & \msd{1.034}{0.005} & \msd{1.632}{0.004}  & \msd{2.364}{0.030} & \msd{8.700}{0.329}  & \msd{22.220}{0.639} \\[15pt]
 & diamond      & 53943      & 26      & \msd{1.346}{0.009}  & \msd{0.898}{0.004}      & \msd{4.410}{0.124}  & \msd{177.624}{6.447}  & \msd{19.190}{3.213}  & \msd{407.630}{13.079}   & \msd{8.176}{8.006}   & \msd{0.190}{0.000} & \msd{0.030}{0.000} & \msd{1.916}{0.005} & \msd{2.956}{0.009}  & \msd{2.080}{0.016} & \msd{29.800}{4.757} & \msd{43.812}{1.863} \\[15pt] \midrule
\multicolumn{4}{l}{Sum (in sec) across all datasets}  & \multicolumn{1}{c}{89.687} & \multicolumn{1}{r}{1,129.997} & \multicolumn{1}{r}{169.650} & \multicolumn{1}{r}{1,639.962} & \multicolumn{1}{r}{305.128} & \multicolumn{1}{r}{5,207.056} & \multicolumn{1}{r}{142.432} & \multicolumn{1}{r}{5.310} & \multicolumn{1}{r}{0.448} & \multicolumn{1}{r}{19.258} & \multicolumn{1}{r}{44.534} & \multicolumn{1}{r}{48.524} & \multicolumn{1}{r}{147.584} & \multicolumn{1}{r}{389.170} \\ 
\multicolumn{4}{l}{Average (in sec) across all datasets} & \multicolumn{1}{r}{4.484}  & \multicolumn{1}{r}{56.500}  & \multicolumn{1}{r}{8.483}   & \multicolumn{1}{r}{81.998}   & \multicolumn{1}{r}{15.256}  & \multicolumn{1}{r}{260.353}  & \multicolumn{1}{r}{7.122}   & \multicolumn{1}{r}{0.266} & \multicolumn{1}{r}{0.022} & \multicolumn{1}{r}{0.963}  & \multicolumn{1}{r}{2.227}  & \multicolumn{1}{r}{2.426}  & \multicolumn{1}{r}{7.379}   & \multicolumn{1}{r}{19.459} \\ \bottomrule
\end{tabular}%
}
\end{table}

\begin{table}[htp!]
\sisetup{group-digits=true, group-separator={,}, group-minimum-digits=4}
\caption{Overview of hyperparameter tuning runtimes in hours.}
\label{tab:tuning_runtime_overview}
\resizebox{\textwidth}{!}{%
\begin{tabular}{
l
S[table-format=4.3]
S[table-format=4.3]
S[table-format=4.3]
S[table-format=4.3]
S[table-format=6.3]
S[table-format=4.3]
S[table-format=4.3]
S[table-format=4.3]
S[table-format=4.3]
S[table-format=4.3]
S[table-format=4.3]
S[table-format=4.3]
S[table-format=4.3]
S[table-format=4.3]@{\hspace{0.5cm}}
S[table-format=7.3]<{\hspace{0pt}}
}
\toprule
 & \multicolumn{9}{c}{\textbf{Interpretable Models}} & \multicolumn{4}{c}{\textbf{Black-box Models}} &  \\ \cmidrule(l){2-10} \cmidrule(l){11-15} 
& \multicolumn{7}{c}{\textbf{GAMs}} & \multicolumn{2}{c}{\textbf{Traditional}} & & & \\ \cmidrule(l){2-8} \cmidrule(l){9-10}
 & \textbf{P-Splines} & \textbf{TP-Splines} & \textbf{EBM} & \textbf{NAM} & \textbf{GAMI-Net} & \textbf{ExNN} & \textbf{IGANN} & \textbf{LR} & \textbf{DT} & \textbf{RF} & \textbf{XGB} & \textbf{CatBoost} & \textbf{MLP} & \textbf{TabNet} & Sum \\ \cmidrule(l){1-15} \cmidrule(l){16-16} 
\begin{tabular}{@{}l@{}} Total runtime for all datasets, \\ 5-folds, and candidates\end{tabular} & 2.49 & 28.25 & 5.66 & 54.67 & 11.44 & 130.18 & 3.56 & 1.63 & 0.06 & 1.04 & 5.57 & 5.39 & 14.76 & 14.59 & 279.28 \\  
 \bottomrule
\end{tabular}%
}
\end{table}

\newpage
\clearpage

\section{Additional Experiments with Spline-based Models}
\label{app:additional_experiments}

Given the multitude of spline-based approaches for \glspl{gam}, a series of preliminary investigations and experiments were conducted to identify the most suitable ones based on the mgcv package written in R \citep{wood_mgcv_2023}. The final selection was made between cubic regression splines (\textit{CR-Splines}), cubic regression splines with shrinkage (\textit{CS-Splines}), and thin plate regression splines with smoothing penalty (\textit{TP-Splines}). The performance of all three spline-based variants was then evaluated for different numbers of basis functions (i.e., basis dimension \textit{k}), using \textit{k} = 5, 10, and 20, as shown in Table~\ref{tab:splines_predictive_performance}. 

\begin{table}[htpb!]
\caption{Predictive performance using different spline-based \glspl{gam} with varying numbers of basis functions (basis dimension). Classification tasks are assessed using \gls{auroc}, whereas regression tasks are measured using \gls{rmse}.}
\label{tab:splines_predictive_performance}
\resizebox{1\textwidth}{!}{%
\begin{tabular}{@{}l@{\hspace{0.1cm}}llllllllll@{}}
\toprule
\textbf{Task} & \textbf{Dataset} & \textbf{TP-Splines$_{5}$} & \textbf{CS-Splines$_{5}$} & \textbf{CR-Splines$_{5}$} & \textbf{TP-Splines$_{10}$} & \textbf{CS-Splines$_{10}$} & \textbf{CR-Splines$_{10}$} & \textbf{TP-Splines$_{20}$} & \textbf{CS-Splines$_{20}$} & \textbf{CR-Splines$_{20}$} \\ \midrule
\multicolumn{1}{l}{\multirow{25}{*}{CLS}} & college      & \msd{0.950}{0.007}   & \msd{0.950}{0.006}   & \msd{0.950}{0.007}   & \textbf{\msd{0.951}{0.006}}    & \msd{0.950}{0.009}    & \msd{0.950}{0.009}    & \msd{0.951}{0.008}    & \msd{0.946}{0.008}    & \msd{0.947}{0.008}    \\[15pt]
 & water        & \textbf{\msd{0.588}{0.021}}   & \textbf{\msd{0.588}{0.021}}   & \msd{0.583}{0.022}   & \msd{0.585}{0.015}    & \msd{0.579}{0.020}    & \msd{0.574}{0.020}    & \msd{0.579}{0.010}    & \msd{0.577}{0.015}    & \msd{0.573}{0.015}    \\[15pt]
 & stroke       & \msd{0.828}{0.022}   & \msd{0.834}{0.024}   & \msd{0.833}{0.022}   & \msd{0.835}{0.024}    & \msd{0.833}{0.021}    & \msd{0.833}{0.021}    & \textbf{\msd{0.836}{0.024}}    & \msd{0.832}{0.025}    & \msd{0.835}{0.028}    \\[15pt]
 & churn        & \msd{0.849}{0.013}   & \msd{0.849}{0.013}   & \msd{0.849}{0.012}   & \msd{0.849}{0.012}    & \msd{0.849}{0.012}    & \msd{0.850}{0.012}    & \msd{0.849}{0.012}    & \msd{0.849}{0.012}    & \textbf{\msd{0.850}{0.011}}    \\[15pt]
 & recidivism       & \msd{0.740}{0.017}   & \msd{0.740}{0.017}   & \msd{0.740}{0.017}   & \msd{0.744}{0.018}    & \msd{0.745}{0.016}    & \msd{0.745}{0.016}    & \msd{0.745}{0.017}    & \textbf{\msd{0.746}{0.016}}    & \textbf{\msd{0.746}{0.016}}    \\[15pt]
 & credit         & \msd{0.804}{0.010}   & \msd{0.803}{0.011}   & \msd{0.802}{0.011}   & \textbf{\msd{0.810}{0.009}}    & \msd{0.806}{0.011}    & \msd{0.806}{0.009}    & \textbf{\msd{0.810}{0.009}}    & \msd{0.808}{0.009}    & \msd{0.808}{0.009}    \\[15pt]
 & income        & \msd{0.917}{0.000}   & \msd{0.915}{0.002}   & \msd{0.915}{0.002}   & \msd{0.916}{0.002}    & \msd{0.920}{0.002}    & \msd{0.921}{0.002}    & \msd{0.914}{0.002}    & \textbf{\msd{0.925}{0.001}}    & \textbf{\msd{0.925}{0.001}}    \\[15pt]
 & bank         & \msd{0.775}{0.011}   & \msd{0.776}{0.012}   & \msd{0.776}{0.012}   & \msd{0.779}{0.011}    & \msd{0.780}{0.012}    & \msd{0.780}{0.012}    & \msd{0.780}{0.011}    & \textbf{\msd{0.781}{0.012}}    & \textbf{\msd{0.781}{0.012} }   \\[15pt]
 & airline      & \msd{0.978}{0.001}   & \msd{0.978}{0.001}   & \msd{0.978}{0.001}   & \textbf{\msd{0.980}{0.001}}    & \textbf{\msd{0.980}{0.001}}    & \textbf{\msd{0.980}{0.001}}    & \textbf{\msd{0.980}{0.001}}    & \textbf{\msd{0.980}{0.001}}    & \textbf{\msd{0.980}{0.001}}    \\[15pt]
 & weather      & \msd{0.873}{0.003}   & \msd{0.873}{0.002}   & \textbf{\msd{0.874}{0.002}}     & \msd{0.874}{0.003}    & \msd{0.874}{0.003}    & \msd{0.874}{0.003}    & \msd{0.874}{0.003}    & \msd{0.874}{0.003}    & \msd{0.874}{0.003}  \\[15pt] \midrule
\multicolumn{1}{l}{\multirow{25}{*}{REG}} & car          & \msd{0.475}{0.197}   & \msd{0.388}{0.127}   & \msd{0.393}{0.081}   & \msd{0.405}{0.113}  & \msd{0.584}{0.428}    & \textbf{\msd{0.384}{0.070}}   & \msd{0.529}{0.258}    & \msd{0.632}{0.486}    & \msd{0.584}{0.428}    \\[15pt]
 & student      & \textbf{\msd{0.859}{0.133}}   & \msd{0.868}{0.125}   & \msd{0.869}{0.122}   & \msd{0.863}{0.132}    & \msd{0.870}{0.122}    & \msd{0.869}{0.122}    & \msd{0.863}{0.132}    & \msd{0.867}{0.124}    & \msd{0.870}{0.122}    \\[15pt]
 & productivity & \msd{0.793}{0.068}   & \msd{0.776}{0.086}   & \msd{0.808}{0.123}   & \msd{0.789}{0.081}    & \msd{1.475}{1.634}    & \msd{1.286}{1.201}    & \textbf{\msd{0.764}{0.052}}    & \msd{1.735}{2.257}    & \msd{1.475}{1.634}    \\[15pt]
 & insurance      & \textbf{\msd{0.497}{0.019}}   & \msd{0.498}{0.018}   & \msd{0.498}{0.018}   & \textbf{\msd{0.497}{0.019} }   & \msd{0.499}{0.018}    & \textbf{\msd{0.497}{0.019}}    & \msd{0.498}{0.018}    & \msd{0.499}{0.018}    & \msd{0.499}{0.018}    \\[15pt]
 & crimes       & \textbf{\msd{0.572}{0.057}}   & \msd{0.574}{0.059}   & \msd{0.577}{0.059}   & \msd{0.586}{0.059}    & \msd{0.653}{0.067}    & \msd{0.608}{0.061}    & \msd{0.617}{0.079}    & \msd{0.647}{0.067}    & \msd{0.654}{0.067}    \\[15pt]
 & farming         & \msd{0.665}{0.043}   & \textbf{\msd{0.664}{0.043}}   & \msd{0.664}{0.042}   & \msd{0.665}{0.043}    & \msd{0.667}{0.042}    & \msd{0.665}{0.044}    & \msd{0.666}{0.044}    & \msd{0.667}{0.042}    & \msd{0.667}{0.042}    \\[15pt]
 & wine         & \msd{0.833}{0.048}   & \textbf{\msd{0.811}{0.024}}   & \textbf{\msd{0.811}{0.024}}   & \msd{0.815}{0.026}    & \msd{0.825}{0.054}    & \msd{0.824}{0.047}    & \msd{0.897}{0.195}    & \msd{0.826}{0.055}    & \msd{0.825}{0.054}    \\[15pt]
 & bike         & \msd{0.673}{0.016}   & \msd{0.667}{0.017}   & \msd{0.667}{0.017}   & \msd{0.581}{0.013}    & \msd{0.554}{0.010}    & \msd{0.569}{0.012}    & \textbf{\msd{0.553}{0.010}}    & \msd{0.554}{0.010}    & \msd{0.554}{0.010}    \\[15pt]
 & housing      & \msd{0.559}{0.017}   & \msd{0.558}{0.019}   & \msd{0.558}{0.019}   & \msd{0.536}{0.018}    & \textbf{\msd{0.520}{0.015}}    & \msd{0.534}{0.018}    & \msd{0.522}{0.016}    & \textbf{\msd{0.520}{0.015}}    & \textbf{\msd{0.520}{0.015}}    \\[15pt]
 & diamond      & \msd{0.330}{0.128}   & \msd{0.311}{0.070}   & \textbf{\msd{0.311}{0.069}}   & \msd{0.634}{0.724}    & \msd{0.549}{0.473}    & \msd{0.607}{0.574}    & \msd{0.316}{0.106}  & \msd{0.549}{0.473}    & \msd{0.549}{0.473} \\[15pt] \midrule
 \multicolumn{2}{l}{Average rank CLS} & \multicolumn{1}{c}{6.95}  & \multicolumn{1}{c}{6.40}   & \multicolumn{1}{c}{6.15}   & \multicolumn{1}{c}{3.85}   & \multicolumn{1}{c}{4.90}   & \multicolumn{1}{c}{4.60}   & \multicolumn{1}{c}{3.95}   & \multicolumn{1}{c}{4.50}  & \multicolumn{1}{c}{\textbf{3.70}} \\ 
\multicolumn{2}{l}{Average rank REG}  & \multicolumn{1}{c}{4.65}  & \multicolumn{1}{c}{\textbf{3.65}}   & \multicolumn{1}{c}{4.10}   & \multicolumn{1}{c}{4.30}   & \multicolumn{1}{c}{6.40}   & \multicolumn{1}{c}{4.75}   & \multicolumn{1}{c}{4.35}   & \multicolumn{1}{c}{6.30}  & \multicolumn{1}{c}{6.50} \\ \midrule
\multicolumn{2}{l}{Average rank total}  &  \multicolumn{1}{c}{5.80}  & \multicolumn{1}{c}{5.03}   & \multicolumn{1}{c}{5.13}   & \multicolumn{1}{c}{\textbf{4.08}}   & \multicolumn{1}{c}{5.65}   & \multicolumn{1}{c}{4.68}   & \multicolumn{1}{c}{4.15}   & \multicolumn{1}{c}{5.40} & \multicolumn{1}{c}{5.10} \\ \bottomrule
\end{tabular}%
}
\end{table}

\noindent As illustrated in the Table~\ref{tab:splines_predictive_performance}, all three types of splines exhibit comparable predictive performance. Notable deviations can be observed in some regression datasets, especially in the case of \textit{cars}, \textit{productivity}, \textit{crimes}, and \textit{diamonds}. Based on the total average rank, \gls{tpsplines} turns out to be the best-performing candidate, with TP-Splines$_{10}$ and TP-Splines$_{20}$ achieving scores of 4.08 and 4.15, respectively. For this reason, \gls{tpsplines} is included in our full model comparison. The corresponding hyperparameter tuning grid can be found in Appendix~\ref{app:hyperparameter}.

\newpage
\clearpage

\section{Visual Model Outputs}
\label{app:visual_output}

\subsection{Shape Plot Comparison with Confidence Bands}
\label{app:visual_output_confidence_bands}

\begin{figure}[htp!]
\centering
    \includegraphics[width=1\textwidth]{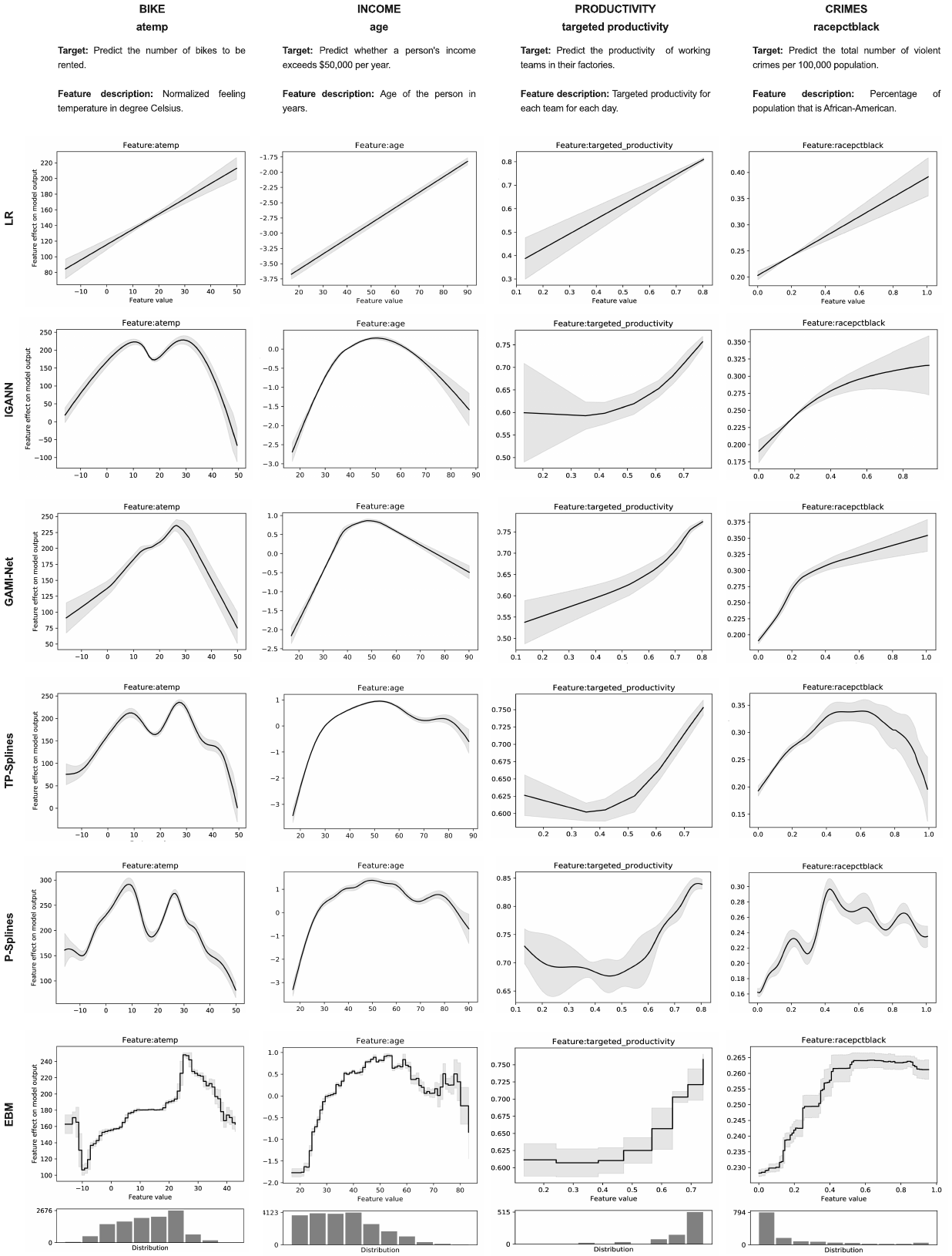}
    \caption{Comparison of shape plots learned by five different \glspl{gam} and a linear model including confidence bands.}
    \label{fig:shape_comparison_conf_bands}
\end{figure}

While the shape plots shown in Figure~\ref{fig:shape_comparison} in the main paper focus on the visual properties of the shape functions learned by different \gls{gam} variants/architectures, the shape plots in Figure~\ref{fig:shape_comparison_conf_bands} below also contain confidence bands that give an insight into model uncertainty/variability in certain feature regions. The confidence bands in the shape plots are generated based on a method following the idea of bagging. For this purpose, ten different subsets of the datasets are generated using ten different seeds.\footnote{As \gls{psplines} and \gls{tpsplines} do not converge for every seed, a suitable seed is sought until these models can also be trained ten times.} The resulting subsets are then used to train ten independent models for each \gls{gam} variant, resulting in ten shape plots for each \gls{gam}. The shape plots shown in black are based on the average of all ten shape plots. The confidence bands, shown in gray, represent the positive and negative standard deviations across all ten shape plots. Therefore, they provide a useful indication of the extent to which each \gls{gam} can vary in a certain feature region with small changes to the input data. In this regard, it can be noted that sparsely populated feature regions generally lead to much greater variability than densely populated feature regions, with each \gls{gam} showing a slightly different behavior depending on the feature and feature regions.

\newpage
\clearpage

\subsection{Degree of Smoothness Based on the Number of Splines}
\label{app:visual_output_n_splines}

\begin{figure}[htp!]
\centering
    \includegraphics[width=0.9\textwidth]{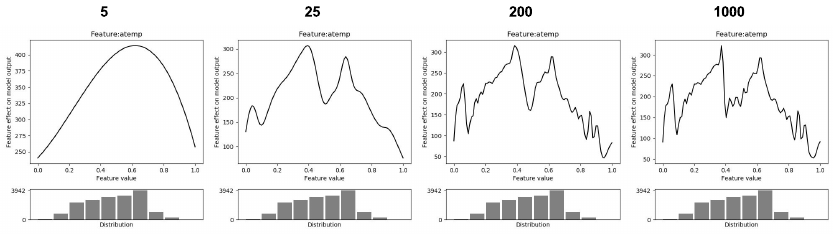}
    \caption{Generated shape functions for \gls{psplines} model using a varying number of splines.}
    \label{fig:n_splines_comparison}
\end{figure}

In Figure~\ref{fig:n_splines_comparison}, we visualize the impact of the hyperparameter "number of splines" on the degree of smoothness for a spline-based \gls{gam}. The results are based on the \textit{atemp} feature of the \textit{bike} dataset using the \gls{psplines} model from the pyGAM package \citep{serven_pygam_2021}. It can be seen that the greater the number of splines, the greater the flexibility of the model and the lower the degree of smoothness of the resulting shape functions. 

\newpage
\clearpage

\subsection{Visual Outputs for ExNN and NAM}
\label{app:visual_output_exnn_nam}

\begin{figure}[htp!]
\centering
    \includegraphics[width=0.9\textwidth]{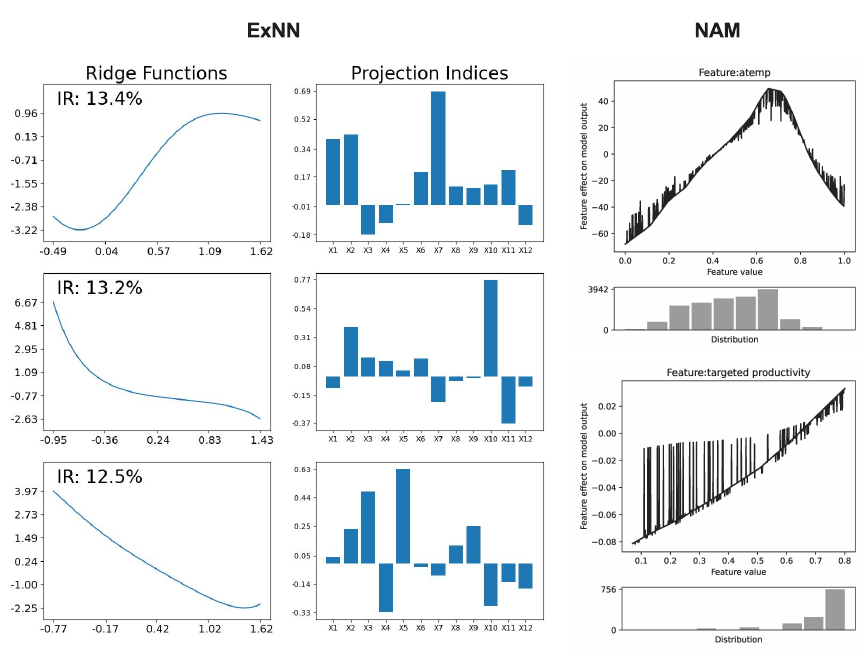}
    \caption{Visual model outputs for ExNN (left) based on the bike dataset and NAM (right) for the bike dataset (top) and the productivity dataset (bottom).}
    \label{fig:plot2_exnn_and_nam}
\end{figure}

In Figure~\ref{fig:plot2_exnn_and_nam}, we show exemplary visual outputs for the two models \gls{exnn} (left) and \gls{nam} (right). 

The demonstration example of \gls{exnn} is based on the \textit{bike} dataset. The three line plots on the left represent three trained shape/ridge functions that are captured by the structure of the underlying projection-based additive index model. This means that not only a single feature is covered by a corresponding shape function, but that a whole set of features can provide partial contributions to each shape function. For example, the features \textit{X7}, \textit{X2}, and \textit{X1} have a high positive contribution to the first shape function, whereas feature \textit{X10} has a high positive contribution to the second shape function. As a result, the overall model output is barely interpretable in a meaningful way because it is not clear what a single shape function may represent. 

The \gls{nam}, on the other hand, suffers from strong overfitting in all our experiments, which could not be resolved even with extensive hyperparameter tuning and additional model implementations. As a result, the model exhibits below-average performance and produces extremely jagged shape functions that are hardly able to convey a comprehensible decision logic. In Figure~\ref{fig:plot2_exnn_and_nam}, we show two exemplary shape plots from two different datasets. The upper shape plot is based on the \textit{bike} dataset and the lower shape plot is based on the \textit{productivity} dataset. In both cases, it can be seen that the noisy shape functions raise concerns regarding their ability to facilitate comprehensible model insights.

\newpage
\clearpage

\section{Model Assessment via Interpretability Criteria}
\label{app:interpret_criteria}

To evaluate the level of model-based interpretability, we adopted the qualitative assessment framework proposed by \citet{Sudjianto2021DesigningII}. At its core, the framework includes six evaluation criteria, which are summarized with their original definitions in Table~\ref{tab:original_interpret_metrics}. In addition, the framework includes a scoring system with two rating columns for global and local assessments and a four-point rating mechanism for each criterion. Originally, the following four rating levels are proposed:

\begin{itemize}
    \item The model automatically meets the criterion (4 points).
    \item The model can be easily constrained to meet the criterion (3 points).
    \item The model can be roughly constrained to meet the criterion (2 points).
    \item The model cannot meet the criterion (1 point).
\end{itemize}

\noindent During the use of the framework, we found that some specifications created ambiguity in the model evaluation process. Therefore, some adjustments have been made. On the one hand, we adjusted the scoring system by combining global and local considerations into a single view to assess both aspects simultaneously. At the same time, we simplified the scales from four to three levels. This was necessary because we could not clearly distinguish a situation where a model could be "easily" constrained (3 points) from a situation where a model could be "roughly" constrained (2 points). Therefore, to avoid any ambiguity, we merged the two levels and added the requirement that a model modification (i.e., model constraint) must be facilitated by predefined hyperparameters. As a result, the following three levels were defined:

\begin{itemize}
    \item The model meets the criterion by default (2 points).
    \item The model meets the criterion after adjusting related hyperparameters (1 point).
    \item The model cannot meet the criterion (0 points).
\end{itemize}

\noindent On the other hand, we also refined the evaluation criteria to increase their conceptual sharpness. This was necessary in order to achieve a better understanding of each criterion among all three raters involved in the model evaluation process. For example, it turned out that there was a disagreement on the \textit{visualizability} criterion. While one rater considered the general ability of a model to provide supporting plot functions, another rater considered the extent to which a model provides shape functions that unambiguously visualize how a particular feature with all feature values affects the predicted outcome. Thus, to reach a better consensus on all criteria, we discussed each original definition and developed a revised, more concrete understanding after going through all individual evaluation results from the first round of scoring. The adjusted definitions are summarized in Table~\ref{tab:original_interpret_metrics}.

\begin{table}[htbp!]
\centering
\caption{Overview of interpretability criteria with original and adjusted definitions.}
\label{tab:original_interpret_metrics}
\scriptsize
\begin{tabular}{p{1.8cm}p{5.7cm}p{7.0cm}}
     \toprule
     \textbf{Criteria} & \textbf{Original Definition \citep{Sudjianto2021DesigningII}} & \textbf{Adjusted Definition}\\
     \midrule
     \emph{Additivity} & Whether/how model takes the additive or modular form. & Whether the model allows for additive decomposition and modular aggregation of individual features so that each feature effect can be analyzed separately. \\[8mm] 
     \emph{Sparsity} &  Whether/how features or model components are regularized. & Whether the model uses an internal selection or regularization mechanism to identify a relevant subset of features, with the goal of maximizing their relevance while reducing model complexity.\\[8mm]
     \emph{Linearity} &  Whether/how feature effects are linear (constant as special case). & Whether the model is able to capture linear or piecewise linear relationships between the features and the target, despite the presence of noise in the dataset\\[8mm]
     \emph{Smoothness} &  Whether/how feature effects are continuous and smooth. & Whether the model avoids erratic curves and abrupt jumps by ensuring that small changes in the input lead to small changes in the output.\\[8mm]
     \emph{Monotonicity} &  Whether/how feature effects can be modeled to be monotone. & Whether monotonic constraints can be injected into the model to maintain a steadily increasing or decreasing trend in the feature-target relationships.\\[8mm]
     \emph{Visualizability} &  Whether/how the feature effects can be directly visualized. & Whether changes in the model's output can be visually understood unambiguously without re-evaluating the overall model when modifying a single feature.\\
     \bottomrule
\end{tabular}
\end{table}

\vspace{2.5mm}

\noindent Furthermore, all raters agreed on the following principles to avoid ambiguities in the scoring process: 

\begin{itemize}
    \item Random forest, XGBoost, and CatBoost can be converted into simple decision trees by adjusting certain hyperparameters (e.g., number of estimators) to allow a higher degree of interpretability. However, this means that they no longer follow their actual model idea of parallel or sequentially combined subtrees, which is why we have not considered/allowed such model adjustments.
    
    \item The multi-layer perceptron can be converted into linear models by adjusting certain hyperparameters (e.g., single-layer architecture) to allow a higher degree of interpretability. However, this means that it no longer follows the actual model idea of a multi-layer neural network, which is why we have not considered/allowed such model adjustments.
    
    \item The \gls{exnn} can be converted into a \gls{nam}-like additive network as well as into a simple linear model by adjusting certain hyperparameters/model configurations. However, this means that it no longer follows the actual model idea of a projection-based network architecture, which is why we have not considered/allowed such model adjustments.

    \item Random forest, XGBoost, CatBoost, and TabNet provide internal functions to compute feature importance scores, which indicate how much each feature contributes to the overall output of a prediction model. However, since feature importance scores do not reveal the modeled relationships between the input features and the target, we do not consider them to be a means of increasing the models' interpretability. For this reason, the models do not receive any points in the evaluation of the visualization criteria.
\end{itemize}

\begin{figure}
    \centering
    \includegraphics[width=1\textwidth]{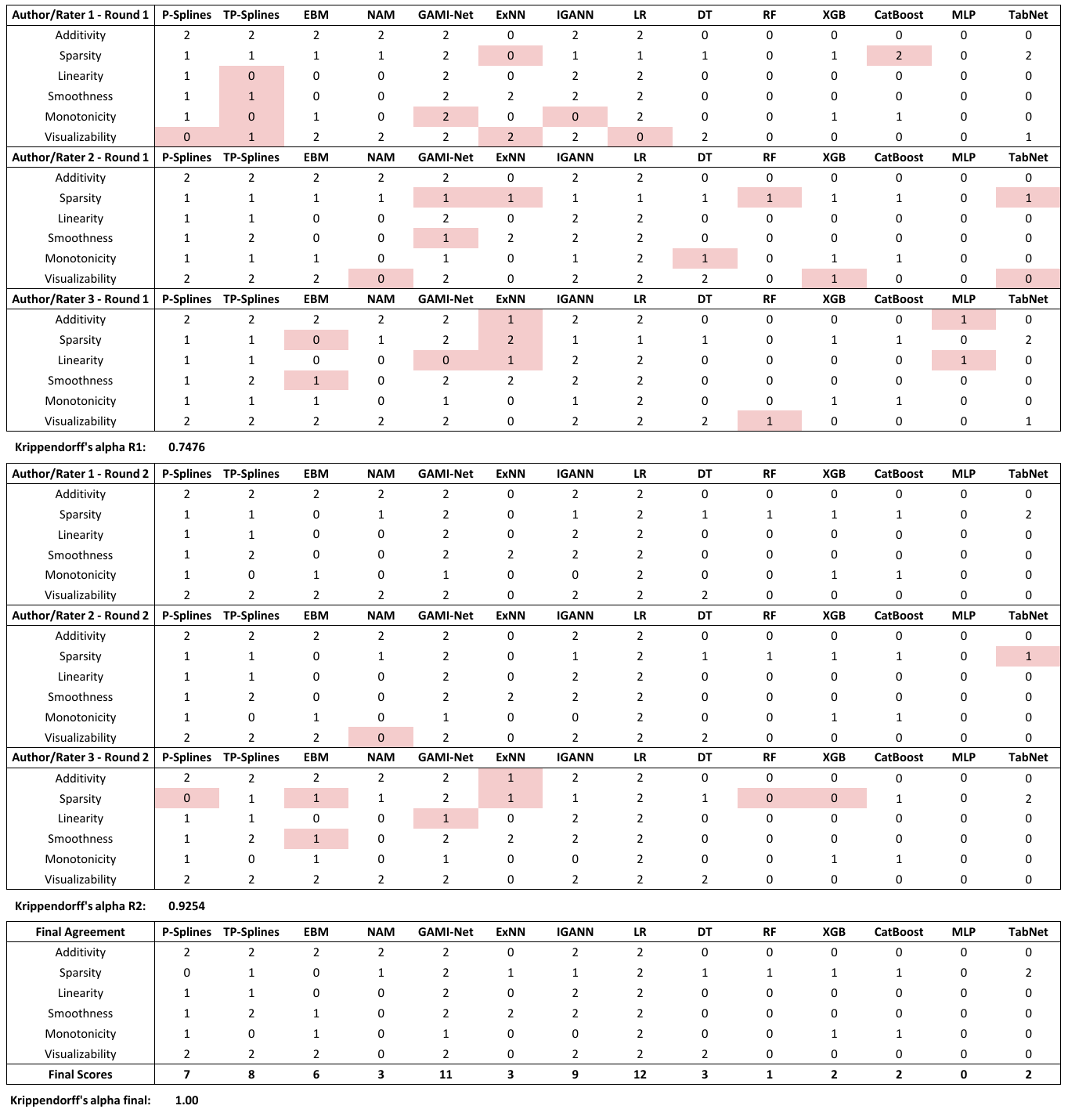}
    \caption{Results of the interpretability assessment for all three rounds of scoring. The cells highlighted in red represent cases of disagreement between the raters.}
    \label{fig:disagreement_cases}
\end{figure}

\noindent The results of all three rounds of scoring are summarized in Figure~\ref{fig:disagreement_cases}. Cells highlighted in red represent cases of disagreement. Whenever there was disagreement between the raters or uncertainty about the effect of individual model properties or the impact of certain hyperparameters, relevant documents and related publications of the corresponding models were consulted. If it was not possible to reach a clear conclusion, additional experiments were performed to systematically investigate the effects of specific hyperparameters (e.g., the effect of regularization mechanisms on model sparsity). In Table~\ref{tab:interpret_scoring+hp}, we list the relevant hyperparameters of the models that can be adjusted to satisfy the respective interpretability criteria, leading to 1 out of 2 possible points in the scoring process.

It should also be noted that \gls{nam} and \gls{exnn} receive 0 points for the visualizability criterion because it was not possible to clearly understand from the visualizations how a change in the input affects the predicted output. We illustrate this aspect for both models with demonstration examples in Appendix~\ref{app:visual_output}.

\begin{table}[htpb!]
\centering
\caption{Final results of the interpretability assessment with reference to relevant hyperparameters for adequate model adjustments.}
\label{tab:interpret_scoring+hp}
\resizebox{\columnwidth}{!}{%
\begin{tabular}{lcccccccccccccc}
\toprule
& \multicolumn{9}{c}{\textbf{Interpretable Models}} & \multicolumn{4}{c}{\textbf{Black-box Models}} &  \\ \cmidrule(l){2-10} \cmidrule(l){11-15} 
& \multicolumn{7}{c}{\textbf{GAMs}} & \multicolumn{2}{c}{\textbf{Traditional}} & & & \\ \cmidrule(l){2-8} \cmidrule(l){9-10}
 \textbf{Criterion}  & \textbf{P-Splines} & \textbf{TP-Splines} & \textbf{EBM} & \textbf{NAM} & \textbf{GAMI-Net} & \textbf{ExNN} & \textbf{IGANN} & \textbf{LR} & \textbf{DT} & \textbf{RF} & \textbf{XGB} & \textbf{CatBoost} & \textbf{MLP} & \textbf{TabNet}  \\ \midrule
Additivity      & 2                           & 2                   & 2                 & 2            & 2                              & 0             & 2              & 2            & 0                         & 0              & 0                     & 0                         & 0            & 0               \\ \hline
Sparsity        & 0                           & 1                   & 0                 & 1            & 2                              & 1             & 1              & 2            & 1                         & 1              & 1                     & 1                         & 0            & 2               \\
                &                             & smooth.terms        &                   & feature      &                                & l1\_proj,     & sparse         &              & max\_leaf\_nodes,         & n\_estimators, & gamma, alpha,         & l2\_leaf\_reg, depth              &              &                 \\
                &                             &                     &                   & dropout      &                                & l1\_subnet    &                &              & max\_features, max\_depth & max\_depth     & lambda                &  &              &                 \\ \hline
Linearity       & 1                           & 1                   & 0                 & 0            & 2                              & 0             & 2              & 2            & 0                         & 0              & 0                     & 0                         & 0            & 0               \\
                & pygam.terms.l               & smooth.terms        &                   &              &                                &               &                &              &                           &                &                       &                           &              &                 \\ \hline
Smoothness      & 1                           & 2                   & 1                 & 0            & 2                              & 2             & 2              & 2            & 0                         & 0              & 0                     & 0                         & 0            & 0               \\
                & lam, n\_splines             &                     & smoothing\_rounds &              &                                &               &                &              &                           &                &                       &                           &              &                 \\ \hline
Monotonicity    & 1                           & 0                   & 1                 & 0            & 1                              & 0             & 0              & 2            & 0                         & 0              & 1                     & 1                         & 0            & 0               \\
                & constraints:monotonic\_inc, &                     & monotonize        &              & mono\_increasing\_list={[}{]}, &               &                &              &                           &                & monotone\_constraints & monotone\_constraints     &              &                 \\
                & constraints:monotonic\_dec  &                     &                   &              & mono\_decreasing\_list={[}{]}  &               &                &              &                           &                &                       &                           &              &                 \\ \hline
Visualizability & 2                           & 2                   & 2                 & 0            & 2                              & 0             & 2              & 2            & 2                         & 0              & 0                     & 0                         & 0            & 0               \\ \hline \\
\textbf{Total score}     & \textbf{7}                           & \textbf{8}                   & \textbf{6}                 & \textbf{3}            & \textbf{11}                             & \textbf{3}             & \textbf{9}              & \textbf{12}           & \textbf{3}                         & \textbf{1}              & \textbf{2}                     & \textbf{2}                         & \textbf{0}            & \textbf{2}               \\ \bottomrule \bottomrule
\end{tabular}%
}
\end{table}

\newpage
\clearpage

\section{Additional Details for Performance-Interpretability Evaluation}
\label{app:trade-off}

The following figures illustrate different perspectives on the performance-interpretability evaluation from Section~\ref{sec:results_summary}. While Figure~\ref{fig:perf_interpret_trade-off} in the main paper shows the average performance results of classification and regression tasks in combination, Figure~\ref{fig:trade-off_classification_regression} below shows the performance results of classification and regression tasks separately. However, compared to the average result of both tasks, the individual diagrams show only small deviations. 

\begin{figure}[htp!]
    \centering
    \begin{subfigure}[b]{0.49\textwidth}
        \centering
        \includegraphics[width=\textwidth]{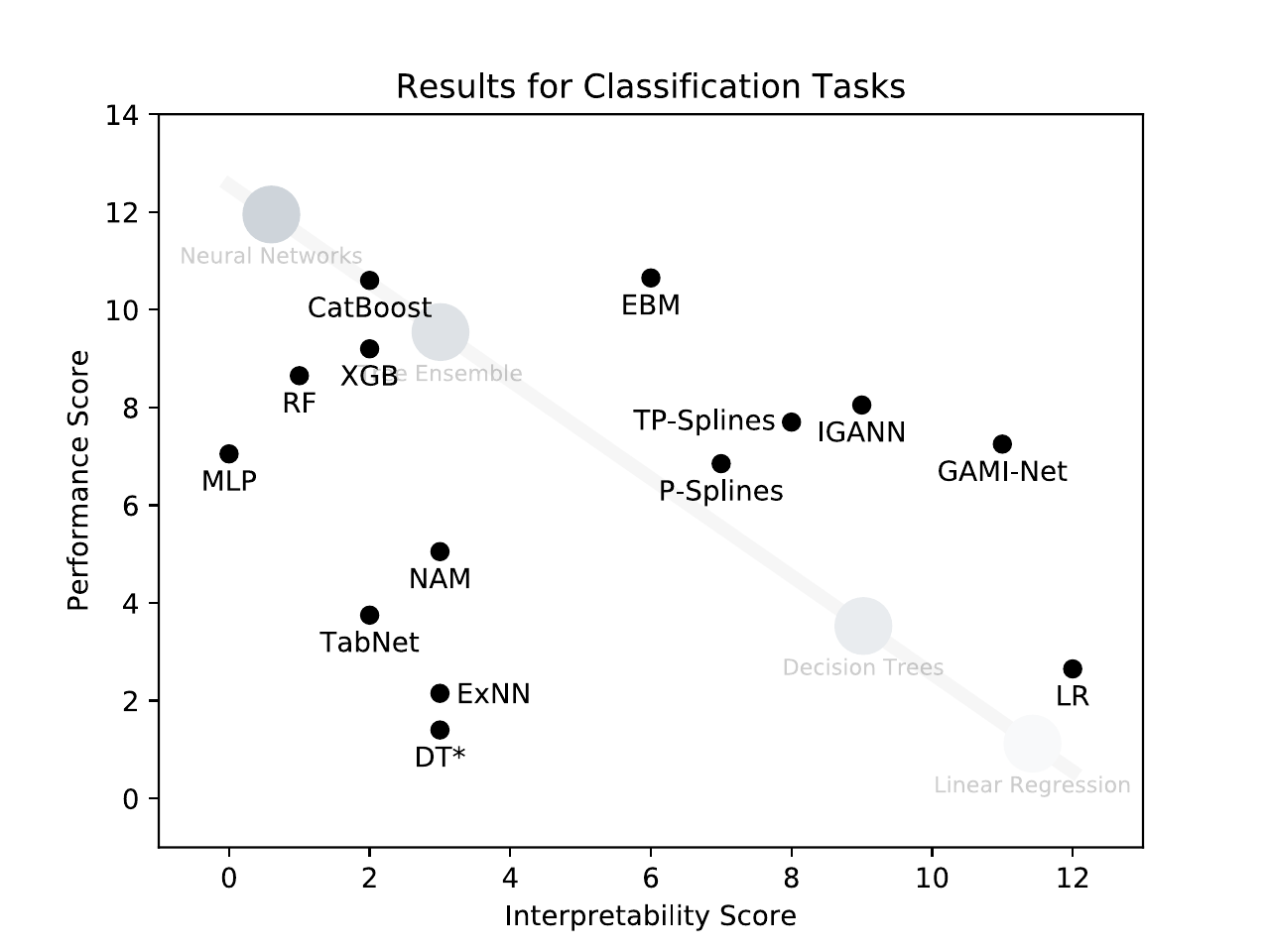}
    \end{subfigure}
    \hfill
    \begin{subfigure}[b]{0.49\textwidth}
        \centering
        \includegraphics[width=\textwidth]{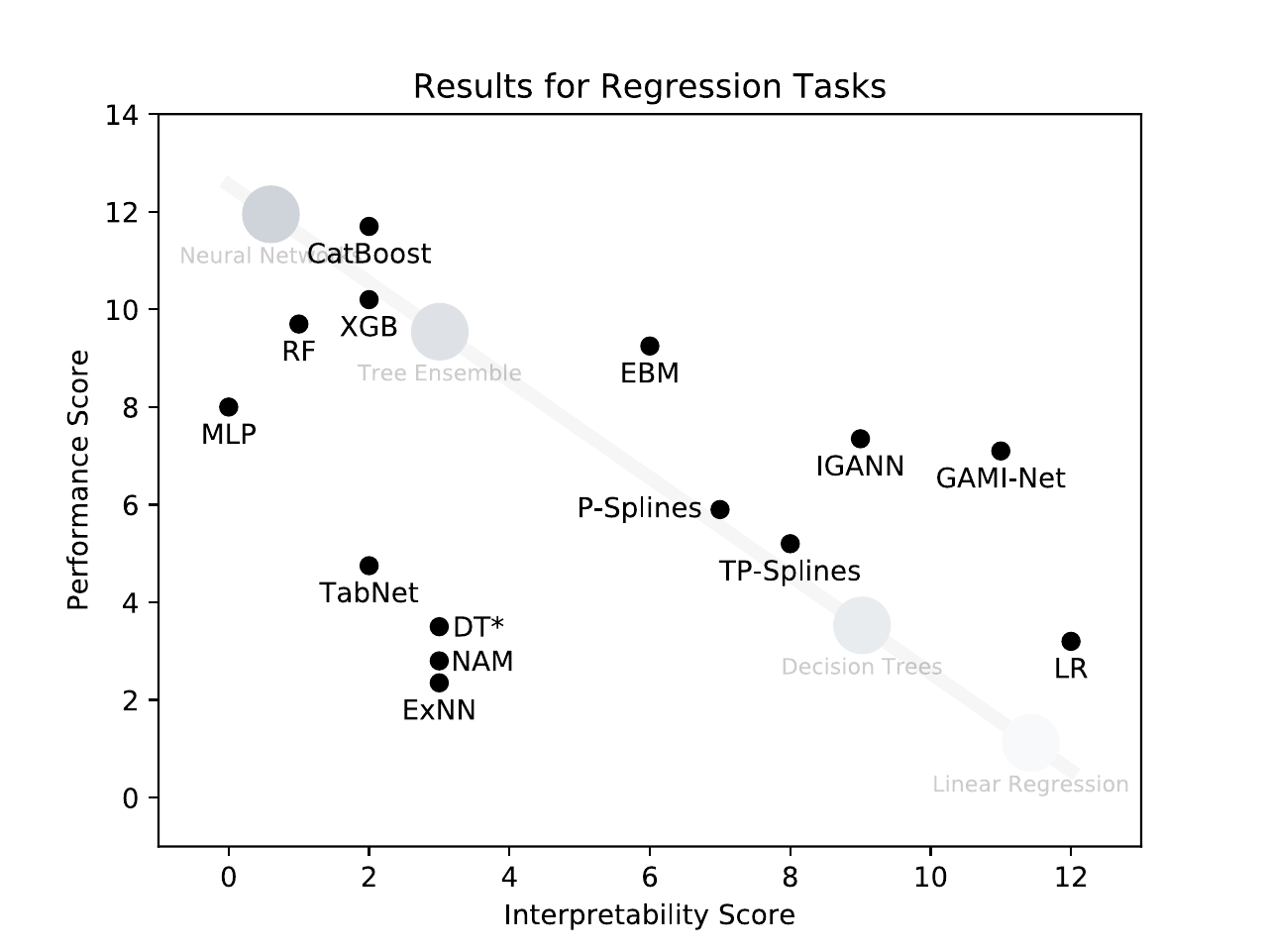}
    \end{subfigure}
    \caption{Comparison of the performance-interpretability evaluation between the average performance results of the classification tasks (left) and the average performance results of the regression tasks (right).}
    \label{fig:trade-off_classification_regression}
\end{figure}

\noindent Furthermore, it should be noted that our evaluation framework is currently based on the assumption that all six interpretability criteria (i.e., additivity, sparsity, linearity, smoothness, monotonicity, and visualizability) are equally important. However, depending on the specifics of different application domains, some criteria may have a higher weight, while others may be neglected. For this purpose, we consider two additional scenarios with modified, context-specific weights to illustrate the impact on the model ranking within the performance-interpretability diagram.

\begin{figure}[h!]
    \centering
    \begin{subfigure}[b]{0.49\textwidth}
        \centering
        \includegraphics[width=\textwidth]{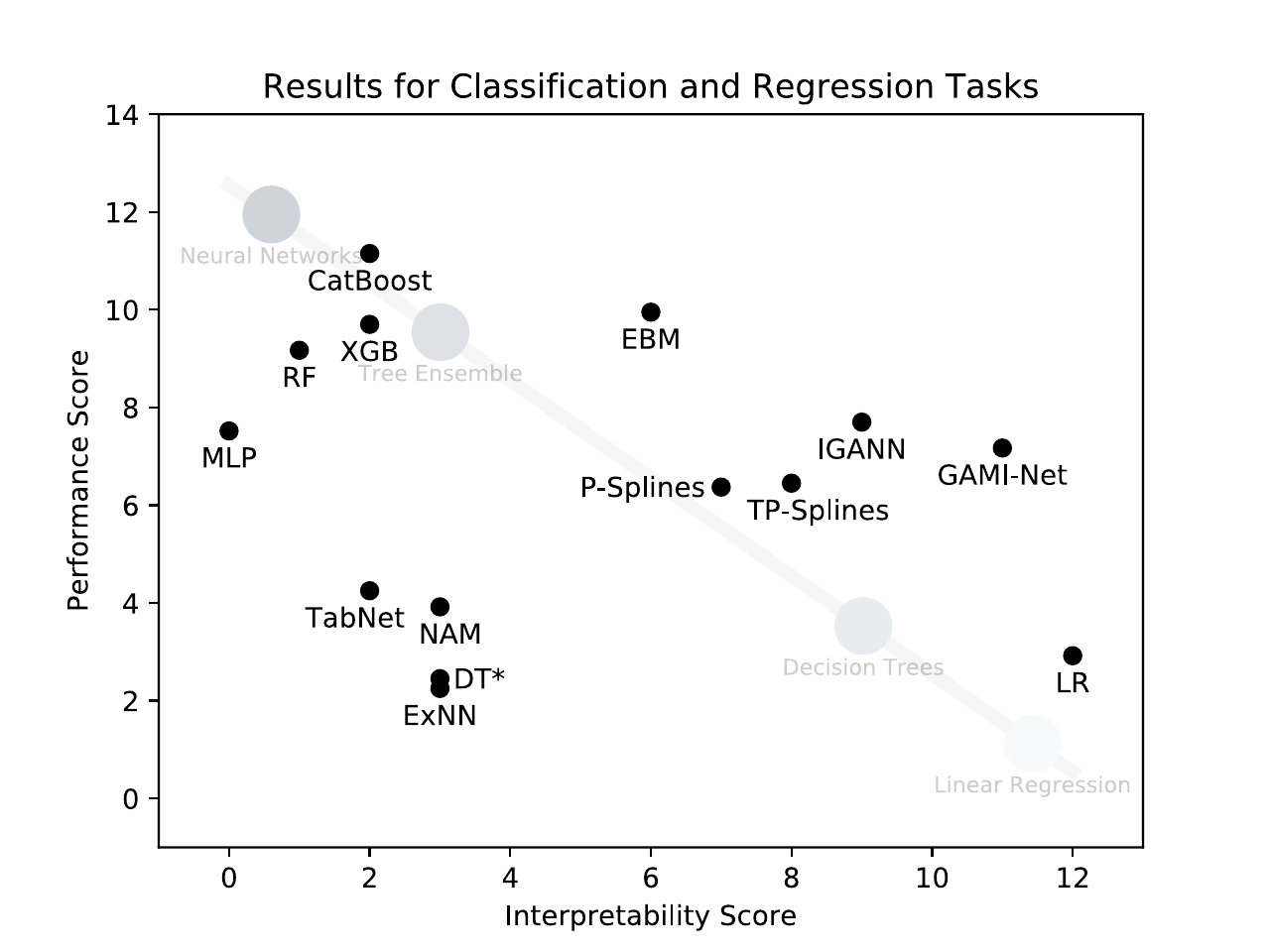}
    \end{subfigure}
    \hfill
    \begin{subfigure}[b]{0.49\textwidth}
        \centering
        \includegraphics[width=\textwidth]{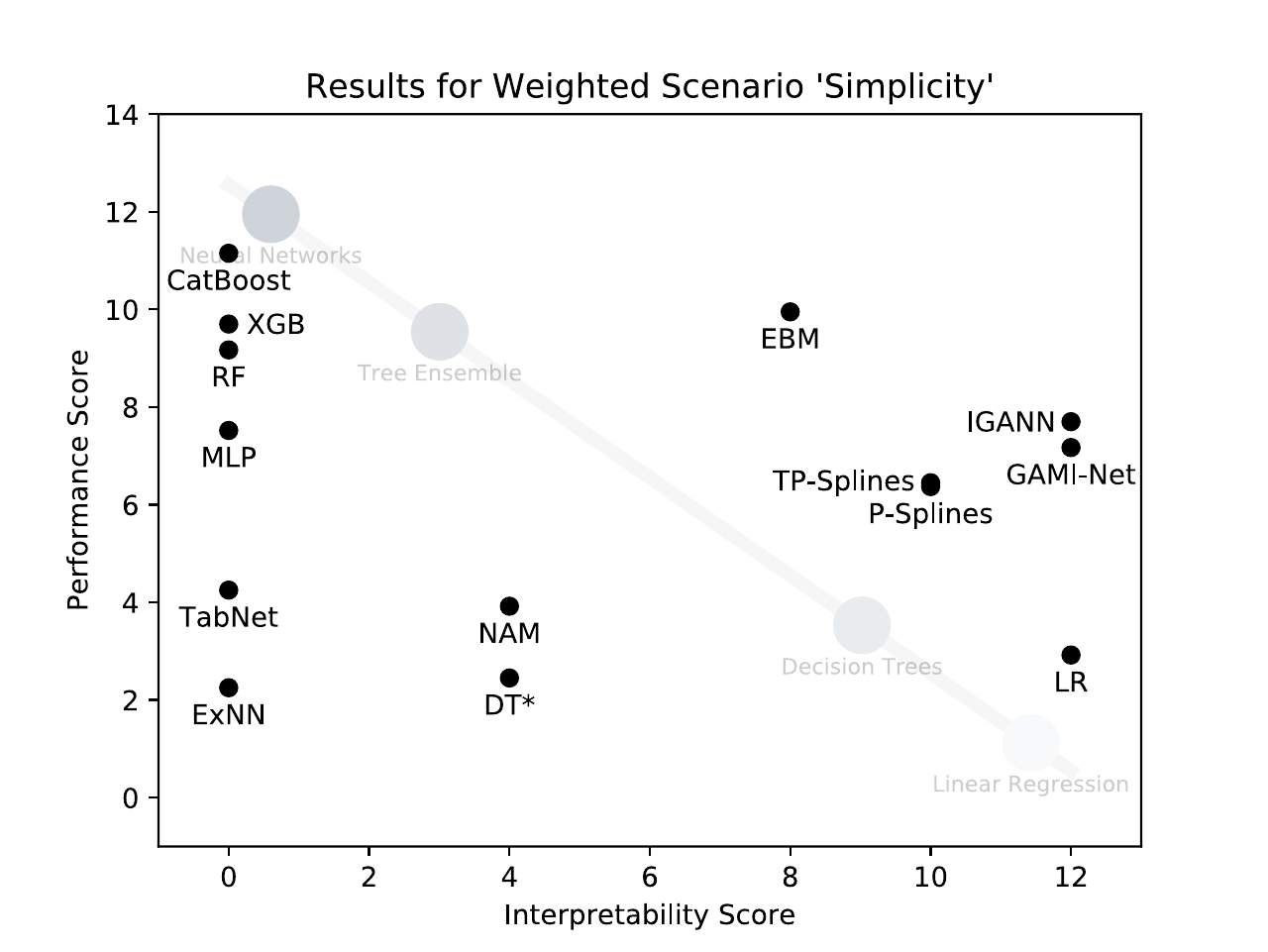}
    \end{subfigure}
    \caption{Comparison of the performance-interpretability evaluation between equally weighted interpretability criteria (left) and an adjusted weighting (right). In the adjusted scenario, additivity, linearity, and visualizability have a weighting of 10, and all other criteria a weighting of 0.}
    \label{fig:trade-off_re-weighting1}
\end{figure}

In the first scenario, we consider a use case with the requirements that a model has to be characterized by simple (ideally linear/piecewise linear) feature effects that can be analyzed individually for instant decision-making with the possibility to visually comprehend how changes in the feature values affect the predicted outcome. Consequently, the criteria of additivity, linearity, and visualizability are assigned a higher weight of 10, while the remaining criteria are considered irrelevant and are assigned a weight of 0. The resulting performance-interpretability diagram is shown in Figure~\ref{fig:trade-off_re-weighting1} on the right. The re-weighting process\footnote{To retain the original scaling of the $x$-axis from 0 - 12, the re-weighted values are scaled as follows and rounded to whole numbers: \(x_i' = \left\lfloor \left( ({x_i - \min(x)})\div{(\max(x) - \min(x))} \right) \times 12 \right\rfloor\).} results in a stronger separation between the group of black-box models and the majority of \glspl{gam}. Furthermore, some models like \gls{exnn}, \gls{nam}, and \gls{dt} remain in a lower score region.

\begin{figure}[h!]
    \centering
    \begin{subfigure}[b]{0.49\textwidth}
        \centering
        \includegraphics[width=\textwidth]{figures_appendix/Figure_A6+7_tradeoff-overall.pdf}
    \end{subfigure}
    \hfill
    \begin{subfigure}[b]{0.49\textwidth}
        \centering
        \includegraphics[width=\textwidth]{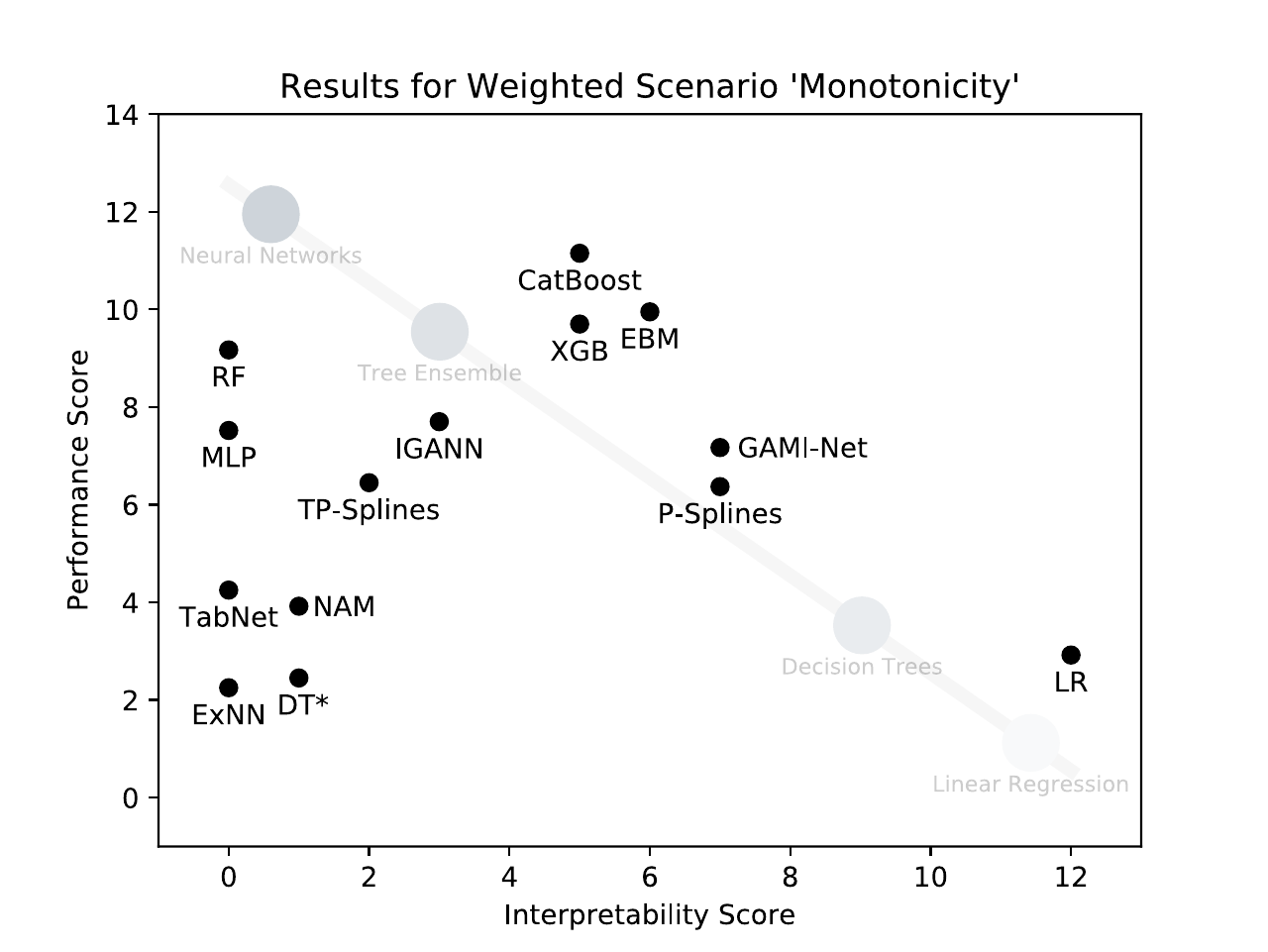}
    \end{subfigure}
    \caption{Comparison of the performance-interpretability evaluation between equally weighted interpretability criteria (left) and an adjusted weighting (right). In the adjusted scenario, additivity, linearity, and visualizability have a weighting of 1, monotonicity an increased weighting of 10, and all other criteria a weighting of 0.}
    \label{fig:trade-off_re-weighting2}
\end{figure}

In a second scenario, we consider a use case that is characterized by more domain-specific requirements, such as those that occur in the credit scoring area. Due to strict regulations, it is usually not permitted, for example, to use non-monotonic models for credit scoring decisions. This shall ensure that no misleading feature effects are learned, for example, that piecewise increasing debt-to-income ratios or piecewise increasing numbers of delinquencies lead to a lower risk of credit default. For this reason, monotonicity is given a higher weight of 10 in this scenario, while additivity, linearity, and visualizability retain the weights of 1. The remaining criteria are assigned weights of 0. The resulting performance-interpretability diagram is shown in Figure~\ref{fig:trade-off_re-weighting2} on the right. While linear/logistic regression is still the most interpretable model, all other models move much closer together showing only low to medium interpretability scores. 

The two scenarios described above represent only two exemplary use cases and illustrate that different perspectives on the performance-interpretability-diagram are conceivable depending on the context-specific weighting of the individual interpretability criteria. However, due to the strong predictive performance of several \glspl{gam}, such as \gls{ebm}, \gls{gaminet}, and \gls{igann}, they are always among the most promising models, as they can satisfy both central requirements to a high degree.

\end{document}